\documentclass[journal]{IEEEtran}

\usepackage{amsmath,amsfonts}
\usepackage{algorithmic}
\usepackage{algorithm}
\usepackage{array}
\usepackage[caption=false,font=normalsize,labelfont=sf,textfont=sf]{subfig}
\usepackage{textcomp}
\usepackage{stfloats}
\usepackage{amsmath}
\DeclareMathOperator*{\argmax}{argmax}
\usepackage{url}
\usepackage{verbatim}
\usepackage{graphicx}
\usepackage{cite}
\usepackage{booktabs}
\usepackage{pgfplots}
\usetikzlibrary{patterns}
\usepackage{multirow}
\usepackage{makecell}
\newcolumntype{d}[1]{D{.}{.}{#1}}

\usepackage[normalem]{ulem}

\begin{document}

\title{MiniZero: Comparative Analysis of AlphaZero and MuZero on Go, Othello, and Atari Games}

\author{
Ti-Rong Wu, \IEEEmembership{Member, IEEE},
Hung Guei, \IEEEmembership{Member, IEEE},
Pei-Chiun Peng,
Po-Wei Huang,
Ting Han Wei,
Chung-Chin Shih, \IEEEmembership{Member, IEEE},
and Yun-Jui Tsai
\thanks{
This research is partially supported by the National Science and Technology Council (NSTC) of the Republic of China (Taiwan) under Grant NSTC 111-2222-E-001-001-MY2 and 112-2634-F-A49-004. \textit{(Corresponding author: Ti-Rong Wu)}
}

\thanks{Ti-Rong Wu and Hung Guei are with the Institute of Information Science, Academia Sinica, Taipei, Taiwan (e-mail: tirongwu@iis.sinica.edu.tw; hguei@iis.sinica.edu.tw).}
\thanks{Pei-Chiun Peng and Po-Wei Huang are with the Department of Computer Science, National Yang Ming Chiao Tung University, Hsinchu, Taiwan (e-mails: pcpeng.ee11@nycu.edu.tw; a311551048.cs11@nycu.edu.tw).}
\thanks{Ting Han Wei is with the School of Informatics, Kochi University of Technology, Kami City, Japan (e-mail: tinghan.wei@kochi-tech.ac.jp).}
\thanks{Chung-Chin Shih is with the Institute of Information Science, Academia Sinica, Taipei, Taiwan (e-mail: rockmanray@iis.sinica.edu.tw)}
\thanks{Yun-Jui Tsai is with the Department of Computer Science, National Yang Ming Chiao Tung University, Hsinchu, Taiwan (e-mails: b08202011.cs12@nycu.edu.tw).}
}

\date{October 2023}

\markboth{}
{Shell \MakeLowercase{\textit{et al.}}: A Sample Article Using IEEEtran.cls for IEEE Journals}


\maketitle

\begin{abstract}
This paper presents \textit{MiniZero}, a zero-knowledge learning framework that supports four state-of-the-art algorithms, including AlphaZero, MuZero, Gumbel AlphaZero, and Gumbel MuZero.
While these algorithms have demonstrated super-human performance in many games, it remains unclear which among them is most suitable or efficient for specific tasks. 
Through \textit{MiniZero}, we systematically evaluate the performance of each algorithm in the two board games, 9x9 Go and 8x8 Othello, as well as 57 Atari games. 
For the two board games, using more simulations generally results in higher performance.
However, the choice between AlphaZero and MuZero may differ based on game properties.
For Atari games, both MuZero and Gumbel MuZero are worth considering.
Since each game has unique characteristics, different algorithms and simulations yield varying results. 
In addition, we introduce an approach, called progressive simulation, which progressively increases the simulation budget during training to allocate computation more efficiently.
Our empirical results demonstrate that progressive simulation achieves significantly superior performance in the two board games. 
By making our framework and trained models publicly available, this paper contributes a benchmark for future research on zero-knowledge learning algorithms, assisting researchers in algorithm selection and comparison against these zero-knowledge learning baselines.
Our code and data are available at https://rlg.iis.sinica.edu.tw/papers/minizero.
\end{abstract}

\begin{IEEEkeywords}
AlphaZero, Atari games, deep reinforcement learning, Go, Gumbel AlphaZero, Gumbel MuZero, MuZero
\end{IEEEkeywords}

\section{Introduction}
\IEEEPARstart{A}{lphaGo}, AlphaGo Zero, AlphaZero, and MuZero are a family of algorithms that have each made groundbreaking strides in model-based deep reinforcement learning. 
AlphaGo \cite{silver_mastering_2016} was the first program to achieve super-human performance in the game of Go, demonstrated by its victory over world champion Lee Sedol in 2016.
Building on this triumph, AlphaGo Zero \cite{silver_mastering_2017} introduced the concept of \textit{zero-knowledge learning} by removing the need for external human knowledge. More specifically, it does not use expert game records for move prediction. Instead, AlphaGo Zero is trained solely with self-play from scratch. Experiments at the time indicate that this approach resulted in stronger play, with AlphaGo Zero beating AlphaGo 100-0.
Additionally, without the need for domain-specific knowledge, self-play training could be conducted across a much wider class of games.
AlphaZero \cite{silver_general_2018} demonstrated that zero-knowledge learning can be extended to chess, shogi, and Go, beating state-of-the-art programs in each game. 
Lastly, MuZero \cite{schrittwieser_mastering_2020} improves upon its predecessors by removing knowledge of the task environment. This includes knowledge of game rules and state transitions. By learning additional representation and dynamics models, MuZero does not have to interact with environments as it is planning ahead. This allows it to master both board games and Atari games, opening up the potential for extension to complex real-world scenarios.

While the zero-knowledge learning algorithms, AlphaZero and MuZero, have achieved super-human performance in many games, recent research highlighted that neither AlphaZero nor MuZero ensures policy improvement unless all actions are evaluated at the root of the Monte Carlo search tree \cite{danihelka_policy_2022}. This implies the potential for failure when training with limited simulations.
Indeed, common settings for both algorithms use 800 Monte Carlo tree search (MCTS) \cite{browne_survey_2012, coulom_efficient_2007, kocsis_bandit_2006} simulations for each move when playing board games, which can be computationally costly.
Gumbel Zero algorithms, \textit{Gumbel AlphaZero} and \textit{Gumbel MuZero}, incorporate Gumbel noise into the original algorithms to guarantee policy improvement \cite{danihelka_policy_2022}, even with fewer simulations.
The two algorithms match the performance of AlphaZero and MuZero for both board games and Atari games.
Moreover, Gumbel Zero demonstrates significant performance with as low as only two simulations.

Since AlphaGo's publication, many open-source projects have emerged to reproduce these algorithms for different purposes.
Several projects are particularly dedicated to reproducing either the AlphaZero or MuZero algorithms for specific games.
For instance, KataGo \cite{wu_accelerating_2020a}, LeelaZero \cite{pascutto_leela_2023}, ELF OpenGo \cite{tian_elf_2019}, and CGI \cite{wu_accelerating_2020} primarily apply the AlphaZero algorithm to the game of Go, while Leela Chess Zero \cite{gary_leela_2023} is designed for chess.
Moreover, EfficientZero \cite{ye_mastering_2021} introduces enhancements and implements the MuZero algorithm for Atari games.

Meanwhile, other projects aim to provide a general framework that accommodates a wide range of games.
OpenSpiel \cite{lanctot_openspiel_2020}, Polygames \cite{cazenave_polygames_2020}, and AlphaZero General \cite{thakoor_learning_2016} all offer frameworks based on the AlphaZero algorithm, supporting various board games.
In contrast, MuZero General \cite{wernerduvaud_muzero_2019} provides training across both board and Atari games based on the MuZero algorithm.
The recently introduced LightZero \cite{niu2023lightzero} framework supports extensive zero-knowledge learning algorithms, such as AlphaZero, MuZero, and two Gumbel Zero algorithms for various environments.

Despite the variety of open-source zero-knowledge learning projects available, there is no available analysis on which algorithm -- AlphaZero, MuZero, Gumbel AlphaZero, or Gumbel MuZero -- is most suitable or efficient for specific tasks.
To answer this question, this paper introduces \textit{MiniZero}, a zero-knowledge learning framework that supports all four algorithms.
Based on this framework, we conduct comprehensive experiments and offer detailed analyses of performances across algorithms and tasks.
The evaluated games include the two board games, 9x9 Go and 8x8 Othello, as well as 57 Atari games.
Furthermore, we propose a novel approach, named \textit{progressive simulation}, which gradually increases the simulation budget during Gumbel Zero training to allocate computation more efficiently.
Notably, with progressive simulation, both Go and Othello achieve higher performance compared to the original Gumbel Zero with a fixed simulation budget.
We have also made our framework and all trained models publicly accessible\footnote{Available at: https://github.com/rlglab/minizero/tree/d29ef42}, in the hopes that our empirical findings can serve as benchmarks for the community, and assist future researchers in comparing novel algorithms against these zero-knowledge learning baselines.

\section{Background}
This section reviews four popular zero-knowledge learning algorithms, including the AlphaZero algorithm in Section \ref{bg:az}, the MuZero algorithm in Section \ref{bg:mz}, and both the Gumbel AlphaZero and Gumbel MuZero algorithms in Section \ref{bg:gz}.
The comparison between these algorithms is summarized in Table \ref{tab:alg-compare}.

\subsection{AlphaZero}\label{bg:az}
AlphaZero \cite{silver_general_2018} is a zero-knowledge learning algorithm that masters a variety of board games without using human knowledge.
The network architecture consists of several residual blocks \cite{he_deep_2016} and two heads, including a policy head and a value head.
Given a board position, the policy head outputs a policy distribution $p$ for possible actions, while the value head predicts an estimated outcome $v$.
The training process comprises two components: self-play and optimization.

Self-play performs MCTS \cite{browne_survey_2012, coulom_efficient_2007, kocsis_bandit_2006} for all players and every move, starting from an initial board position until the end of the game.
MCTS contains three phases: selection, expansion, and backpropagation.
In the selection phase, an action $a$ is chosen for a given state $s$ using the PUCT \cite{rosin_multiarmed_2011, silver_mastering_2017} formula:
\begin{equation}\label{eq:az_puct}
a^{*} = \argmax_a \{ Q(s,a)+c_{puct} P(s,a)\frac{\sqrt{\Sigma_bN(s,b)}}{1+N(s,a)}\},
\end{equation}
where $Q(s,a)$, $P(s,a)$, and $N(s,a)$ denote the estimated value, prior probability, and visit count of $a$ at node $s$, and $c_{puct}$ is an exploration hyperparameter.
State $s$ is initialized to the root of the search tree. The chosen action $a$ is performed, upon which $s$ is set to the resulting state. This process is repeated until $s$ is a leaf node.
Next, the leaf node is evaluated by the network during the expansion phase. All children are expanded with $Q(s,a)=N(s,a)=0, P(s,a)=p_a$.
Finally, during backpropagation, the estimated outcome $v$ obtained from the value network is updated along the selection path, upwards towards the root.
Complete self-play game records are stored in a replay buffer for optimization.

Data is sampled randomly from the replay buffer to optimize the network.
Specifically, the policy network aims to learn $p$ such that it matches MCTS search policy distribution $\pi$.
Simultaneously, the value network is updated to minimize the error between $v$ and the game outcome $z$.
The optimization loss is shown in equation \eqref{eq:az_loss}:
\begin{equation}\label{eq:az_loss}
L=(z-v)^{2}-\pi^{\mathsf{T}} \log p+c||\theta||^{2},
\end{equation}
where the last term is for regularization, in which $||\theta||^{2}$ denotes the L2 normalization of the model parameters $\theta$, and $c$ is a hyperparameter.

Since AlphaZero can achieve superhuman performance without the need for human knowledge, it has been widely extended to other games \cite{cazenave_polygames_2020, lanctot_openspiel_2020} as well as non-game applications, such as optimization problems
like matrix multiplication discovery \cite{fawzi_discovering_2022} and sorting algorithm improvement \cite{mankowitz_faster_2023}.

\subsection{MuZero}\label{bg:mz}
MuZero \cite{schrittwieser_mastering_2020} builds upon the AlphaZero algorithm but incorporates neural network-based learned models to learn the environment.
These models allow MuZero to plan ahead without requiring additional environment interactions, as summarized in Table \ref{tab:alg-compare}. 
This is especially useful when environment simulators are not easily accessible or costly.
MuZero not only matches the super-human performance of AlphaZero in board games such as chess, shogi, and Go, it also achieves new state-of-the-art performance in Atari games.

\begin{table}
\caption{Comparison between AlphaZero, MuZero, Gumbel AlphaZero, and Gumbel MuZero}
    \centering
    \setlength{\tabcolsep}{5pt}
    \begin{tabular}{lcccc}
    \toprule
     & \multirowcell{2}{AlphaZero} & \multirowcell{2}{MuZero} & \multirowcell{2}{Gumbel\\AlphaZero} & \multirowcell{2}{Gumbel\\MuZero}\\
     \\
    \midrule
    Planning w/o simulator &   & V &   & V \\
    Policy improvement     &   &   & V & V \\
    Apply to board games            & V & V & V & V \\
    Apply to Atari games            &   & V &   & V \\
    \bottomrule
    \end{tabular}
    \label{tab:alg-compare}
\end{table}

MuZero employs three networks: \textit{representation}, \textit{dynamics}, and \textit{prediction}.
The prediction network, denoted by $f(s)=(p,v)$, is similar to the two-headed network used in AlphaZero.
The representation and dynamics networks play an important role in learning abstract hidden states that represent the real environment.
First, the representation network, denoted by $h(o)=s$, converts environment observations $o$ into a hidden state $s$.
Next, the dynamics network, denoted by $g(s,a)=(r,s')$, learns environment transitions. For a given hidden state $s$ and action $a$, it generates the next hidden state $s'$ and the state-action pair's associated reward $r$. 
Essentially, the dynamics network serves as a surrogate environment simulator, allowing MuZero to simulate trajectories during planning without having to sample actions.

The training process for MuZero is the same as AlphaZero, which includes both self-play and optimization.
In self-play, MCTS is executed for each move, and the completed games are stored in a replay buffer.
However, different from AlphaZero, the MCTS in MuZero first converts the environment observation into a hidden state, then conducts planning through the dynamics network.
During optimization, MuZero unrolls $K$ steps, ensuring that the model is in alignment with sequences sampled from the replay buffer.
In addition, the loss function is modified:
\begin{equation}\label{eq:mz_loss}
l=\sum_{k=0}^{K}l^{p}(\pi_k,p_k)+\sum_{k=0}^{K}l^{v}(z_k,v_k)+\sum_{k=0}^{K}l^{r}(u_k,r_k)+c||\theta||^{2},
\end{equation}
where $l^p$, $l^v$, and $l^r$ denote the policy, value, and reward losses respectively.
The policy and value losses are similar to AlphaZero.
Note that $z$ is calculated based on the n-step return in Atari games.
The reward loss minimizes the error between the predicted immediate reward $r$ and the observed immediate reward obtained from the environment $u$.

There are several enhancements for MuZero.
Sampled MuZero \cite{hubert_learning_2021} adapts MuZero for environments with continuous actions, while Stochastic MuZero \cite{antonoglou_planning_2021} modifies MuZero to support stochastic environments.
Additionally, both EfficientZero \cite{ye_mastering_2021} and Gumbel MuZero \cite{danihelka_policy_2022} introduce different methods to improve the learning efficiency.

\subsection{Gumbel AlphaZero and Gumbel MuZero}\label{bg:gz}
The Gumbel Zero algorithms \cite{danihelka_policy_2022} were proposed to guarantee policy improvement, as summarized in Table \ref{tab:alg-compare}.
This improves training performance with smaller simulation budgets.
The modifications include MCTS root node action selection, environment action selection, and policy network update.

For MCTS root node action selection, Gumbel Zero algorithms employ the Gumbel-Top-k trick \cite{kool_stochastic_2019} and the sequential halving algorithm \cite{karnin_almost_2013} to select actions at the root node.
The Gumbel-Top-k trick samples the top $k$ actions with higher $G(a)+logits(a)$, where $G(a)$ is the sampled Gumbel variable and $logits(a)$ is the unnormalized prediction generated by the policy network.
If the simulation count for the MCTS, $n$, is set to $k$, $k$ actions are sampled, using up the simulation budget. The final chosen action is the one with the highest $G(a)+logits(a)+\sigma(q(a))$, where $q(a)$ is the $Q$ value of $a$ and $\sigma$ is a monotonically increasing transformation.
On the other hand, if $n$ is set to larger than $k$, sequential halving is applied to divide the search budget into several phases. This allocates more simulations to better actions. 
In the first phase, the search starts with $k$ sampled actions. 
After each phase, the actions are sorted according to $G(a)+logits(a)+\sigma(q(a))$, where the actions in the bottom half are discarded from consideration. 
In the last phase, only one action is retained. This remaining action is the most visited action and the one that is chosen.

For the environment action selection, we explicitly select the best action from the root node utilizing the aforementioned method. 
This is in contrast to the non-Gumbel algorithms, where the action is sampled according to the search policy -- a distribution that is formed by the visit counts of the root actions.

For policy network updates, due to the very few actions sampled, we cannot use the search visit count distribution as the policy network training target.
The Gumbel trick uses Q values to derive a policy training target.
For visited children, we use the sampled Q value.
For those that have not been visited, we use the value network instead.


\section{MiniZero}
This section first presents the \textit{MiniZero} framework in Section \ref{minizero:design}. Then, section \ref{minizero:estimated_q} introduces an estimated Q value method for non-visited actions. Finally, section \ref{minizero:progressive_simulation} proposes a new training approach for Gumbel Zero algorithms.

\subsection{Framework Design}\label{minizero:design}

\begin{figure*}[!t]
    \centering
    \includegraphics[width=\linewidth]{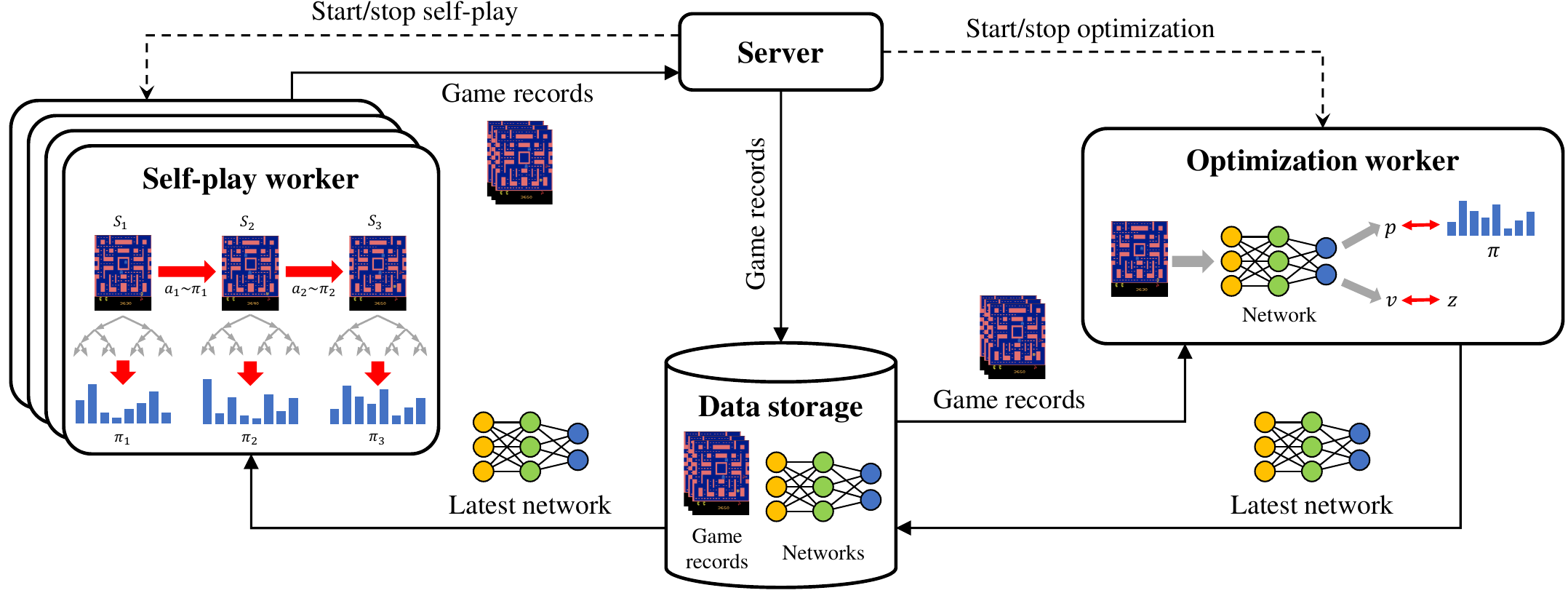}
    \captionsetup[subfigure]{justification=centering}
    \caption{The MiniZero architecture, includes four components: a server, self-play workers, an optimization worker, and data storage.}
    \label{fig:architecture}
\end{figure*}

The architecture of \textit{MiniZero} shown in Fig. \ref{fig:architecture} comprises four components, including a \textit{server}, one or more \textit{self-play workers}, an \textit{optimization worker}, and \textit{data storage}. We describe each component in the next paragraph.

The server is the core component in \textit{MiniZero}, controlling the training process and managing both the self-play and optimization workers.
The training process contains several iterations.
For each iteration, the server first instructs all self-play workers to generate self-play games simultaneously by using the latest network.
Each self-play worker maintains multiple MCTS instances to play multiple games simultaneously with batch GPU inferencing to improve efficiency.
Specifically, the self-play worker runs the selection for each MCTS to collect a batch of leaf nodes and then evaluates them through batch GPU inferencing.
Finished self-play games are sent to the server and forwarded to the data storage by the server.
Once the server accumulates the necessary self-play games, it then stops the self-play workers and instructs the optimization worker to start network updates.
The optimization worker updates the network over steps using data sampled from the replay buffer.
Generally, the number of optimized steps is proportional to the number of collected self-play games to prevent overfitting.
The optimization worker stores updated networks into the data storage.
The server then starts the next iteration. This process continues until the training reaches a predetermined iteration $I$.

The server and workers communicate using TCP connections for message exchange.
The data storage uses the Network File System (NFS) for sharing data across different machines.
This is an implementation choice; a simpler file system can suffice if distributed computing is not employed.

MiniZero supports various zero-knowledge learning algorithms, including AlphaZero, MuZero, Gumbel AlphaZero, and Gumbel MuZero. 
To ensure execution efficiency, our implementation utilizes C++ and PyTorch \cite{paszke_pytorch_2019}. To support experiments on Atari games, it also utilizes the Arcade Learning Environment (ALE) \cite{bellemare_arcade_2013}, \cite{machado_revisiting_2018}.

\subsection{MCTS Estimated Q Value for Non-Visited Actions}\label{minizero:estimated_q}
While our implementation mainly follows the original design, we introduce a modification to enhance exploration.
The method is similar to that used in ELF OpenGo \cite{tian_elf_2019} and EfficientZero \cite{ye_mastering_2021}, in which all non-visited nodes are initialized to an estimated Q value instead of $0$ during MCTS.

To elaborate, $\hat{Q}(s)$ is introduced as an estimated Q value when $N(s,a)=0$:
\begin{equation}
\begin{aligned}
Q(s,a) &= 
\begin{cases}
Q(s,a) &  N(s,a) > 0 \\
\hat{Q}(s) &  N(s,a) = 0.
\end{cases}
\end{aligned}
\end{equation}
The value of $\hat{Q}(s)$ is determined by two statistics $N_{\Sigma}(s)$ and $Q_{\Sigma}$ from MCTS:
\begin{equation}
\begin{aligned}
N_{\Sigma}(s) &= \sum_{b} \bold{1}_{N(s,b)>0} \\
Q_{\Sigma}(s) &= \sum_{b} \bold{1}_{N(s,b)>0}Q(s,b),
\end{aligned}
\end{equation}
where $\bold{1}_{N(s,b)>0}$ is the characteristic function that filters the visited actions $b$ at $s$, $N_{\Sigma}(s)$ represents the number of children that have been visited, and $Q_{\Sigma}$ denotes the value sum of these visited children. 
We use different calculations for $\hat{Q}(s)$ in board games than Atari games. To avoid performing forced exploration (as in breadth-first search), we bias the estimated Q to a smaller value by virtually sampling the action once with a losing outcome:
\begin{equation}
\hat{Q}(s) = \frac{Q_{\Sigma}(s)}{N_{\Sigma}(s)+1}.
\end{equation}
In contrast, we encourage enhanced exploration in Atari games:
\begin{equation}
\hat{Q}(s) = 
\begin{cases}
\frac{Q_{\Sigma}(s)}{N_{\Sigma}(s)} & N_{\Sigma}(s)>0\\
1 & N_{\Sigma}(s)=0.
\end{cases} \\
\end{equation}

\subsection{Progressive Simulation for Gumbel Zero}\label{minizero:progressive_simulation}
Gumbel Zero algorithms can guarantee policy improvement and achieve comparable performance using only a few simulations.
However, more simulations can still improve performance when planning with Gumbel, especially if the lookahead depth is relatively deeper.
To take advantage of this effect, we propose a method called \textit{progressive simulation}.
We wish to weight the number of simulations so that it gradually increases during Gumbel Zero training, all while ensuring no additional computing resources are consumed.

Given a total number of iterations $I$ and the simulation budget $B$, the simulation count $n_i$ for each iteration is computed as $\frac{B}{I}$ in the unaltered Gumbel Zero training process.
Namely, the simulation count remains consistent throughout training.
With progressive simulation, we introduce two hyperparameters, $N_{min}$ and $N_{max}$.
The training uses the same simulation budget $B$ by adjusting the simulation count $n_i$ dynamically during different iterations, where $n_i\in[N_{min}, N_{max}]$.

\begin{algorithm}[t]
\caption{Progressive Simulation Budget Allocation}\label{alg:sim-budget-allocation}
\begin{algorithmic}[1]
\REQUIRE total iterations $I$
\REQUIRE simulation budget $B$
\REQUIRE simulation boundary ($N_{min}$, $N_{max}$)
\STATE $N \gets []$  \hfill$\vartriangleright$ simulation budget list for $I$ iterations
\STATE \label{alg:sim-budget-allocation-step1} $B \gets B - I \times N_{min}$  \hfill$\vartriangleright$ allocate $N_{min}$ to each iteration
\STATE \label{alg:sim-budget-allocation-step2-begin} $n \gets N_{max}$
\WHILE{$B > 0$ \AND $n > N_{min}$ \AND $I - |N| > 0$}
  \STATE $i \gets $ min($\lfloor \frac{B \times 0.5}{n - N_{min}} \rfloor$, $I - |N|$)
  \STATE insert $i$ elements of $n$ at the front of $N$
  \STATE $B \gets B - i \times (n - N_{min})$
  \STATE $n \gets \lfloor \frac{n}{2} \rfloor$
\ENDWHILE
\STATE \label{alg:sim-budget-allocation-step2-end} insert $(I-|N|)$ elements of $N_{min}$ at the front of $N$
\WHILE{$B > 0$} \label{alg:sim-budget-allocation-step3-begin}
  \STATE $k \gets$ count the number of occurrences of min($N$) in $N$
  \STATE $i \gets$ min($B$, $k$) 
  \STATE add 1 to each of $N$ from $(k-i+1)$th through $k$th
  \STATE $B\gets B - i$
\ENDWHILE \label{alg:sim-budget-allocation-step3-end}
\RETURN $N$  \hfill$\vartriangleright$ allocation results s.t. sum($N$) = $B$
\end{algorithmic}
\end{algorithm}

Algorithm \ref{alg:sim-budget-allocation} allocates the number of simulations $n_i$ for each iteration such that it satisfies $B = \Sigma n_i$ as follows. 
First, all $n_i$ are initialized to $N_{min}$ at the beginning, as in line \ref{alg:sim-budget-allocation-step1}.
Then, we progressively set a target simulation of $N_{max}$, $\frac{N_{max}}{2}$, $\frac{N_{max}}{4}$, and so on to remaining unallocated iterations, as in lines \ref{alg:sim-budget-allocation-step2-begin}-\ref{alg:sim-budget-allocation-step2-end}.
For each target simulation, up to half of the remaining budget is allocated.
Finally, when the remaining budget is not enough for a target or all $n_i$ have been allocated once, the remaining budget is allocated to $n_i$ with the lowest simulations, as in lines \ref{alg:sim-budget-allocation-step3-begin}-\ref{alg:sim-budget-allocation-step3-end}.

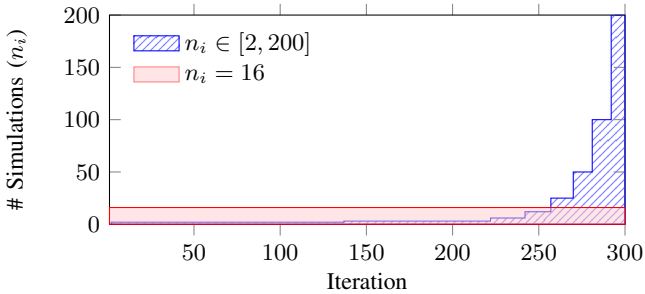
\begin{figure}[t]
\centering
\begin{tikzpicture}
\begin{axis}[
  ybar, ymin=0, ymax=200, xmin=1, xmax=300,
  ylabel={\# Simulations ($n_i$)}, xlabel={Iteration}, font=\small,
  width=\axisdefaultwidth, height=0.6*\axisdefaultheight,
  legend pos=north west, legend style={
    legend cell align=left,
    draw=none
  }
  ]
\fill[pattern=north east lines, pattern color=blue!50]
    (1, 0) -- (1, 2) -- (136, 2) -- (136, 3) -- (221, 3) -- 
    (221, 6) -- (241, 6) -- (241, 12) -- (256, 12) -- (256, 25) -- 
    (269, 25) -- (269, 50) -- (280, 50) -- (280, 100) -- 
    (291, 100) -- (291, 200) -- (299, 200) -- (299, 0) -- cycle;
\draw[blue]
    (1, 0) -- (1, 2) -- (136, 2) -- (136, 3) -- (221, 3) -- 
    (221, 6) -- (241, 6) -- (241, 12) -- (256, 12) -- (256, 25) -- 
    (269, 25) -- (269, 50) -- (280, 50) -- (280, 100) -- 
    (291, 100) -- (291, 200) -- (299, 200) -- (299, 0) -- cycle;
\addlegendimage{area legend, pattern=north east lines, pattern color=blue!50, draw=blue}
\addlegendentry{$n_i\in[2,200]$}
\fill[fill=red!20,opacity=0.5]
    (0, 0) -- (0, 16) -- (299, 16) -- (299, 0) -- cycle;
\draw[red]
    (0, 0) -- (0, 16) -- (299, 16) -- (299, 0) -- cycle;
\addlegendimage{area legend, draw=red, fill=red!20, opacity=0.5}
\addlegendentry{$n_i=16$}
\end{axis}
\end{tikzpicture}
\caption{Simulation budget allocation for progressive simulation with a setting of $I=300$, $B=4800$, and $(N_{min}, N_{max})=(2, 200)$. $n_i\in[2,200]$ depicts the simulations of each iteration for the progressive simulation; $n_i=16$ depicts those for the baseline that uses a fixed simulation of 16. Note that the baseline has the same budget as the progressive simulation.}
\label{fig:sim-allocation-example}
\end{figure}

Take $I=300$, $B=4,800$, and $(N_{min}, N_{max})=(2, 200)$ as an example.
First, $n_{1}$ to $n_{300}$ are initialized to $N_{min}=2$ simulations, with a remaining budget of 4,200 simulations. 
Next, we progressively allocate the target number of simulations, starting from 200.
For the target of 200 simulations, we need to allocate an additional $198$ simulations per iteration (two simulations are already allocated, so $200-2=198$).
Half of the remaining budget, 2,100 simulations, is split among $\lfloor \frac{2,100}{198} \rfloor = 10$ iterations.
Therefore, $n_{291}$ to $n_{300}$ are allocated an additional $198 \times 10$ simulations. 
Next, for the target of 100 simulations (with a remaining budget of $4,200 - 1,980 = 2,220$), $n_{280}$ to $n_{290}$ are allocated. 
The procedure repeats until the target of 3 simulations is allocated to $n_{179}$ to $n_{220}$. The number of simulations $n_{1}$ to $n_{178}$ remain at $N_{min} = 2$ simulations, the same as its initialized value.
Finally, the remaining 43 simulations are assigned to $n_{136}$ to $n_{178}$, resulting in the allocation as shown in Fig. \ref{fig:sim-allocation-example}.

\section{Experiments}
We evaluate the performance of four zero-knowledge learning algorithms, AlphaZero, MuZero, Gumbel AlphaZero, and Gumbel MuZero.
For simplicity, the four algorithms are denoted as $\alpha_0$, $\mu_0$, g-$\alpha_0$, and g-$\mu_0$, respectively.
This section is organized as follows.
First, Section \ref{exp:setup} introduces the training procedure and settings.
Second, sections \ref{exp:board}, \ref{exp:board_same_time}, and \ref{exp:atari} compare the performance differences when using different algorithms on the two board games and Atari games, respectively.
Third, section \ref{exp:estimated_q} conducts an ablation study to compare the performance of the estimated Q value.
Finally, section \ref{exp:psg} analyzes the proposed progressive simulation method.

\begin{table}
    \caption{Hyper-parameters for board games and Atari games}
    \centering
    \begin{tabular}{lcc}
        \toprule
        Parameter & Board Games & Atari Games\\
        \midrule
        Iteration & \multicolumn{2}{c}{300}\\
        Optimizer & \multicolumn{2}{c}{SGD}\\
        Optimizer: learning rate & \multicolumn{2}{c}{0.1}\\
        Optimizer: momentum & \multicolumn{2}{c}{0.9}\\
        Optimizer: weight decay & \multicolumn{2}{c}{0.0001}\\
        Training steps & \multicolumn{2}{c}{60K}\\
        \# Blocks & 3 & 2\\
        Batch size & 1024 & 512\\
        Replay buffer size & 40K games & 1M frames\\
        Max frames per episode & - & 108K\\
        Discount factor & - & 0.997\\
        Priority exponent ($\alpha$) & - & 1\\
        Priority correction ($\beta$) & - & 0.4\\
        Bootstrap step (n-step return) & - & 5\\
        \bottomrule
    \end{tabular}
    \label{tab:hyper-parameters}
\end{table}


\subsection{Setup}\label{exp:setup}
All experiments are conducted on one machine equipped with two Intel Xeon E5-2678 v3 CPUs, and four GTX 1080Ti GPUs.
We apply $\alpha_0$ and g-$\alpha_0$ to the two board games, 9x9 Go and 8x8 Othello, and use $\mu_0$ and g-$\mu_0$ for both board games and 57 Atari games.
Table \ref{tab:hyper-parameters} lists the hyperparameters.
Generally, we follow the same hyperparameters and network architectures as the original AlphaZero, MuZero, and Gumbel Zero paper but with some differences described as follows.

For board games, we train two models for $\alpha_0$ and $\mu_0$ with 200 MCTS simulations.
For g-${\alpha_0}$ and g-${\mu_0}$, each algorithm is trained with two models with 2 and 16 MCTS simulations.
The numbers of simulations are chosen to follow the settings proposed in \cite{danihelka_policy_2022}.
We omit training both g-${\alpha_0}$ and g-${\mu_0}$ with 200 simulations, as the experiments in \cite{danihelka_policy_2022} demonstrate that Gumbel Zero matches the performance of both AlphaZero and MuZero when training with the same number of simulations.
All six models use the same network architecture containing 3 residual blocks.
Note that in our MuZero network design, both the representation and dynamics networks contain 3 residual blocks.
During training, for each iteration, self-play workers generate 2,000 games and the optimization worker updates the network with 200 steps by using randomly sampled data from the most recent 40,000 games in the replay buffer.
For example, with a total of 300 iterations, each model is trained with a total of 600,000 self-play games and optimized for 60,000 training steps with a batch size of 1,024.

For Atari games, we train one $\mu_0$ model with 50 MCTS simulations, and two g-${\mu_0}$ models, one with 2 and the other with 18 simulations.
The choices of simulations follow the original setup in \cite{danihelka_policy_2022}.
The network block size is reduced to 2 residual blocks.
For each iteration, the self-play workers generate 250 intermediate sequences, where each contain 200 moves.
The optimization worker updates the network with 200 steps by using data randomly sampled from the most recent 1 million frames.
We calculate the n-step return by bootstrapping with five steps.
Overall, each model is trained for a total of 600,000 intermediate sequences and optimized for 60,000 training steps with a batch size of 512.

\begin{figure}[!t]
\centering
\includegraphics[width=0.8\columnwidth]{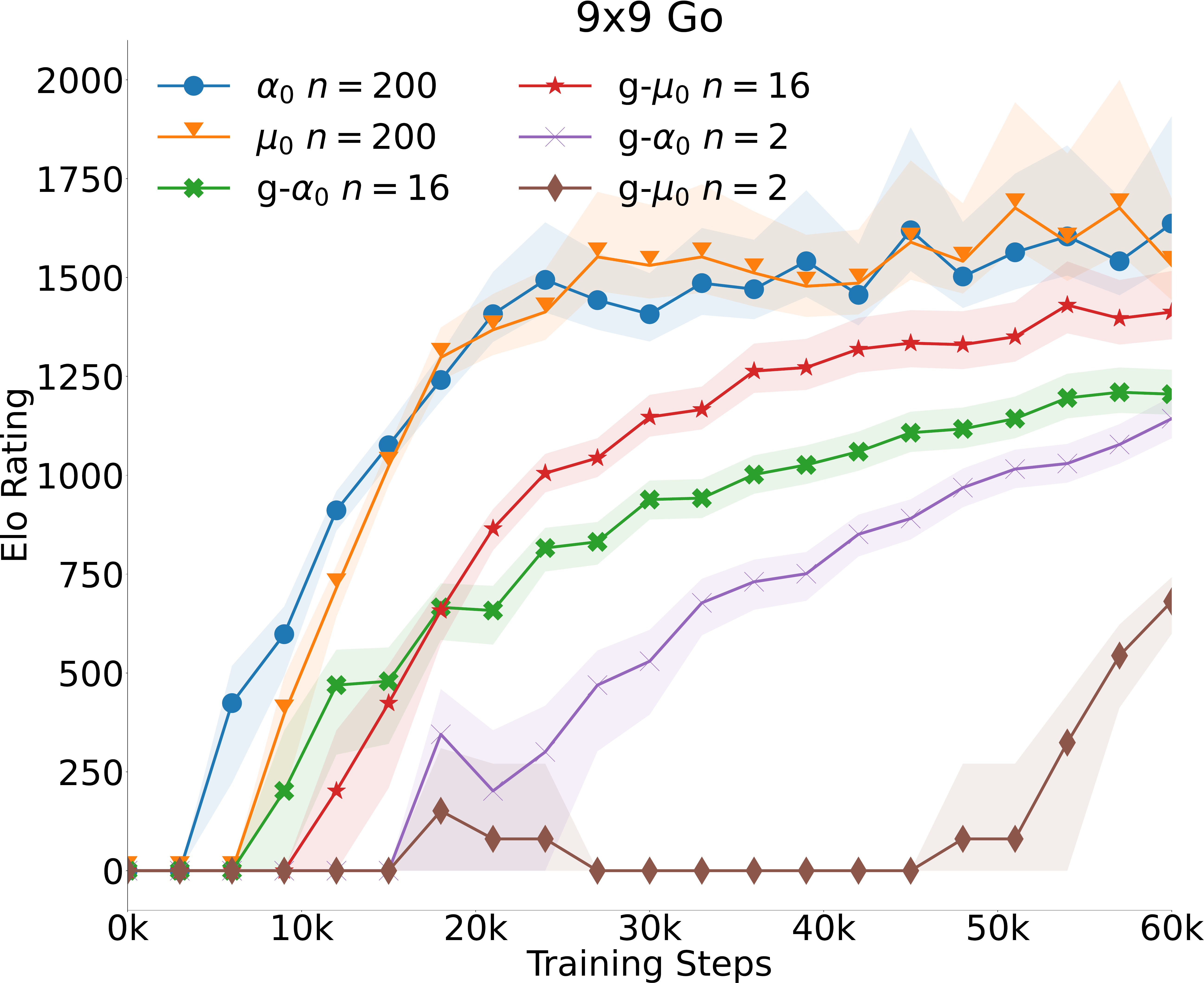}
\caption{Evaluation of different zero-knowledge learning settings in 9x9 Go. The x-axis represents the number of neural network training steps, while the y-axis represents Elo ratings. The shaded area is the error bar with 95\% confidence interval.}
\label{fig:9x9_go_evaluation}
\end{figure}

\begin{figure}[!t]
\centering
\includegraphics[width=0.8\columnwidth]{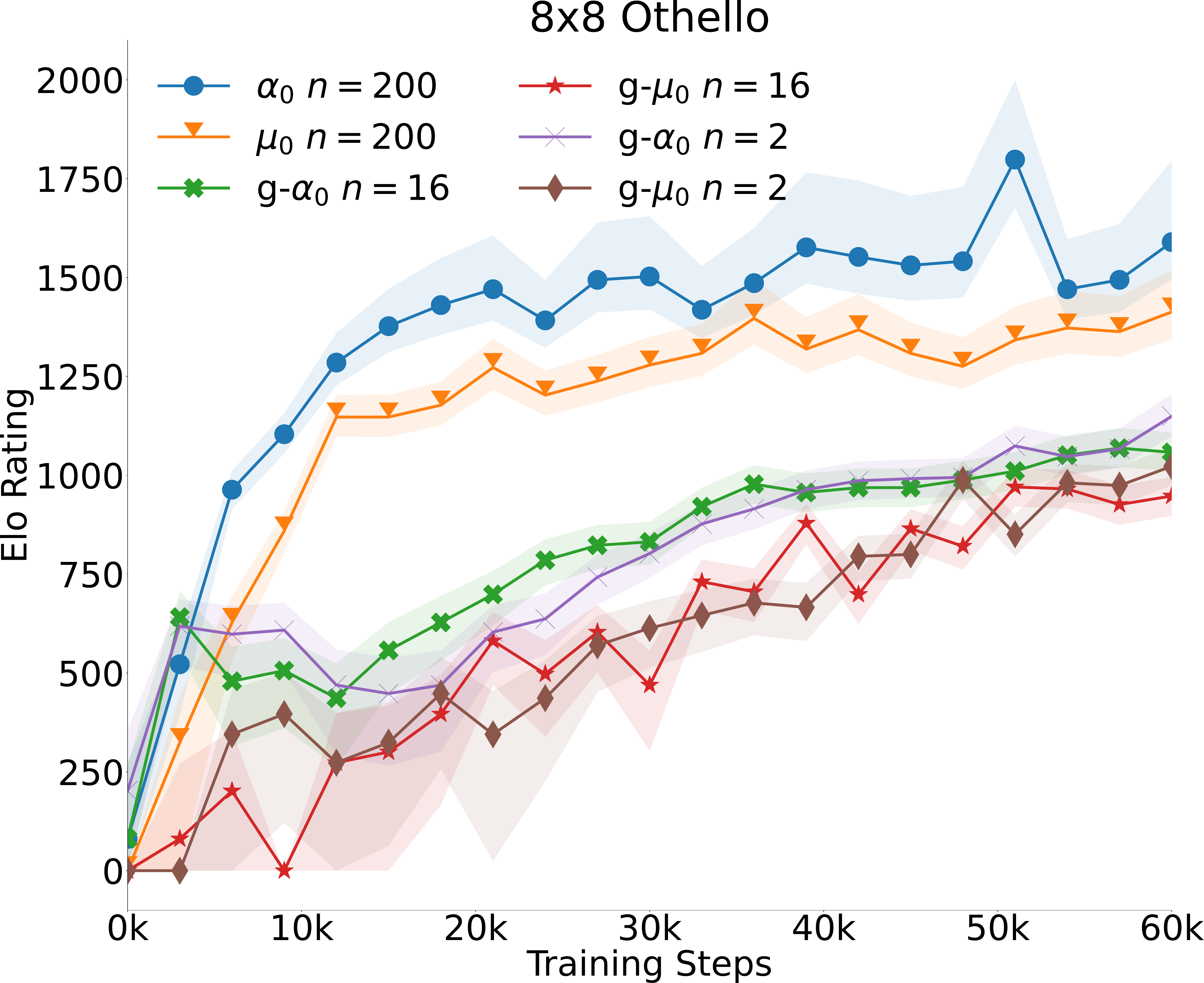}
\caption{Evaluation of different zero-knowledge learning settings in 8x8 Othello. The x-axis represents the number of neural network training steps, while the y-axis represents Elo ratings. The shaded area is the error bar with 95\% confidence interval.}
\label{fig:8x8_othello_evaluation}
\end{figure}

\subsection{Comparison of Different Algorithms on Board Games}\label{exp:board}
We compare the performance of six different zero-knowledge learning models, including $\alpha_0$ $n=200$, $\mu_0$ $n=200$, g-$\alpha_0$ $n=2$, g-$\alpha_0$ $n=16$, g-$\mu_0$ $n=2$, and g-$\mu_0$ $n=16$, on 9x9 Go and 8x8 Othello.
Each model is evaluated against a specific baseline model with Elo ratings \cite{silver_general_2018}, which is chosen to be neither excessively strong nor overly weak, for better discrimination.
Specifically, we choose $\mu_0$ $n=200$ trained for 15,000 training steps (75 iterations) for 9x9 Go and 10,000 training steps (50 iterations) for 8x8 Othello as the baselines.
Fig. \ref{fig:9x9_go_evaluation} and Fig. \ref{fig:8x8_othello_evaluation} show the strength comparison between models for the two games.
For each curve, we sample the networks every 3,000 training steps (15 iterations) and evaluate 200 games against the baseline model.
Namely, a total of 21 network checkpoints (including the initial network) are evaluated for each model.
All models use 400 simulations per turn during evaluation, regardless of the simulation count used during training.

In 9x9 Go, training with more simulations generally yields stronger results than training with fewer simulations.
For $n=200$, both $\alpha_0$ and $\mu_0$ perform similarly, surpassing the performance of the four other Gumbel Zero settings.
Next, for $n=16$, g-$\mu_0$ outperforms g-$\alpha_0$, indicating that the MuZero algorithm might be more suitable for the game of Go.
For $n=2$, the learning curve of g-$\alpha_0$ grows slowly but eventually reaches a similar performance to g-$\alpha_0$ $n=16$.
However, g-$\mu_0$ $n=2$ demonstrates the poorest performance among the six models.
The reason could be that only two simulations are not sufficient to learn the environment representation and dynamics in 9x9 Go for g-$\mu_0$.
In conclusion, for 9x9 Go, both AlphaZero and MuZero are appropriate.
However, without a sufficient number of simulations, MuZero might not perform well.

Next, in 8x8 Othello, $\alpha_0$ and $\mu_0$ still outperform g-$\alpha_0$ and g-$\mu_0$, demonstrating that using more simulations directly affects the performance.
Different from 9x9 Go, $\alpha_0$ performs better than $\mu_0$ with a significant Elo rating difference.
This phenomenon also exists in Gumbel Zero where g-$\alpha_0$ is generally better than g-$\mu_0$.
A possible explanation is that Othello has a more dramatic change on the board with each turn due to the frequent piece flipping across the board.
The dynamics network might therefore find it more difficult to learn state transitions.
Interestingly, in Gumbel Zero, we observe that the performance of $n=2$ is equal to or might even be slightly better than $n=16$ for both g-$\alpha_0$ and g-$\mu_0$.
This contradicts the results in Go, where more simulations usually yield higher Elo ratings.
This discrepancy might be attributed to the fact that Othello often has fewer legal moves relative to Go, so that fewer simulations might be sufficient in exploring optimal moves and avoiding bad ones.

In summary, given sufficient computing resources, the choice of $\alpha_0$ or $\mu_0$ with more simulations, such as $n=200$, is suitable.
However, with limited computing resources, we can consider using g-$\alpha_0$ or g-$\mu_0$ with fewer simulations.
The choice of g-$\alpha_0$ and g-$\mu_0$ with different simulation settings might depend on the property of the game.
For games with more dramatic environment changes like 8x8 Othello, g-$\alpha_0$ is a better choice than g-$\mu_0$ because it can interact with the actual environment during planning, without relying on learned environments.
In terms of the choice of simulation count in Gumbel Zero, for games that generally have many large branching factors like 9x9 Go, a higher number of simulations is essential for both g-$\alpha_0$ and g-$\mu_0$.

\subsection{Comparison of Different Algorithms on Board Games under the Same Training Time}\label{exp:board_same_time}
\begin{figure}[!t]
\centering
\includegraphics[width=0.8\columnwidth]{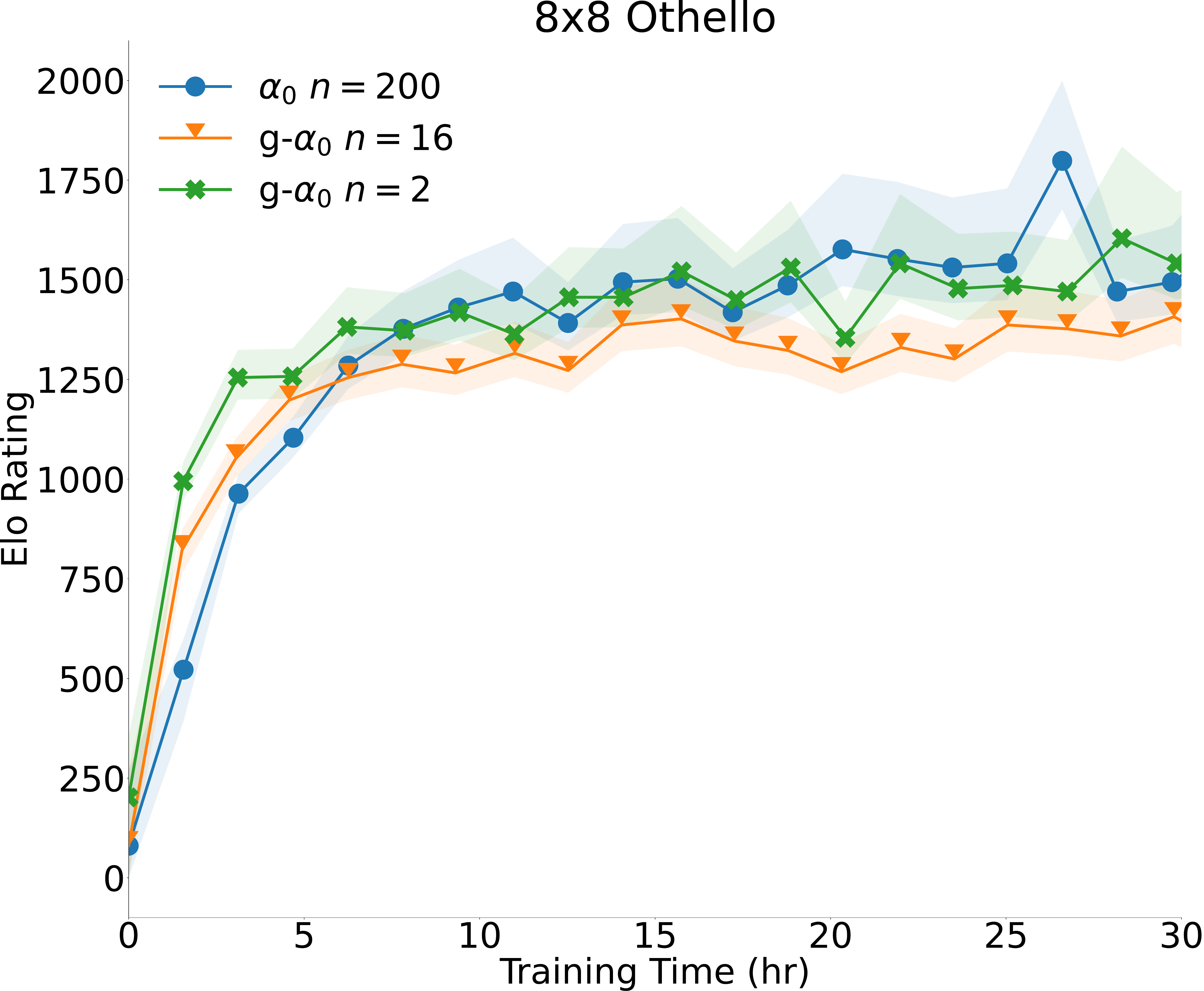}
\caption{Comparing the playing performance of $\alpha_0$ and g-$\alpha_0$ on different simulations under the same amount of training time in 8x8 Othello.}
\label{fig:8x8_othello_az_gaz_same_time}
\end{figure}

The previous subsection mainly compares the performance of algorithms with the same number of training steps (iterations).
However, it is an interesting question whether using $n=2$ and $n=16$ can attain similar performance to $n=200$ under the same training time.
To investigate this issue, we select the game of 8x8 Othello and train g-$\alpha_0$ $n=2$ and g-$\alpha_0$ $n=16$ using the same amount of time as $\alpha_0$ $n=200$.
Ordinarily, for every 200 training steps (one iteration of training), $\alpha_0$ $n=200$, g-$\alpha_0$ $n=16$, and g-$\alpha_0$ $n=2$, take around 378.51 seconds, 40.62 seconds, and 23.34 seconds.
We set the training process for all three models to roughly the same amount of time, for a total of 60,000, 560,000, and 972,000 training steps, i.e. 300, 2,800, and 4,860 iterations, respectively.
The result is shown in Fig. \ref{fig:8x8_othello_az_gaz_same_time}.
During the initial five hours of training, both g-$\alpha_0$ $n=2$ and $n=16$ perform slightly better than $\alpha_0$ $n=200$. 
After training for the same amount of time, g-$\alpha_0$ $n=2$ and $\alpha_0$ $n=200$ achieve similar playing strength.
However, g-$\alpha_0$ $n=16$ performs slightly worse than others, consistent with previous experiments described in section \ref{exp:board}.

\subsection{Comparison of Different Algorithms on Atari Games}\label{exp:atari}
For Atari games, we use a similar evaluation approach as Muesli \cite{hessel_muesli_2021}, where the average returns are calculated based on the last 100 finished training episodes from self-play games.
Table \ref{tab:Atari57-score} lists the average returns of 57 Atari games, while Fig. \ref{fig:Atari57-score} depicts the learning curves for each game.
Among 57 Atari games, we observed that $\mu_0$ $n=50$, g-$\mu_0$ $n=18$, and g-$\mu_0$ $n=2$ perform the best in 40, 10, and 13 games, respectively.
Generally, $\mu_0$ $n=50$ consistently outperforms g-$\mu_0$ $n=18$, and g-$\mu_0$ $n=18$ outperforms g-$\mu_0$ $n=2$.

There are three cases worth discussing the performance of g-$\mu_0$ versus $\mu_0$.
First, we find that $\mu_0$ $n=50$ achieves noticeably higher average returns than g-$\mu_0$ $n=18$ and g-$\mu_0$ $n=2$ in some games\footnote{\textit{breakout}, \textit{centipede}, \textit{crazy\_climber}, \textit{enduro}, \textit{fishing\_derby}, \textit{kangaroo}, \textit{kung\_fu\_master}, \textit{private\_eye}, \textit{up\_n\_down}.}.
For example, in \textit{centipede}, the average returns of $\mu_0$ $n=50$ achieves 49,823.07, while the g-$\mu_0$ $n=18$ and g-$\mu_0$ $n=2$ only reach 30,749.87 and 9,946.83, respectively.
We suspect that these games require meticulous planning and precise action sequences.
Therefore, having more simulations can be critical for achieving higher returns.
In g-$\mu_0$ $n=2$, only two actions are evaluated, which makes exploring better actions more challenging without deeper planning.

\begin{figure*}[!t]
\centering
\subfloat{
    \includegraphics[width=0.32\textwidth]{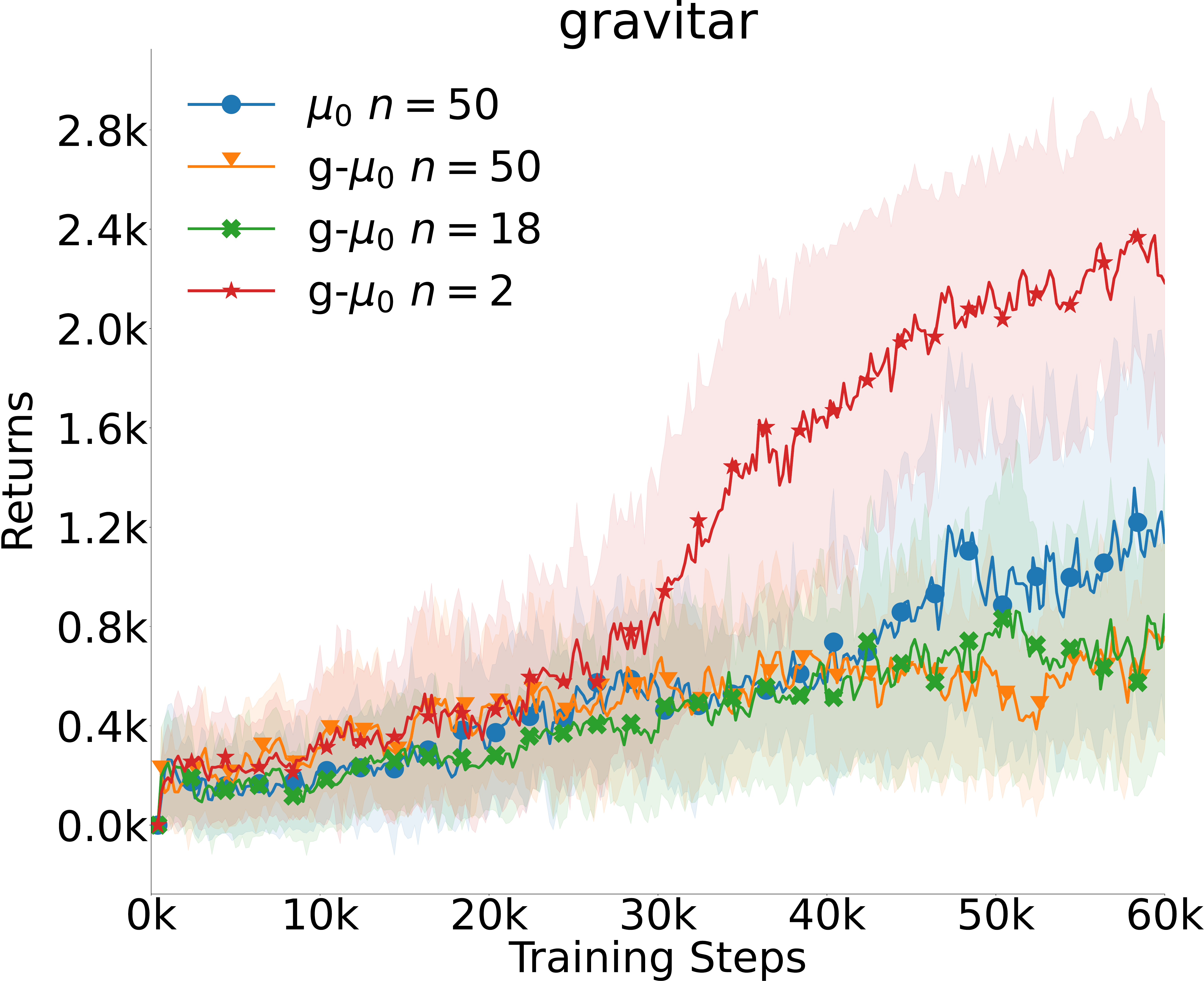}}
\subfloat{
    \includegraphics[width=0.32\textwidth]{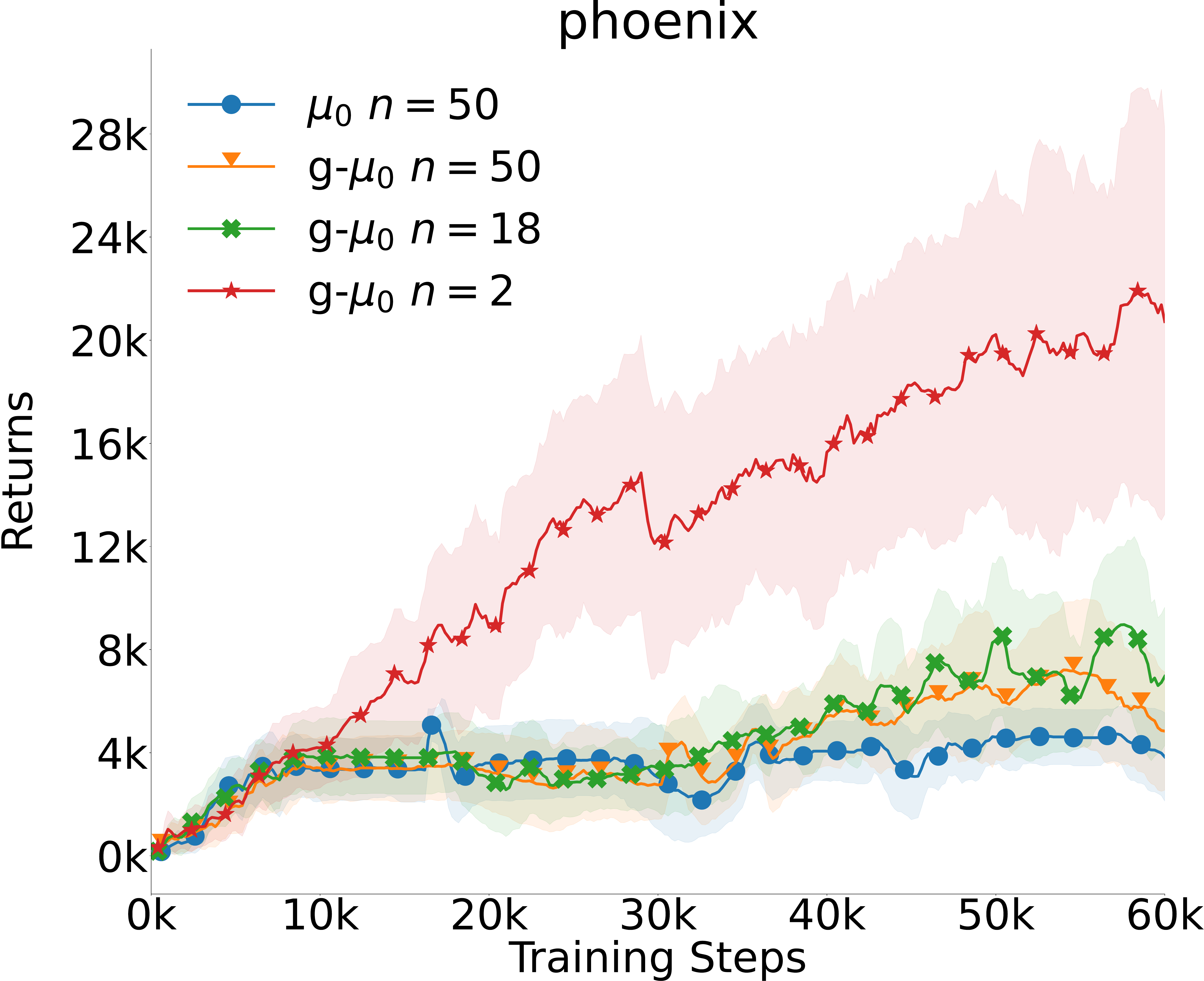}}
\subfloat{
    \includegraphics[width=0.32\textwidth]{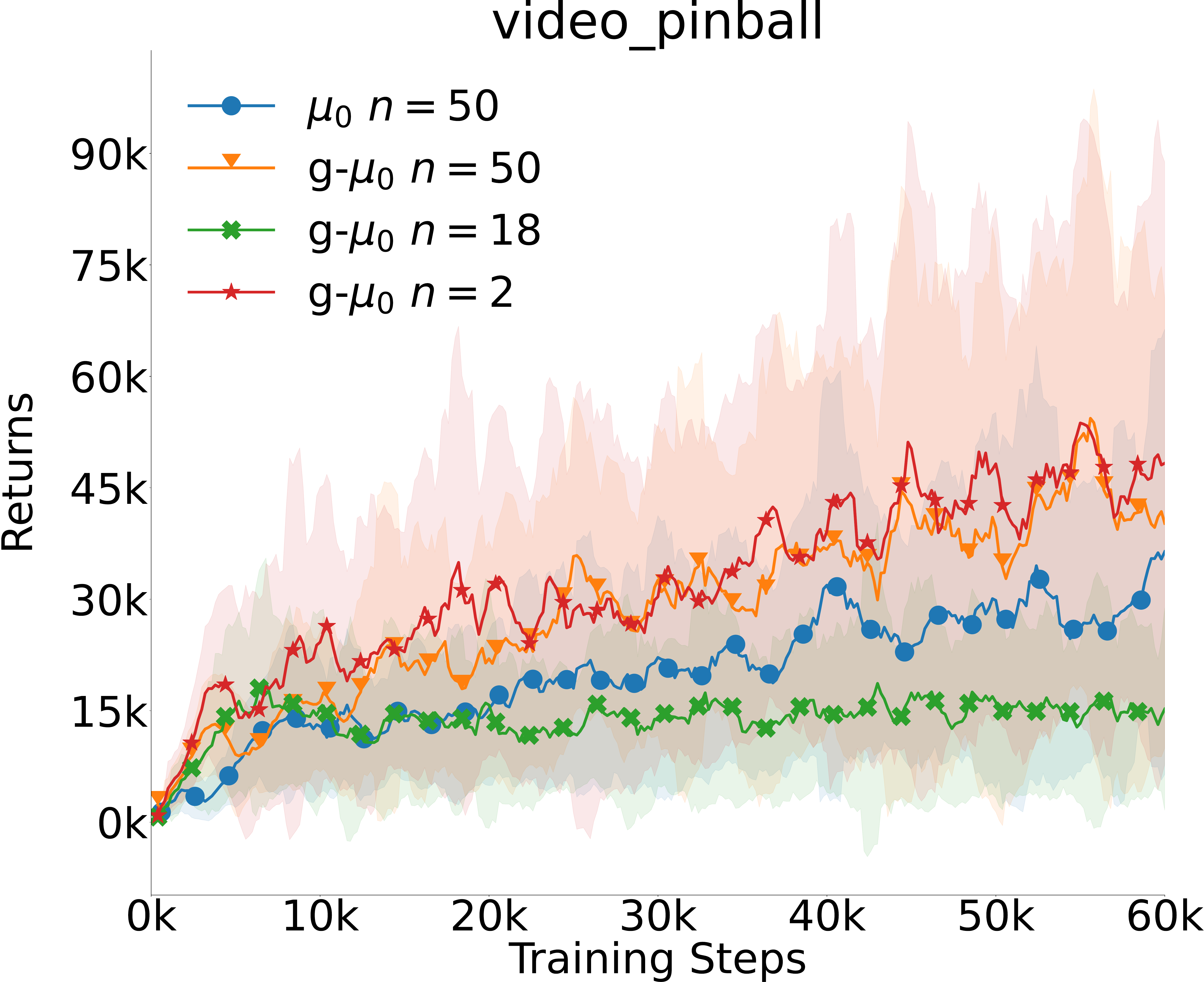}}
\caption{The g-${\mu_0}$ $n=50$ experiment on three Atari games, including \textit{gravitar}, \textit{phoenix}, and \textit{video\_pinball}.}
\label{fig:atari-gmz-n50-compare}
\end{figure*}

\begin{figure}[!t]
\centering
\subfloat[$\mu_0\ n=50$]{
    \includegraphics[width=0.45\columnwidth]{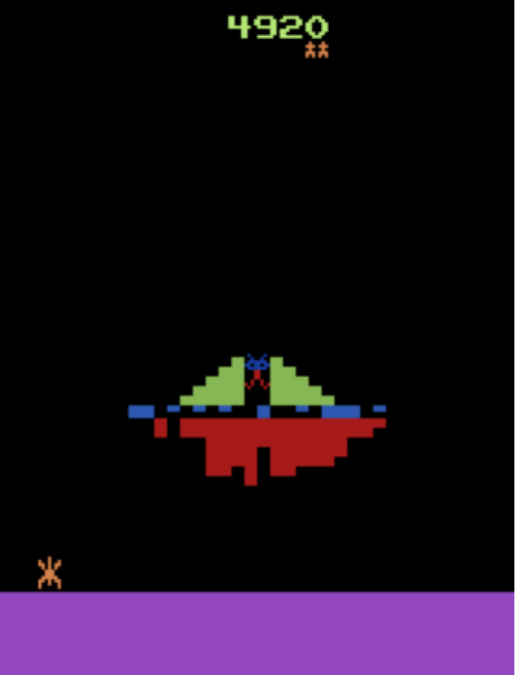}
    \label{fig:phoenix-mz-n50}
    }
\subfloat[g-$\mu_0\ n=2$]{
    \includegraphics[width=0.45\columnwidth]{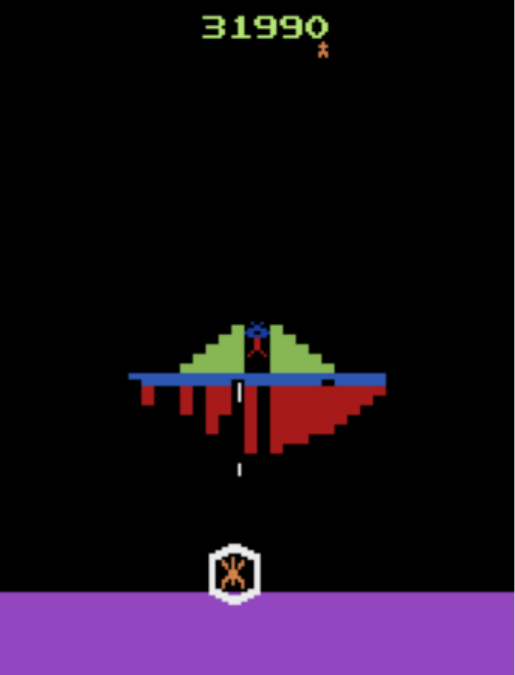}
    \label{fig:phoenix-gmz-n2}
    }
\label{fig:phoenix}
\caption{A screenshot from the final iteration of self-play games in \textit{phoenix}: (a) $\mu_0$ $n=50$ tends to stay at the leftmost position to avoid fire from the boss (alien pilot), and (b) g-$\mu_0$ $n=2$ tries to move to the center and target the boss.}
\end{figure}

Second, we discuss the case for g-$\mu_0$ outperforming $\mu_0$.
Specifically, for the three games, \textit{gravitar}, \textit{phoenix}, and \textit{video\_pinball}, g-$\mu_0$ $n=2$ outperformed significantly better than $\mu_0$ $n=50$.
To investigate which of the two settings, the Gumbel trick or the number of simulations, contribute more to this phenomenon, we additionally trained g-$\mu_0$ $n=50$ for these three games, as shown in Fig. \ref{fig:atari-gmz-n50-compare}.
For the two games, \textit{gravitar} and \textit{phoenix}, performance does not improve even with 50 simulations, regardless of whether the Gumbel trick is used.
We observe that \textit{gravitar}\footnote{\textit{gravitar} is a game that controls a spaceship with various missions on different planets. The mission complexity will become progressively more difficult and the environment will change significantly when the mission is completed.} has highly unpredictable environment changes that are hard to learn, making it difficult to learn an accurate dynamics network.
As the tree digs deeper, the simulated environment diverges more from the real environment, leading to increasingly inaccurate policy and value estimates.
Next, we investigate the game of \textit{phoenix}, which is a shooting game.
We notice that the boss (an alien pilot) constantly fires bullets towards the center of the game, making it more challenging for the player to survive when close to the boss.
Surprisingly, $\mu_0$ $n=50$ tends to stay at the leftmost place, which is outside the firing range of the boss, to avoid death, as shown in Fig. \ref{fig:phoenix-mz-n50}.
With deep planning using 50 simulations, $\mu_0$ $n=50$ sufficiently explores to understand that approaching the boss too closely results in death easily.
This results in a longer survival time, but the returns do not increase without beating the boss, leading to lower average returns.
In contrast, due to a smaller number of two simulations and the Gumbel noise, g-$\mu_0$ $n=2$ inevitably chooses actions that bring it closer to the boss, as shown in Fig. \ref{fig:phoenix-gmz-n2}.
Although this results in a shorter survival time for g-$\mu_0$ $n=2$, it allows the agent to explore another strategy to defeat the boss and obtain higher returns.
For the game of \textit{video\_pinball}, g-$\mu_0$ $n=50$ has nearly the same performance as g-$\mu_0$ $n=2$, suggesting that the Gumbel trick has a positive impact on its performance.

Finally, although g-$\mu_0$ uses fewer simulations than $\mu_0$, there are some games where g-$\mu_0$ $n=2$ and g-$\mu_0$ $n=18$ achieve similar returns to those of $\mu_0$ $n=50$.
For example, in \textit{boxing} and \textit{qbert}, the game environment does not have extensive changes and offers straightforward scoring opportunities.
Hence, the agent can learn the environment transitions easily while also not requiring intricate planning.
Consequently, using the Gumbel approach allows g-$\mu_0$ to attain almost identical training curves as $\mu_0$.

It is also worth mentioning that several games\footnote{\textit{boxing}, \textit{freeway}, \textit{hero}, \textit{pong}, \textit{private\_eye}, \textit{enduro}, and \textit{solaris}.} achieve nearly the same performance as the original MuZero paper even if the models are trained with a smaller network architecture (2 blocks compared to 16 blocks) and fewer environment frames (15 million frames compared to 20 billion frames).

Therefore, these games might serve as exemplary benchmarks to verify the zero-knowledge learning algorithms without requiring a large amount of computational resources.

In summary, our experiments show that the performance of $\mu_0$ and g-$\mu_0$ varies across different Atari games.
This indicates that each Atari game has unique characteristics, and using different zero-knowledge learning algorithms and simulations can yield varying results. 
We present three cases to illustrate that the choice between g-$\mu_0$ and $\mu_0$, and the number of simulations might depend on the complexity of the game, the predictability of the environment, and the benefits of planning for the game.
These results contribute to a deeper understanding of zero-knowledge learning algorithms. We expect that our findings will provide valuable insights for future research studying Atari games with these zero-knowledge learning algorithms.

\begin{figure}[!t]
\centering
\subfloat{
    \includegraphics[width=0.48\columnwidth]{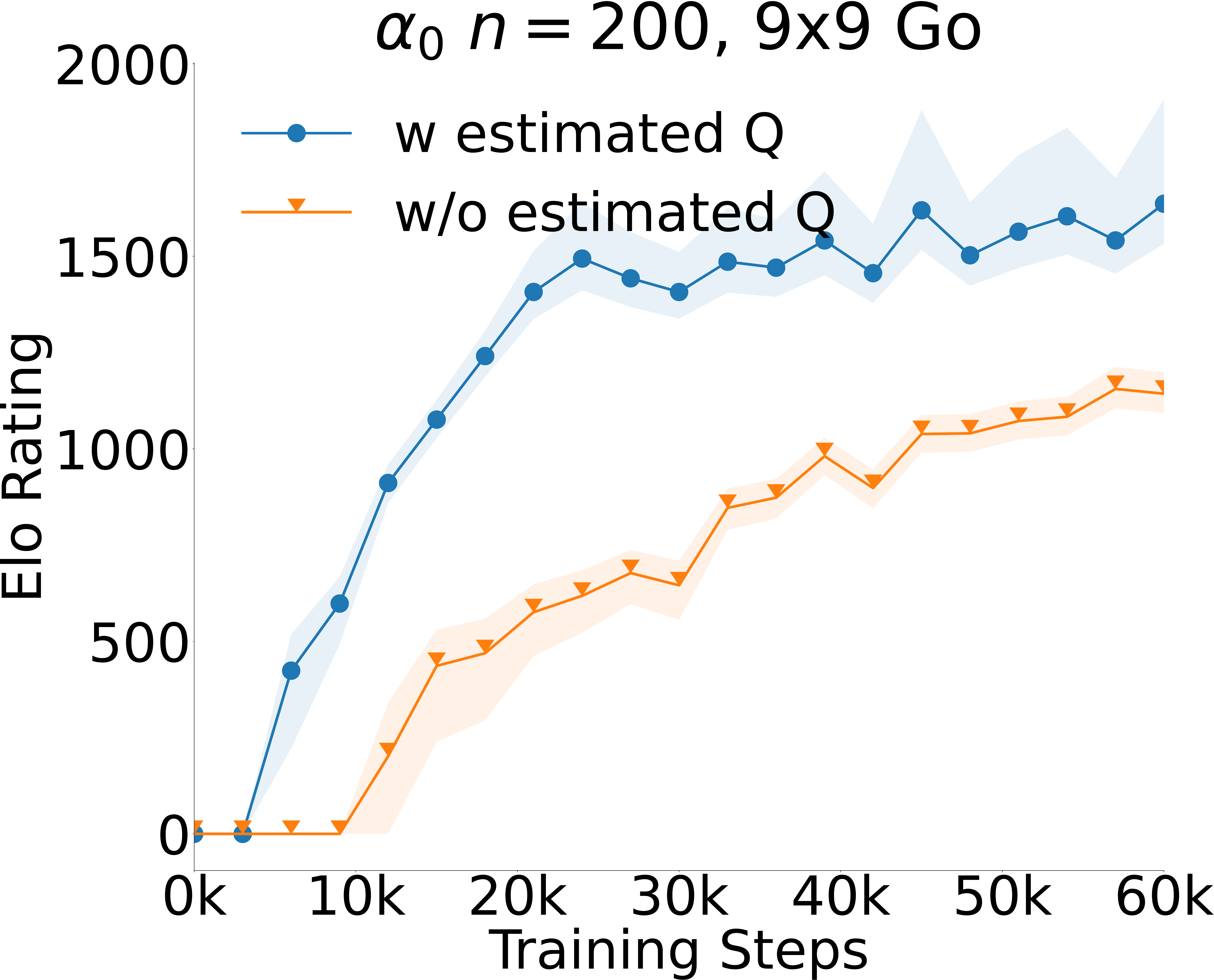}}
\subfloat{
    \includegraphics[width=0.48\columnwidth]{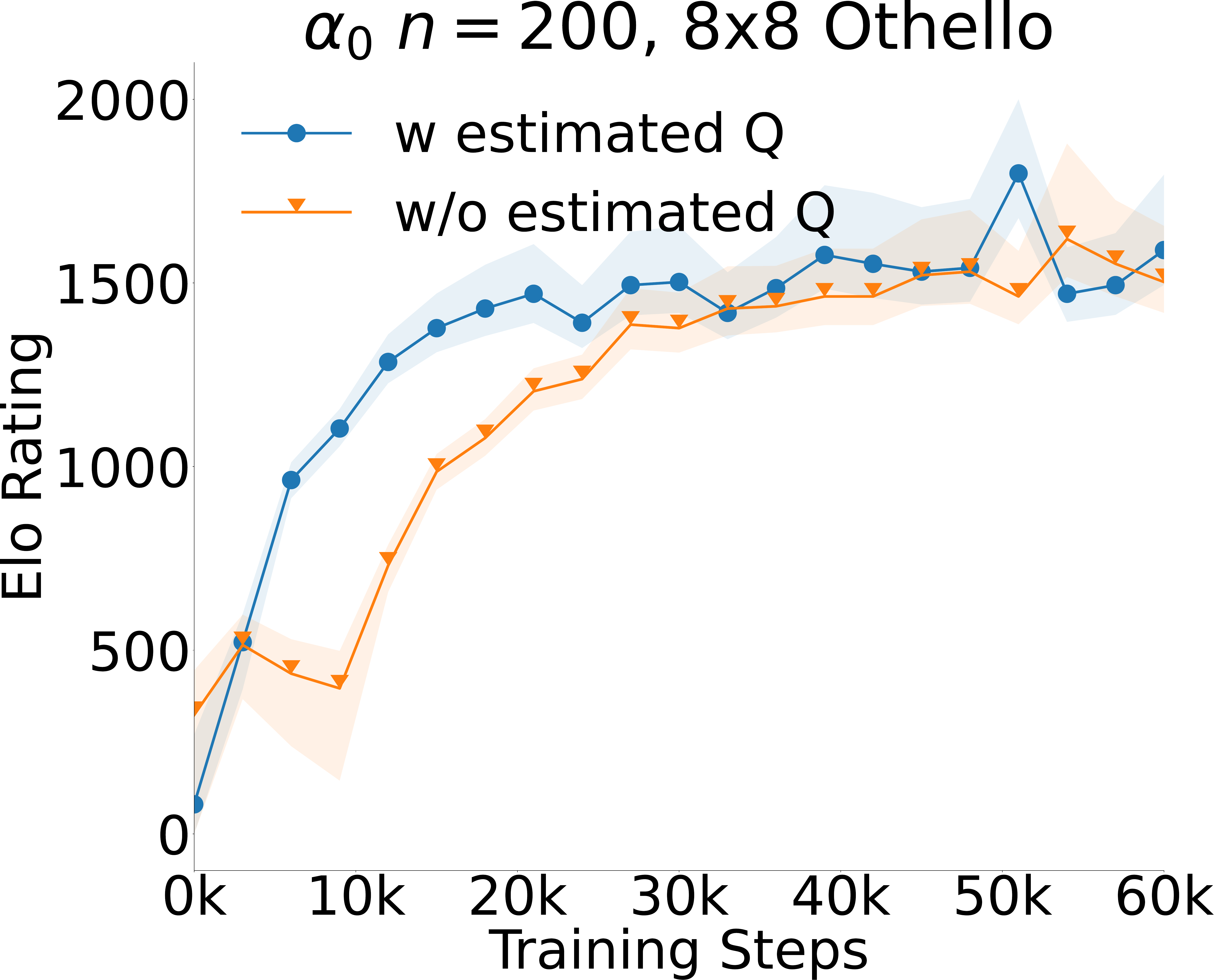}}
\caption{Ablation study of whether using estimated Q value for non-visited actions for $\alpha_0$ $n=200$ on board games.}
\label{fig:estimated_q_board_games}
\end{figure}

\begin{figure}[!t]
\centering
\subfloat{
    \includegraphics[width=0.48\columnwidth]{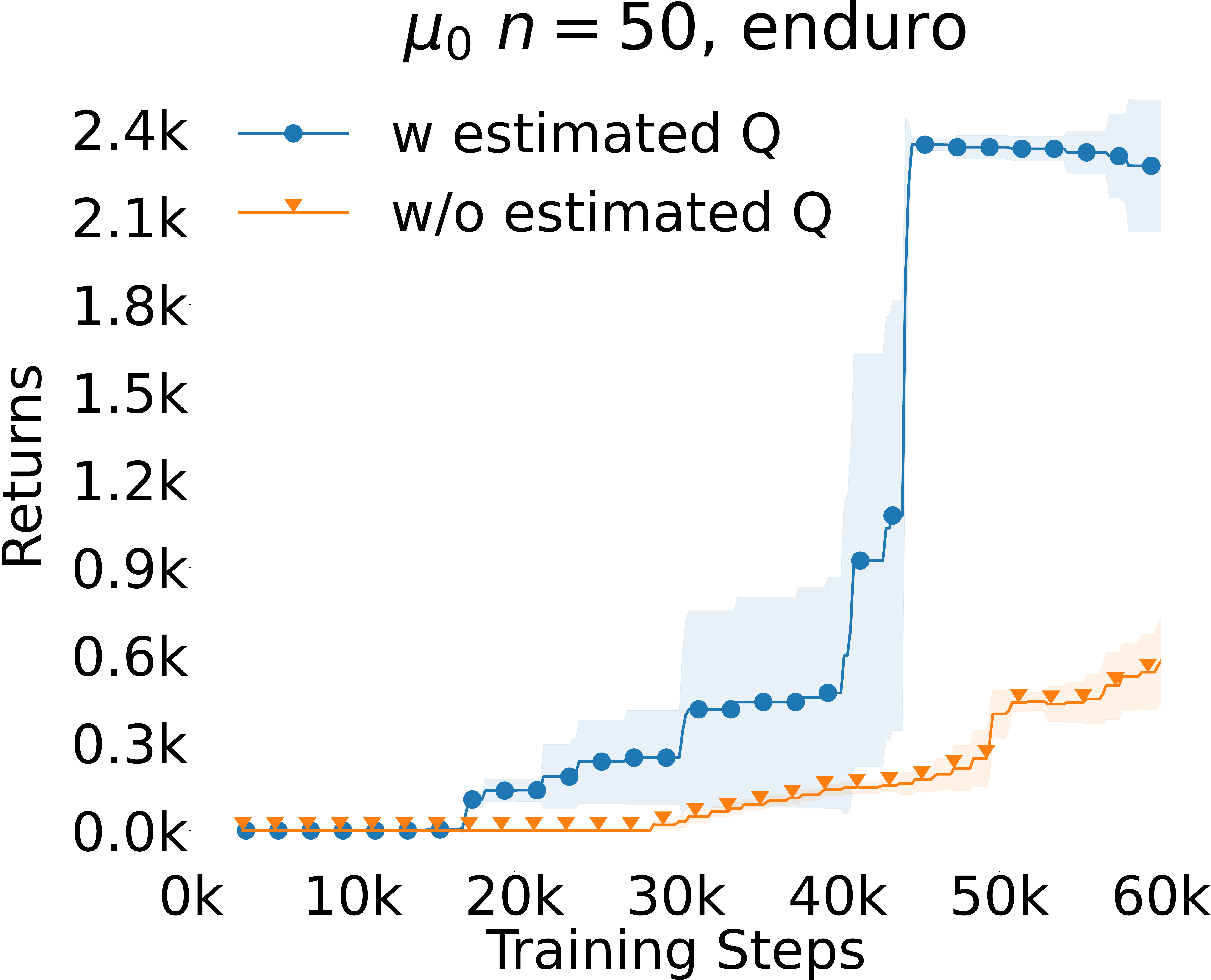}}
\subfloat{
    \includegraphics[width=0.48\columnwidth]{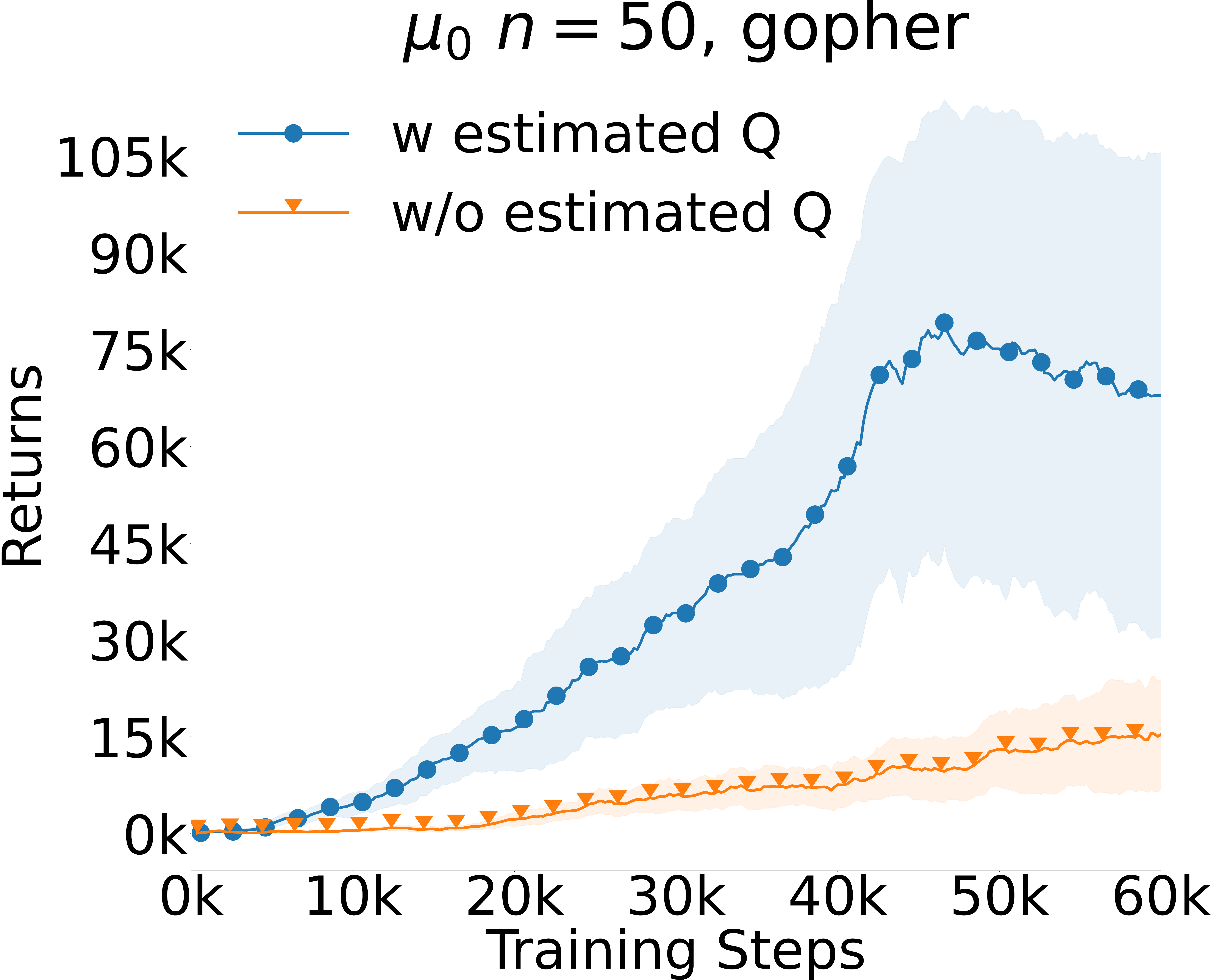}}
\\
\vspace{-0.5em}
\subfloat{
    \includegraphics[width=0.48\columnwidth]{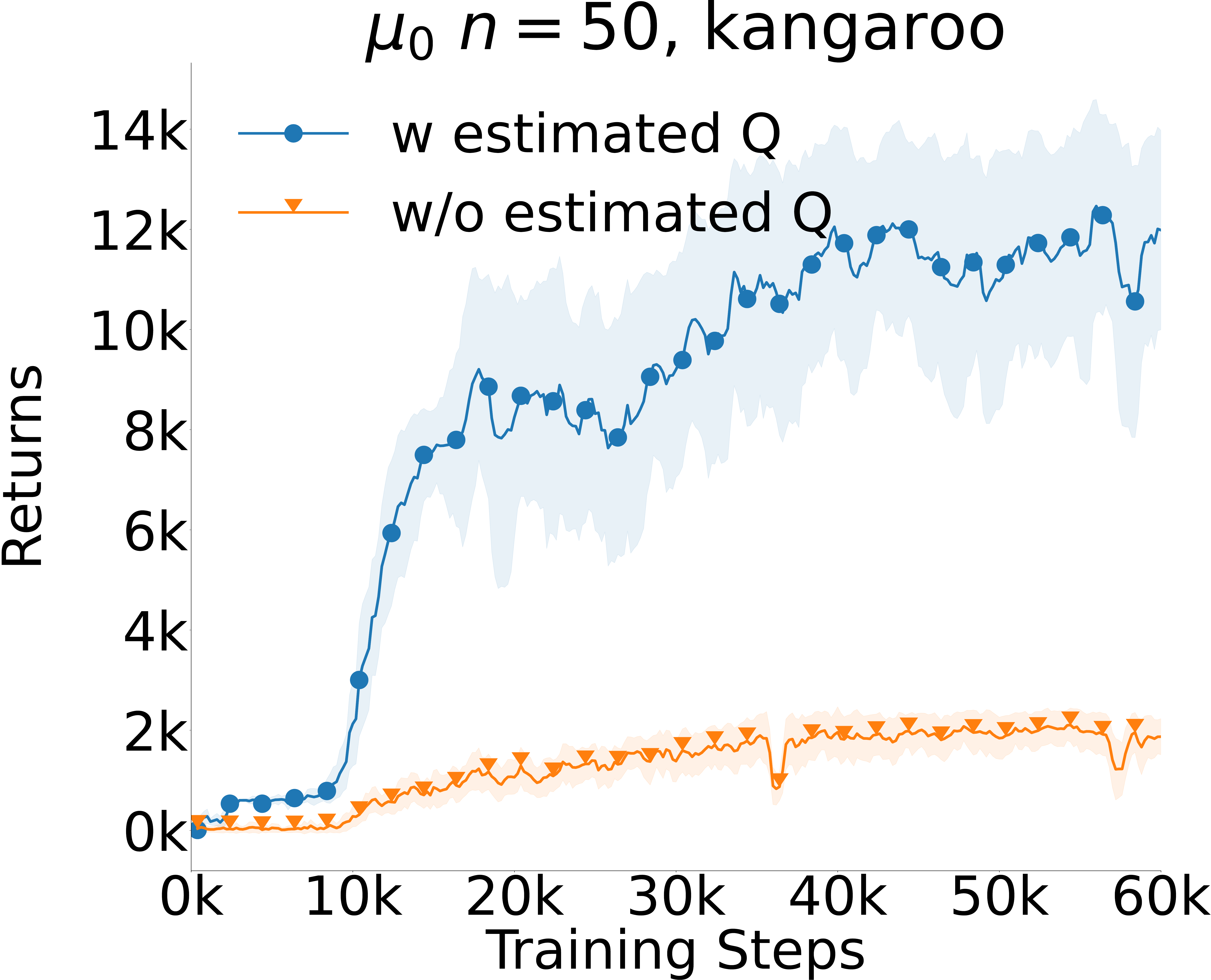}}
\subfloat{
    \includegraphics[width=0.48\columnwidth]{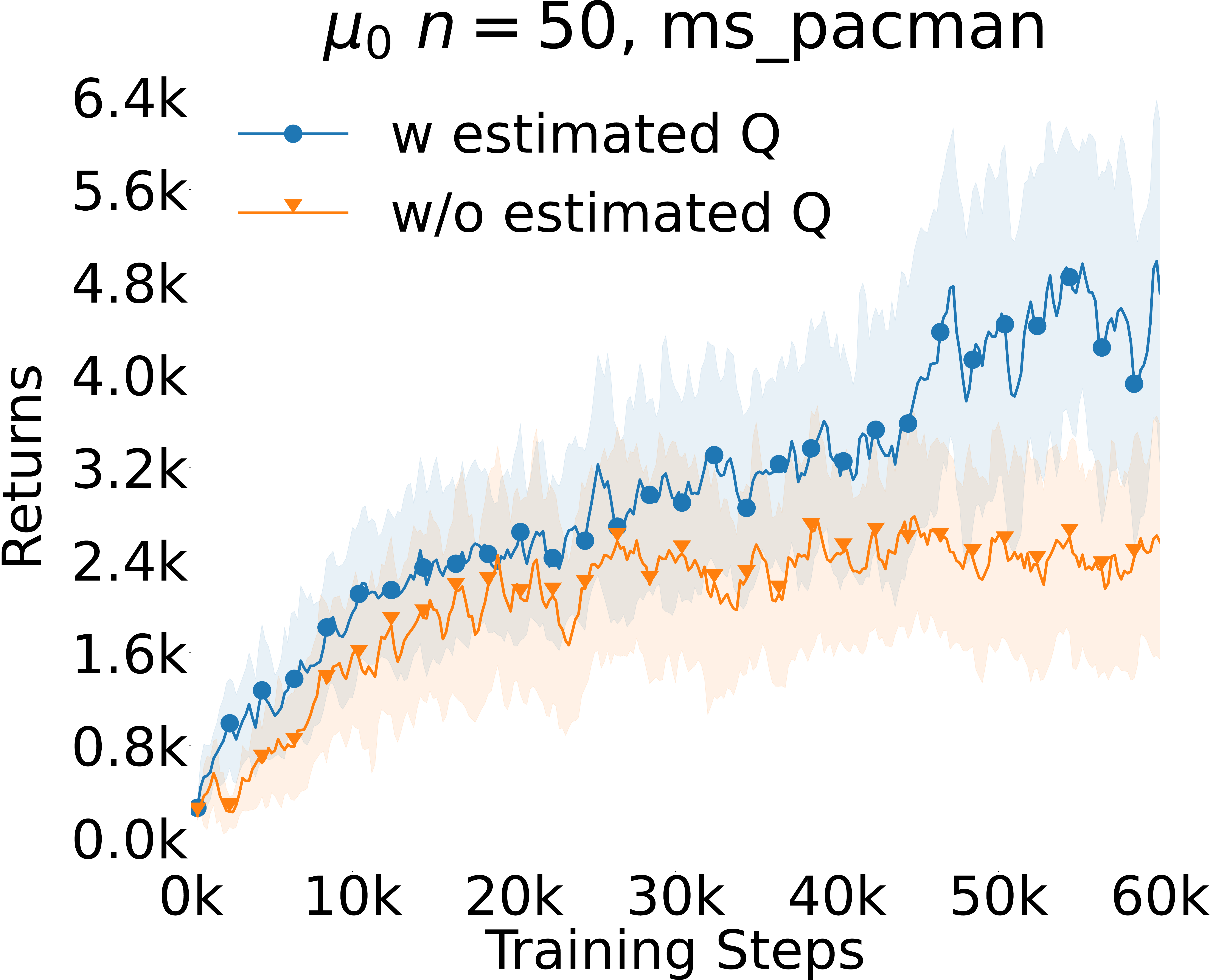}}
\caption{Ablation study of whether using estimated Q value for non-visited actions for $\mu_0$ $n=50$ on four Atari games: \textit{enduro}, \textit{gopher}, \textit{kangaroo}, and \textit{ms\_pacman}.}
\label{fig:estimated_q_atari_games}
\end{figure}

\subsection{Ablation Study on Estimated Q value}\label{exp:estimated_q}
We conduct an ablation study to determine the impact of using the estimated Q value in MCTS, as described in section \ref{minizero:estimated_q}.
Specifically, we compare the performance using estimated Q value with $\alpha_0$ for 9x9 Go and 8x8 Othello, and with $\mu_0$ for four Atari games: \textit{enduro}, \textit{gopher}, \textit{kangaroo}, and \textit{ms\_pacman}.
The training settings follow those introduced above.

Fig. \ref{fig:estimated_q_board_games} and Fig. \ref{fig:estimated_q_atari_games} illustrate the training results in either Elo ratings for board games or average returns for Atari games.
As shown in these figures, using the estimated Q value significantly outperforms the method without the estimated Q value, with the exception of 8x8 Othello.
Furthermore, the learning curves using the estimated Q value consistently outperform those without throughout the training.
This indicates that for planning problems with sufficiently large action spaces, more exploration can be more beneficial.

\begin{figure}[!t]
\centering
\includegraphics[width=0.8\columnwidth]{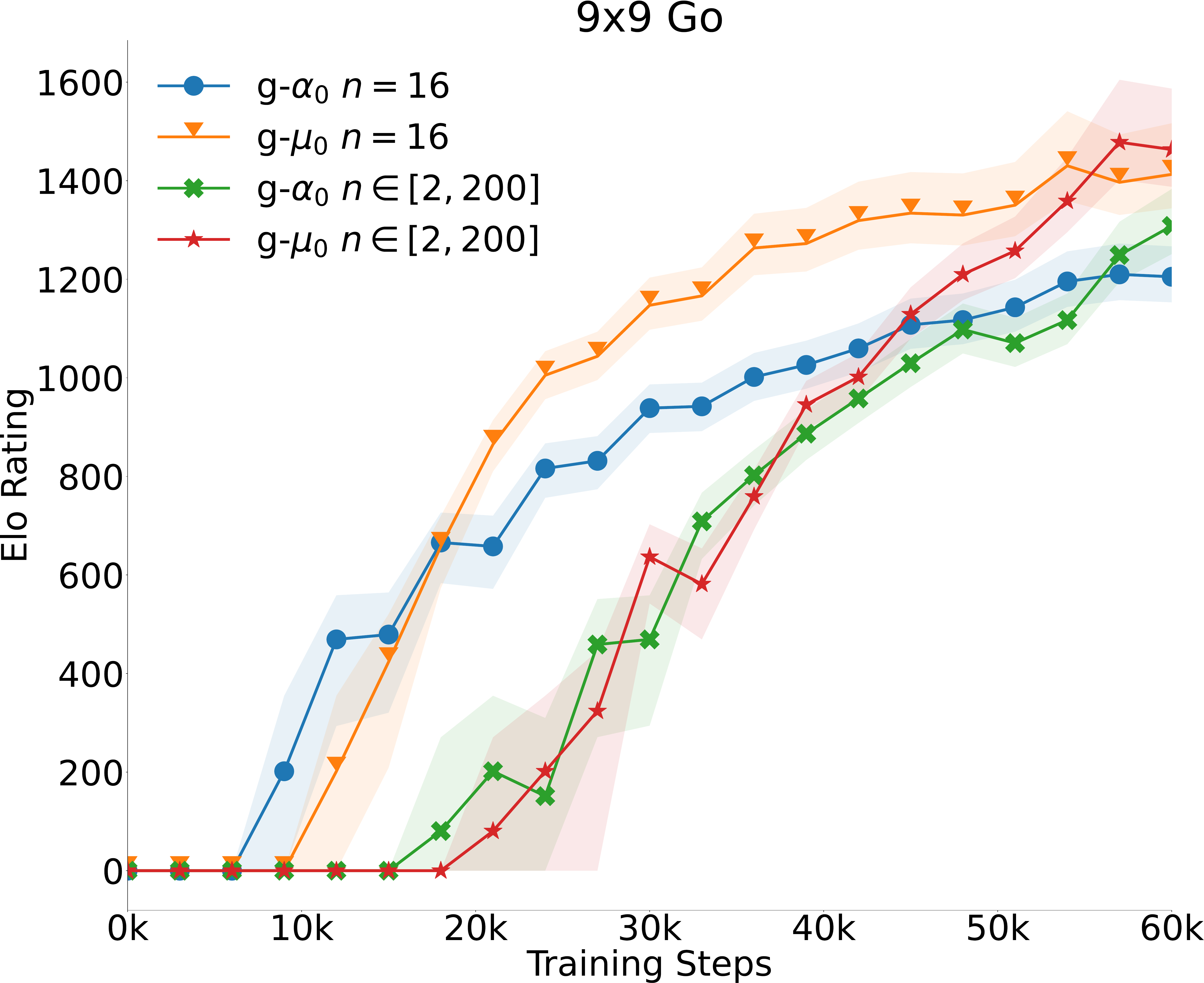}
\caption{Progressive simulation for Gumbel Zero in 9x9 Go games.}
\label{fig:9x9_go_scheduling}
\end{figure}

\begin{figure}[!t]
\centering
\includegraphics[width=0.8\columnwidth]{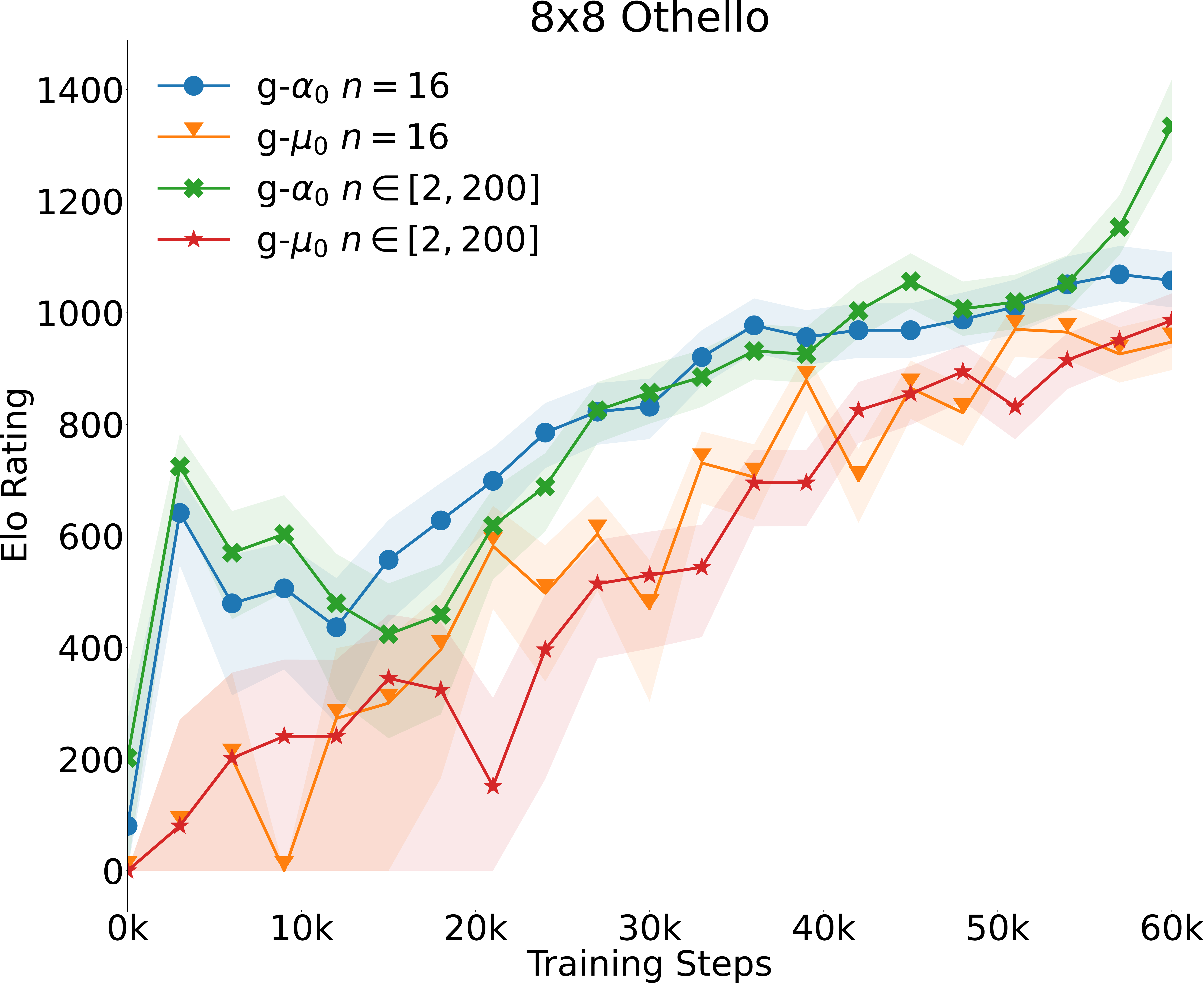}
\caption{Progressive simulation for Gumbel Zero in 8x8 Othello games.}
\label{fig:8x8_othello_scheduling}
\end{figure}

\subsection{Progressive Simulation in Gumbel Zero}\label{exp:psg}
We examine the performance of Gumbel Zero trained with \textit{progressive simulation}, as introduced in section \ref{minizero:progressive_simulation}.
For board games, we train g-$\alpha_0$ and g-$\mu_0$ with the simulation boundary $(N_{min}, N_{max})=(2, 200)$, where the simulation budget is the same as $n=16$.
Note that under this setting, the simulation budget allocation of each iteration is the same as in Fig. \ref{fig:sim-allocation-example}.
The evaluation results for both 9x9 Go and 8x8 Othello are shown in Fig. \ref{fig:9x9_go_scheduling} and Fig. \ref{fig:8x8_othello_scheduling}.
In general, the models trained with progressive simulation outperform those trained without by the end of training.
However, we observe that both g-$\alpha_0$ and g-$\mu_0$ with progressive simulation are weaker than those without before 54,000 training steps (270 iterations) for both games.
Most notably, in the case of 8x8 Othello, g-$\alpha_0$ $n\in [2, 200]$ jumps significantly in Elo rating -- by 276 -- between 50,000 to 60,000 training steps.
More specifically, the simulation count is increased to over $n=16$ after 51,000 training steps, as shown in Fig. \ref{fig:sim-allocation-example}.

These observations corroborate existing findings on combining machine learning with planning.
It has been shown that when evaluations -- both value and policy -- are inaccurate, planning can actually lead to worse decisions \cite{haque_road_2022}. 
In the early training stages of AlphaZero and MuZero, the agent's evaluations are certainly very inaccurate. 
The progressive simulation method addresses this by conserving the simulation budget for later stages of training, where the agent may benefit the most from planning with much more accurate evaluations.

\begin{figure*}[h!t]
\centering
\subfloat{
    \includegraphics[width=0.15\textwidth]{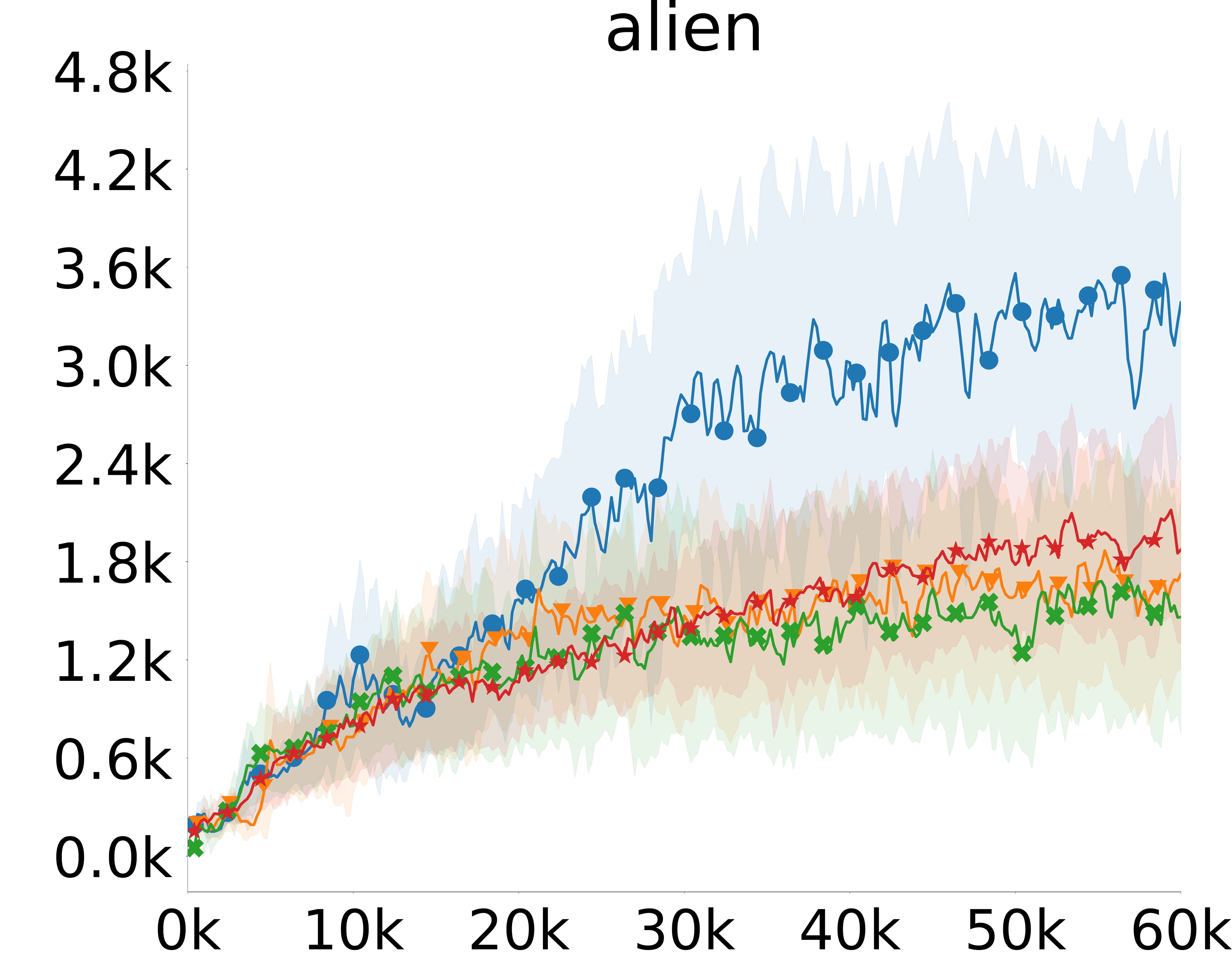}}
\subfloat{
    \includegraphics[width=0.15\textwidth]{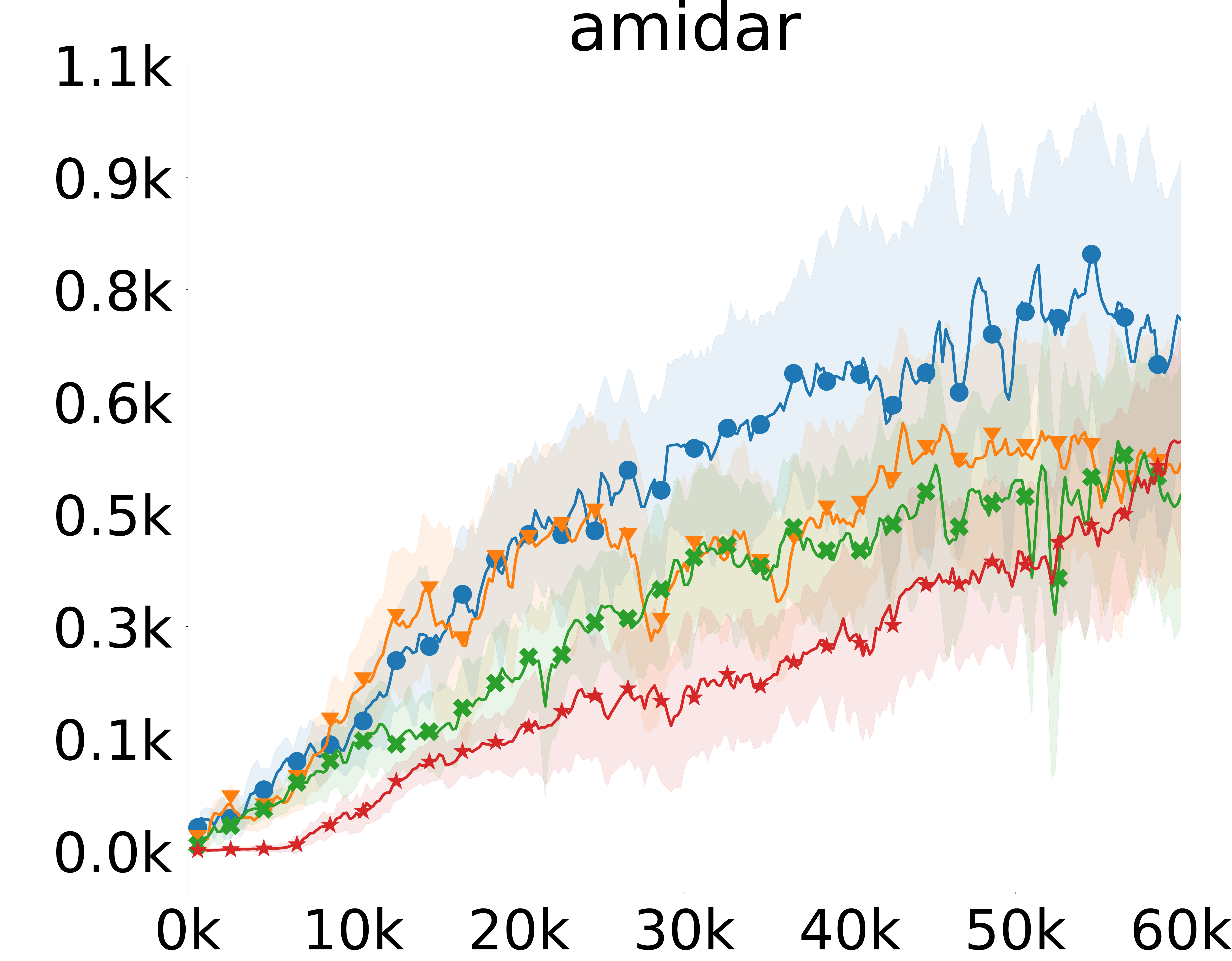}}
\subfloat{
    \includegraphics[width=0.15\textwidth]{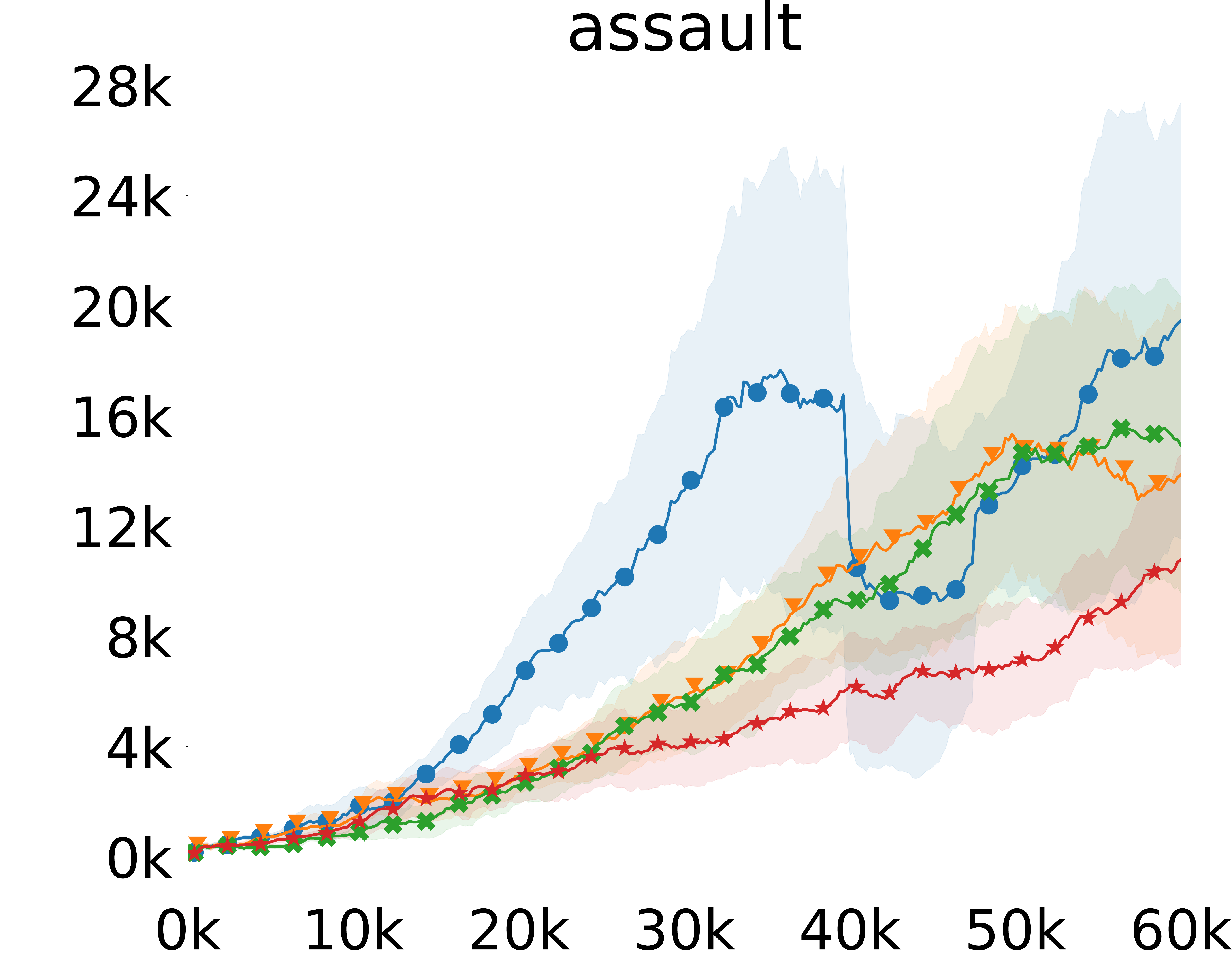}}
\subfloat{
    \includegraphics[width=0.15\textwidth]{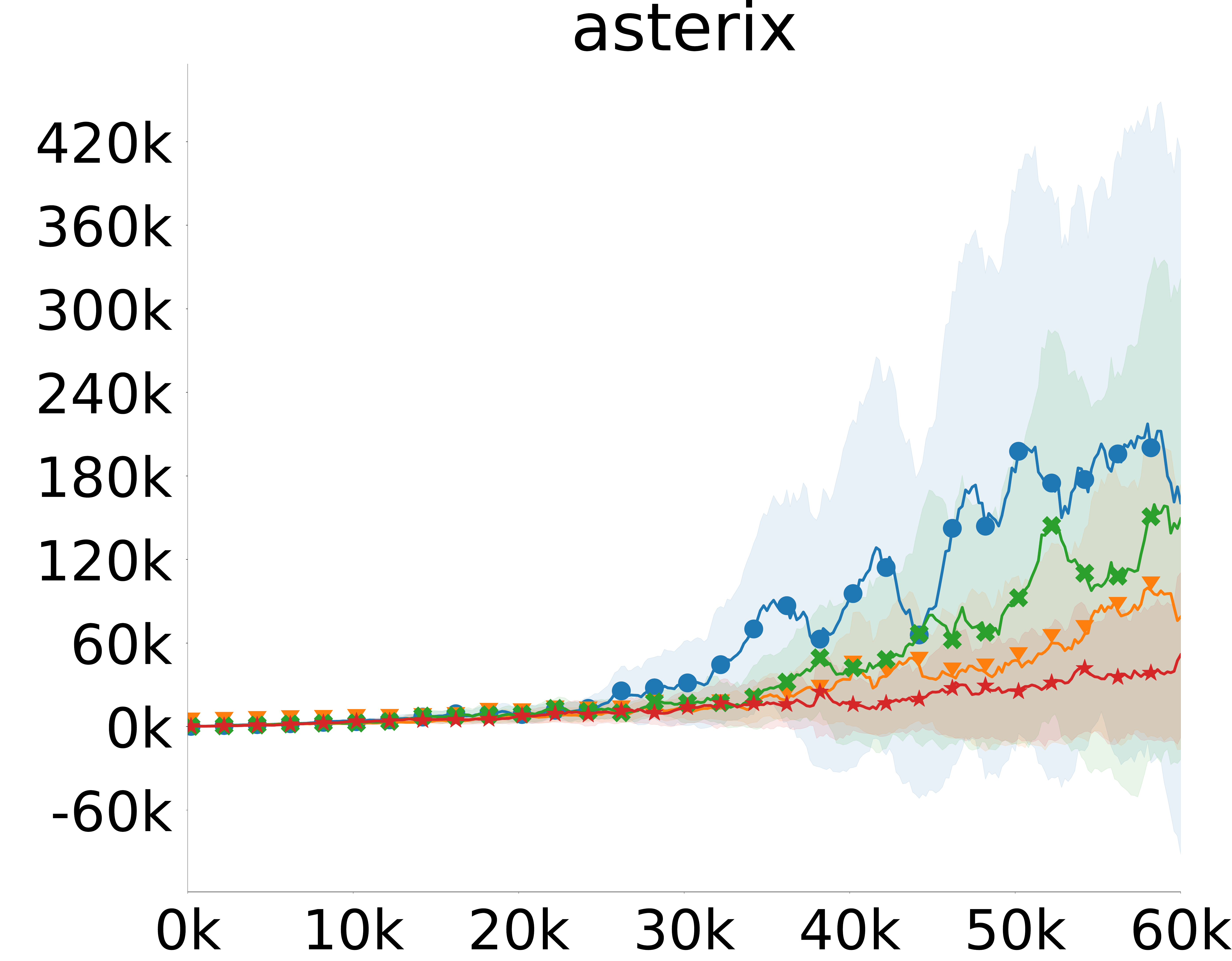}}
\subfloat{
    \includegraphics[width=0.15\textwidth]{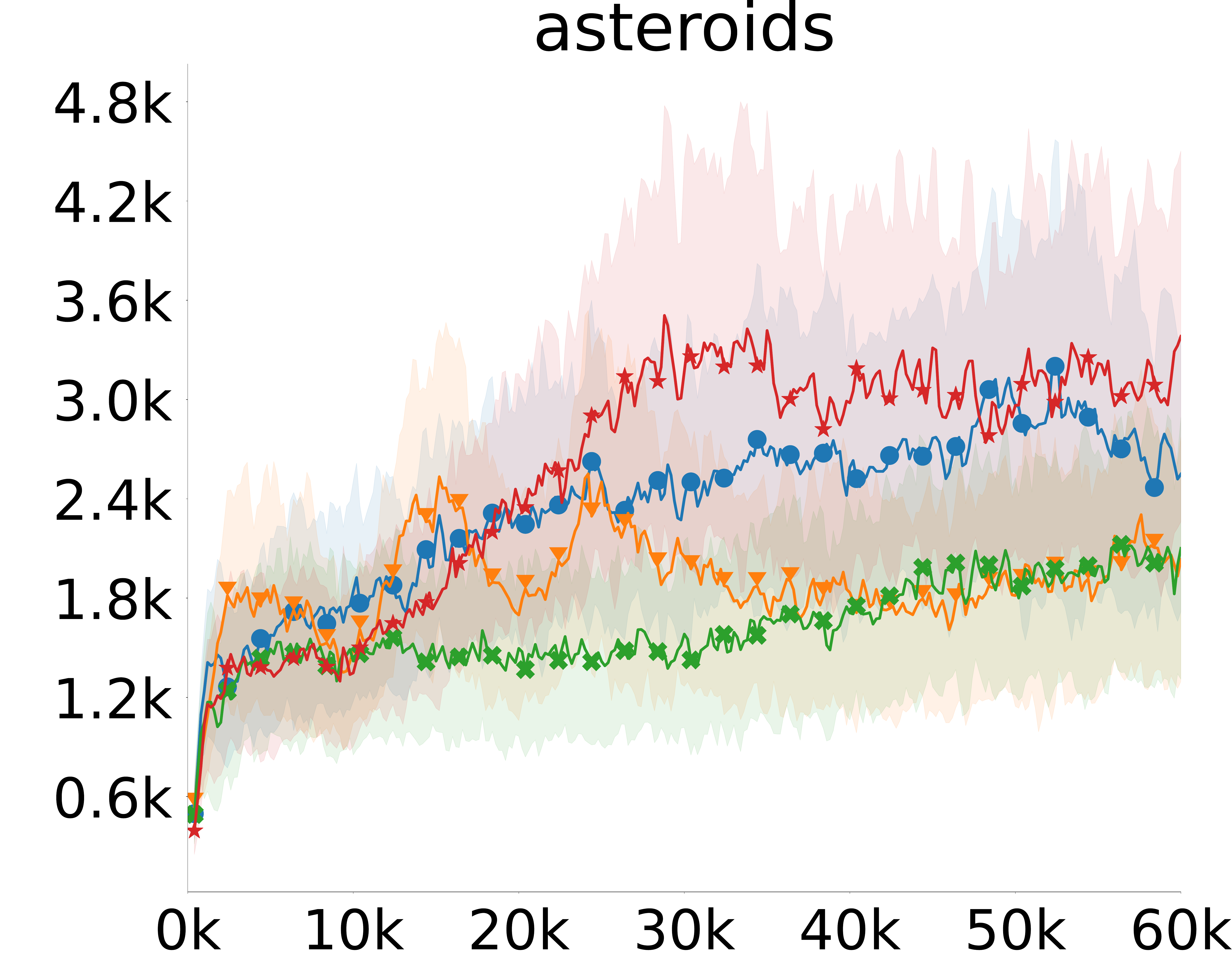}}
\subfloat{
    \includegraphics[width=0.15\textwidth]{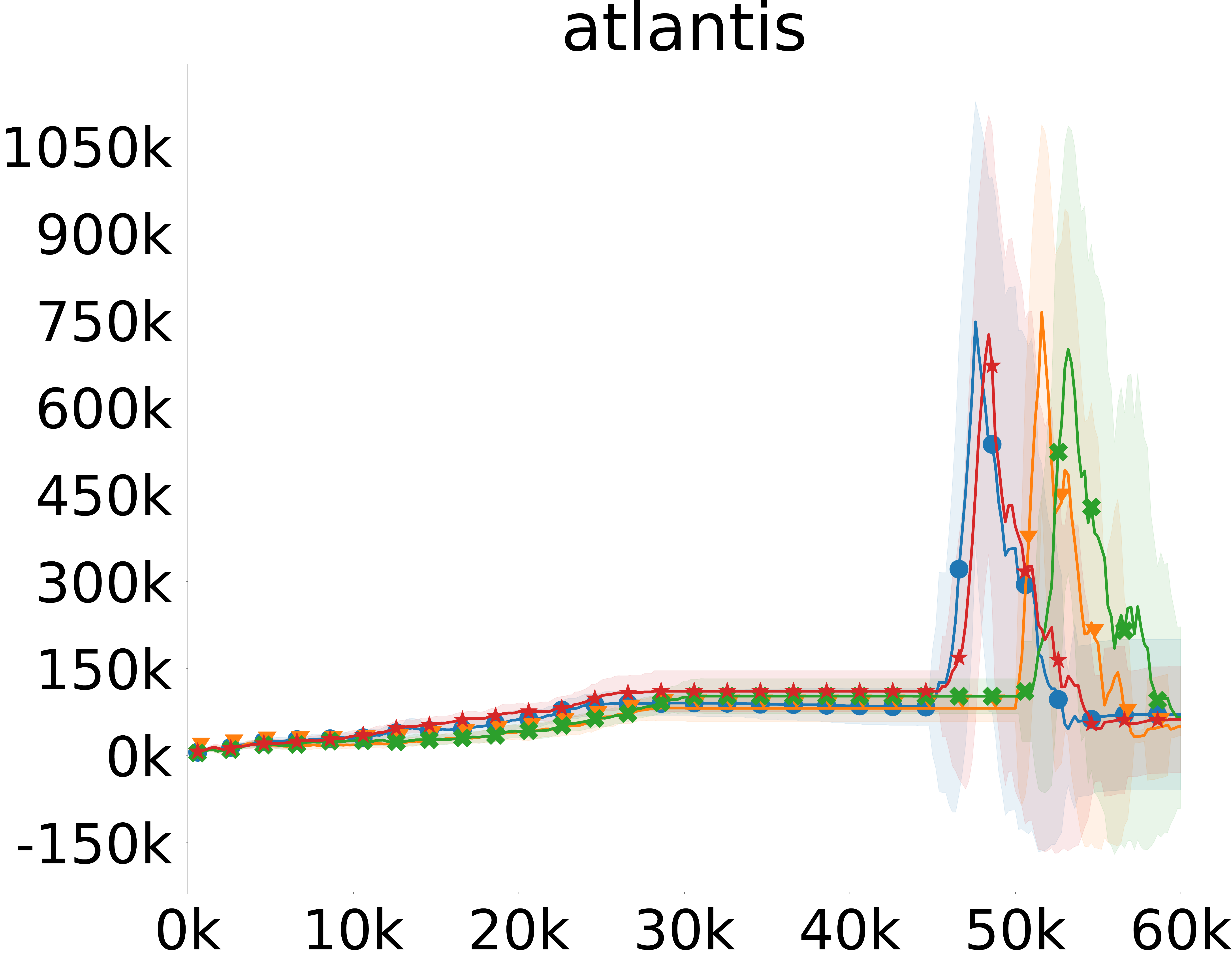}}
\\\vspace*{-0.8em}
\subfloat{
    \includegraphics[width=0.15\textwidth]{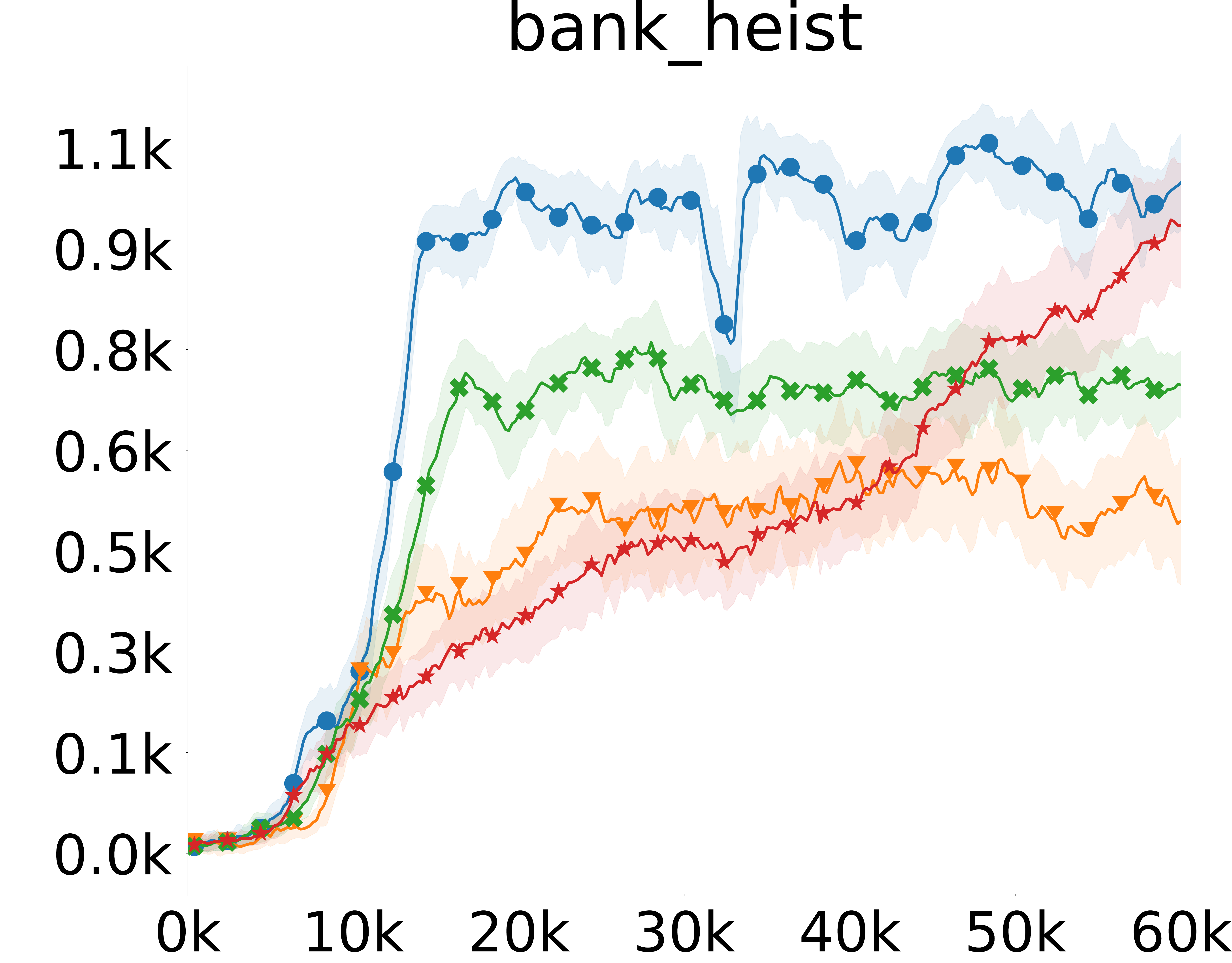}}
\subfloat{
    \includegraphics[width=0.15\textwidth]{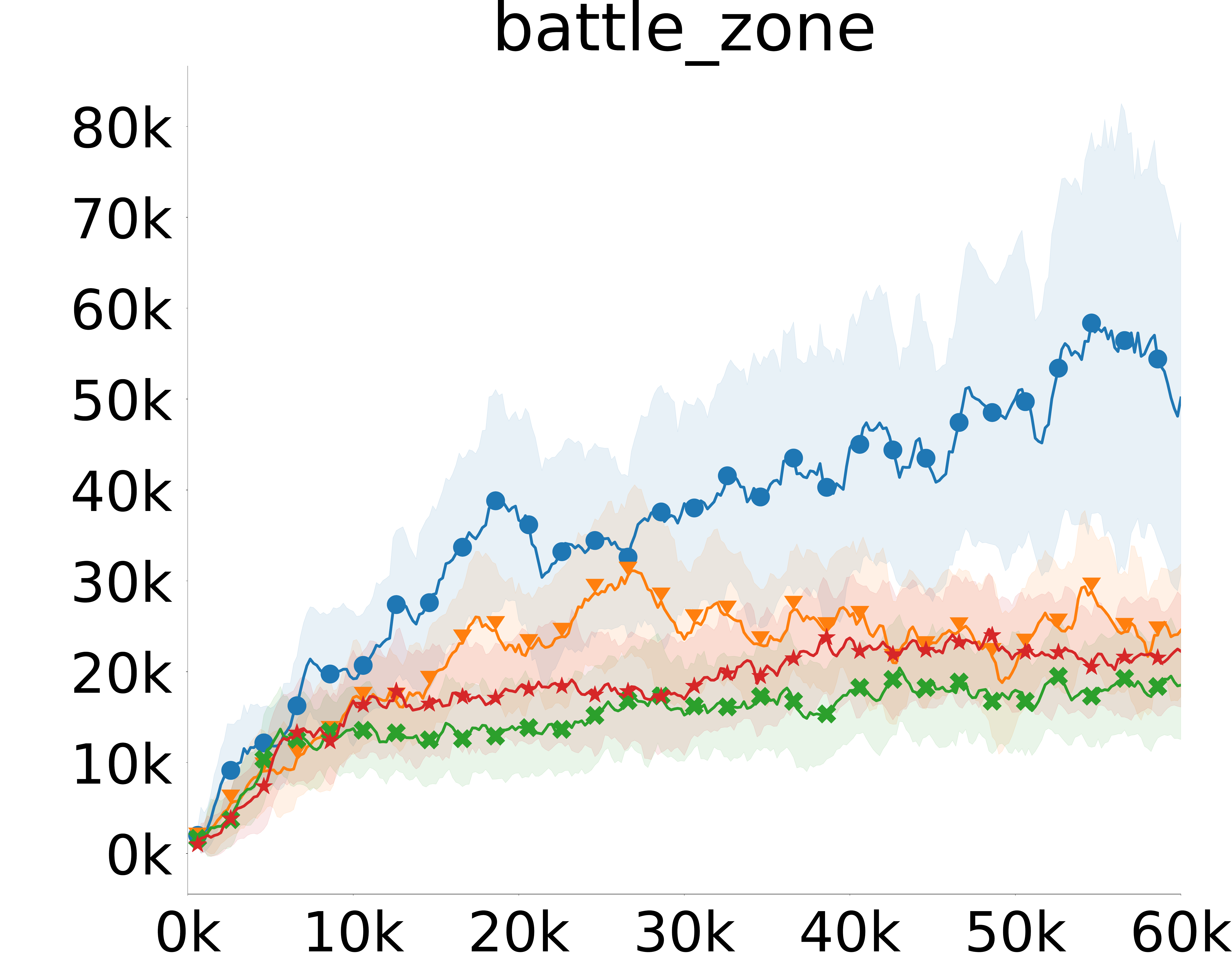}}
\subfloat{
    \includegraphics[width=0.15\textwidth]{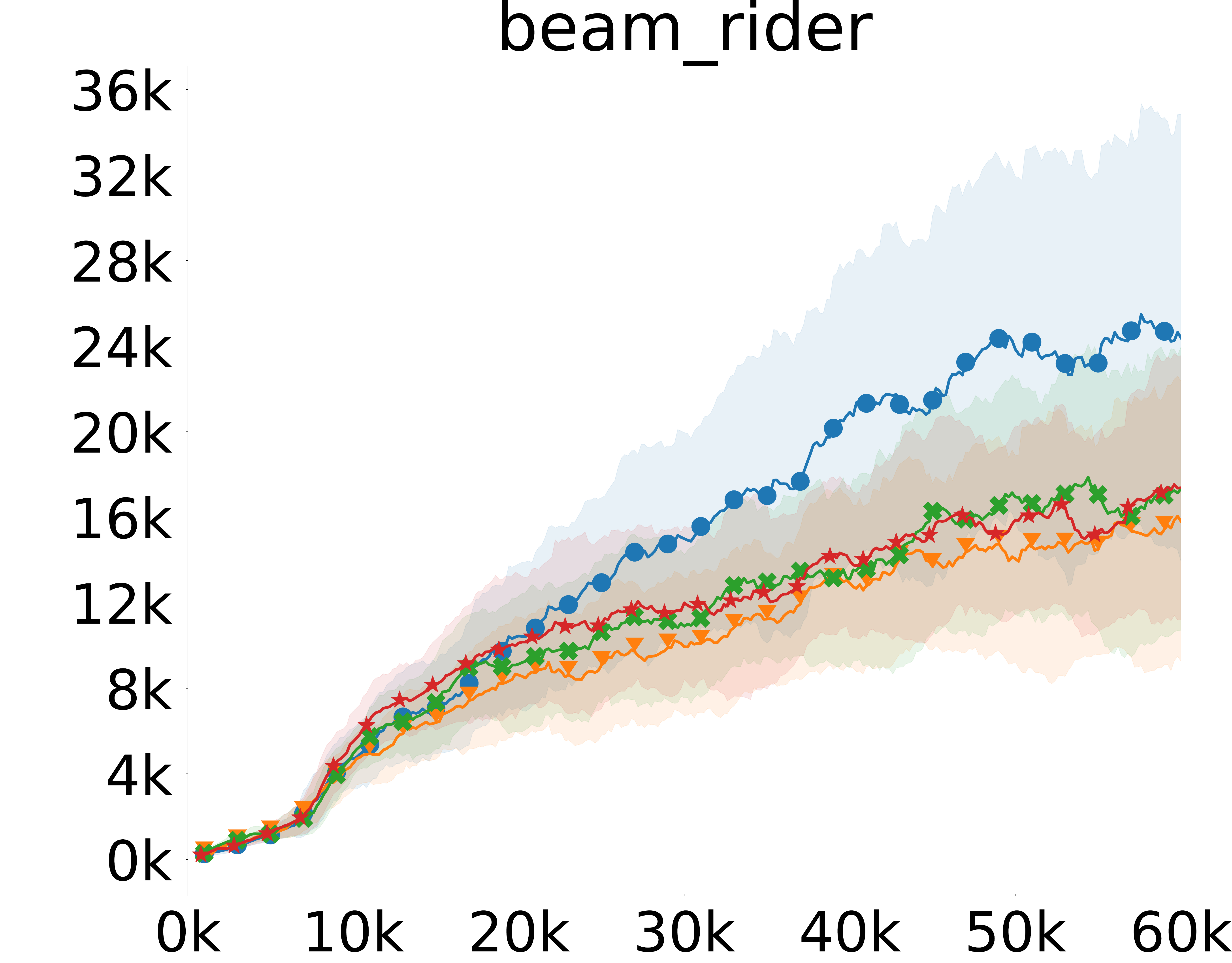}}
\subfloat{
    \includegraphics[width=0.15\textwidth]{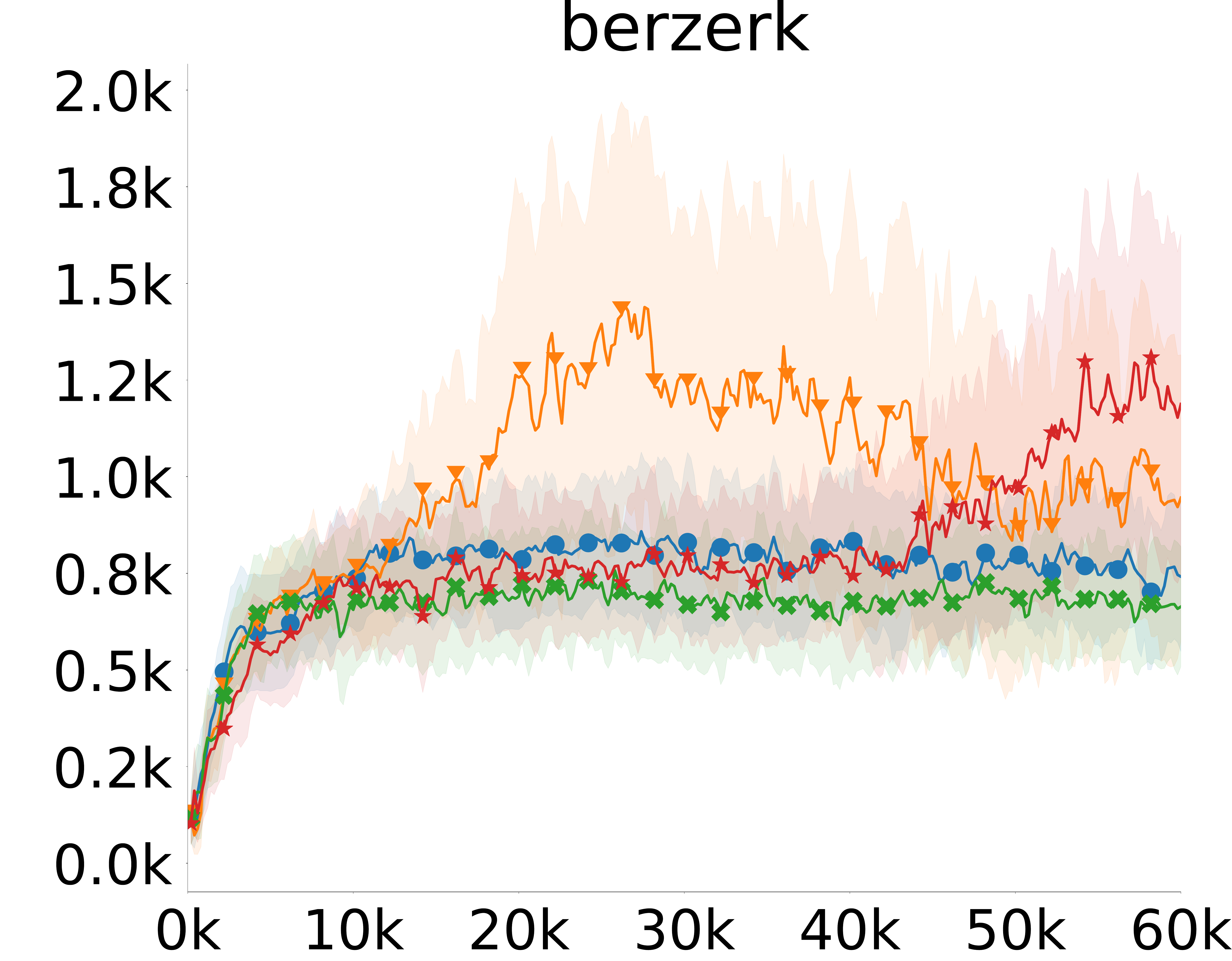}}
\subfloat{
    \includegraphics[width=0.15\textwidth]{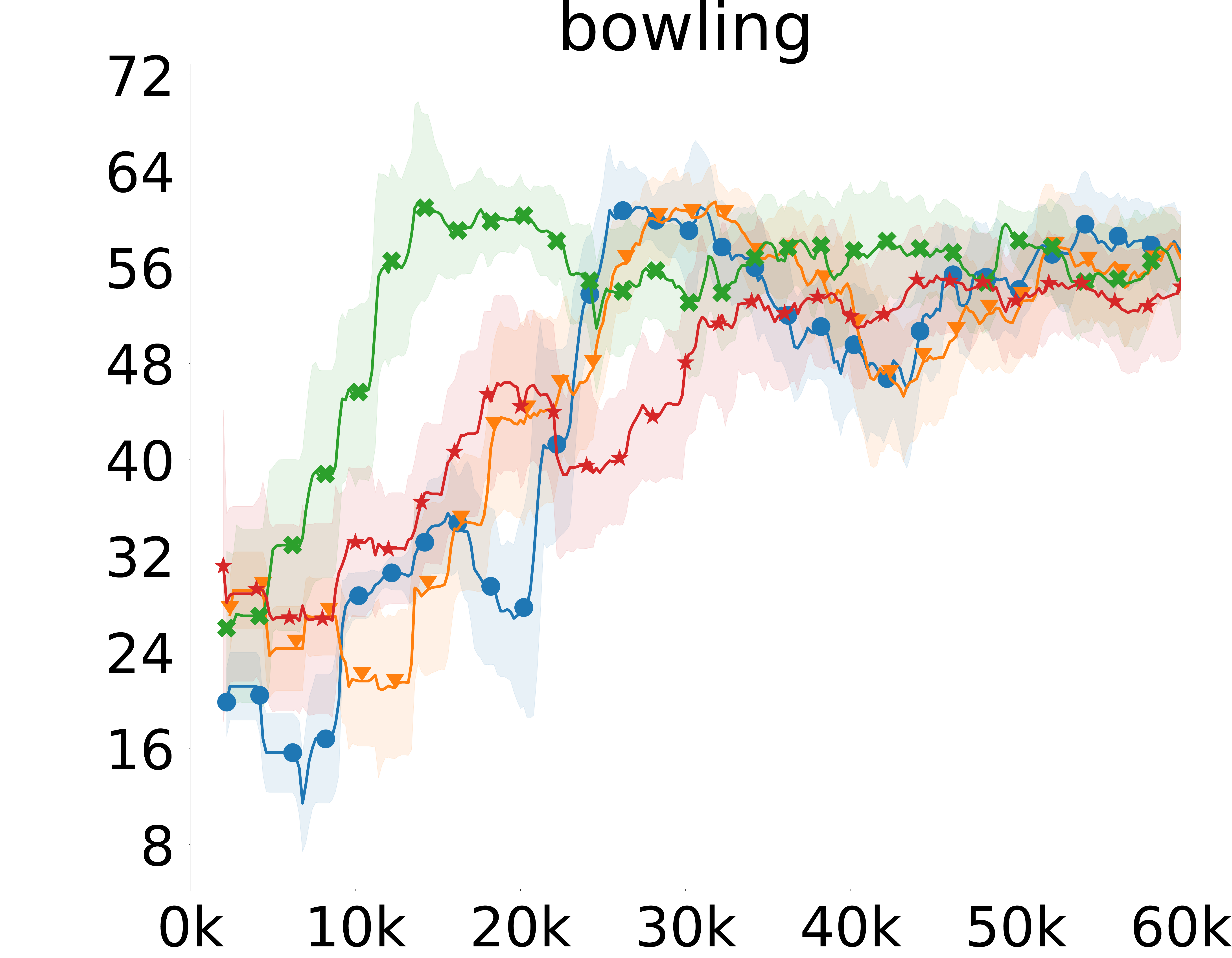}}
\subfloat{
    \includegraphics[width=0.15\textwidth]{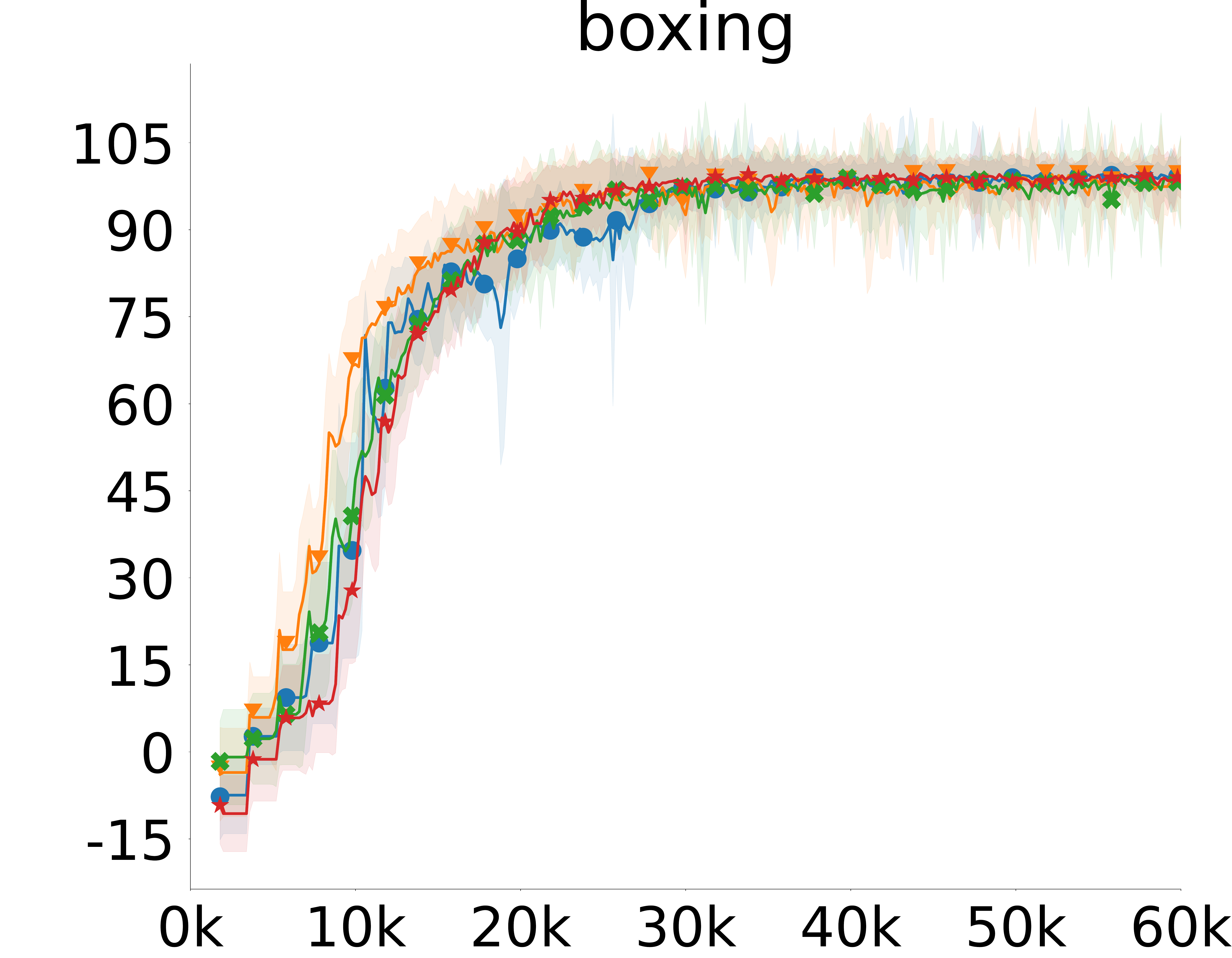}}
\\\vspace*{-0.8em}
\subfloat{
    \includegraphics[width=0.15\textwidth]{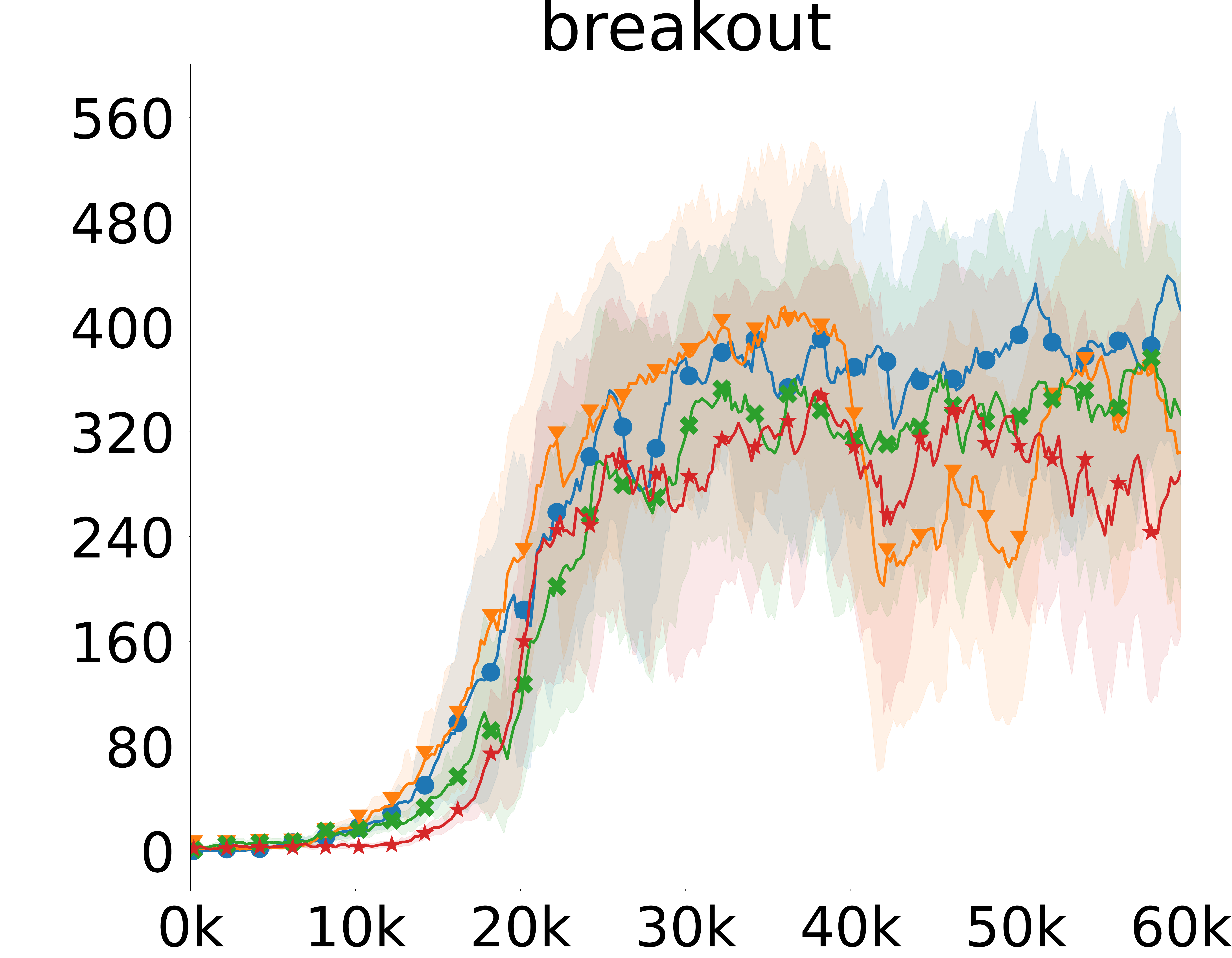}}
\subfloat{
    \includegraphics[width=0.15\textwidth]{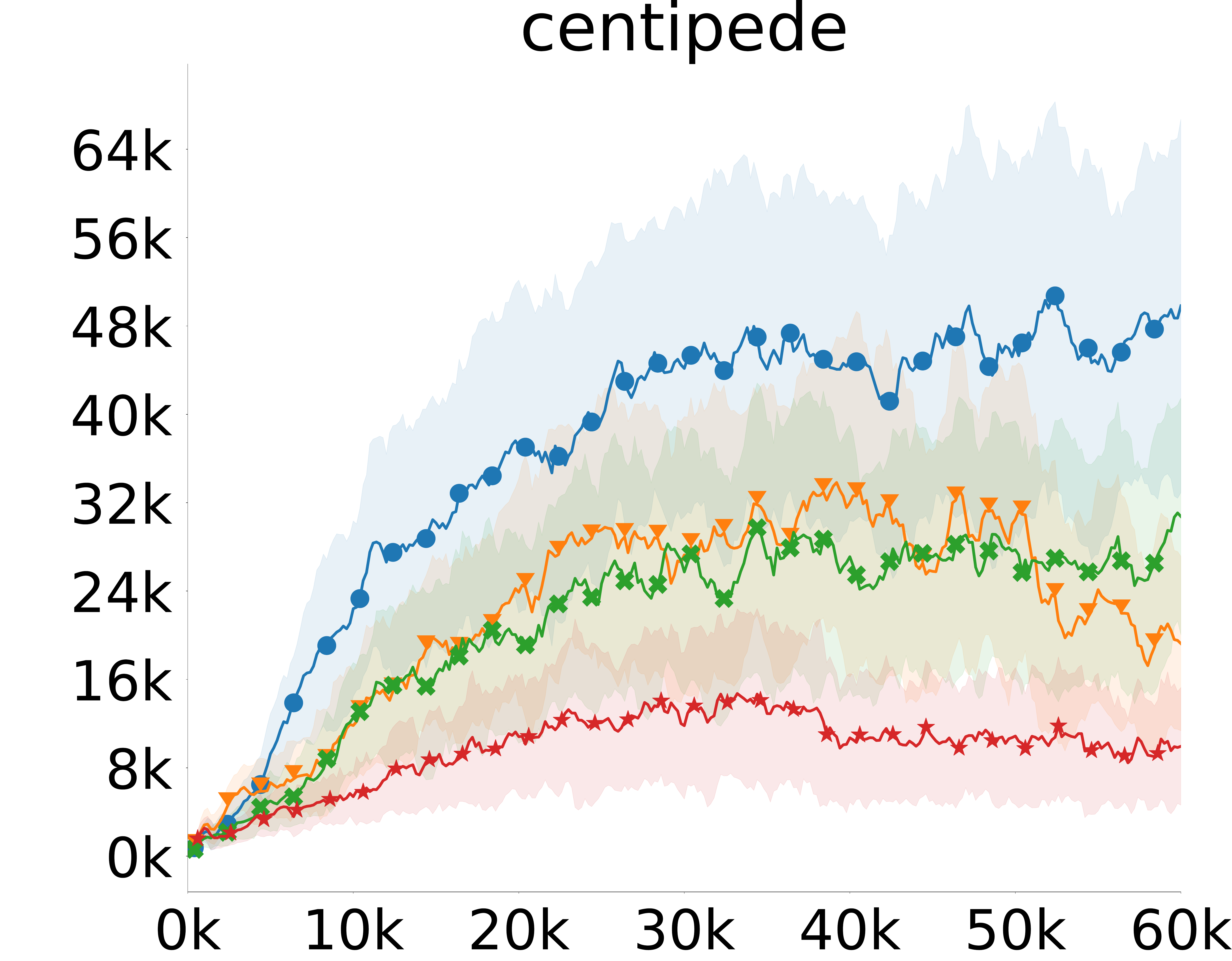}}
\subfloat{
    \includegraphics[width=0.15\textwidth]{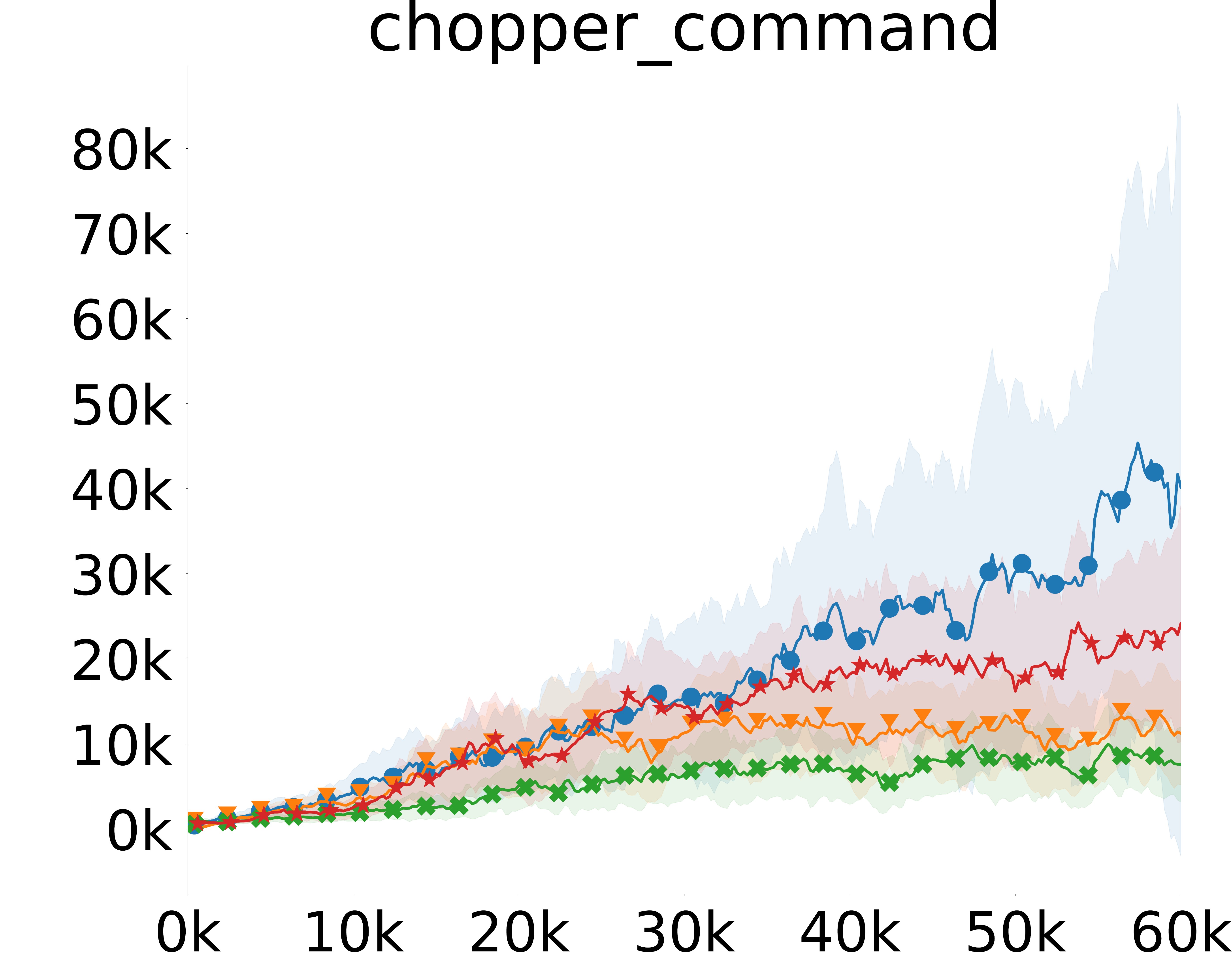}}
\subfloat{
    \includegraphics[width=0.15\textwidth]{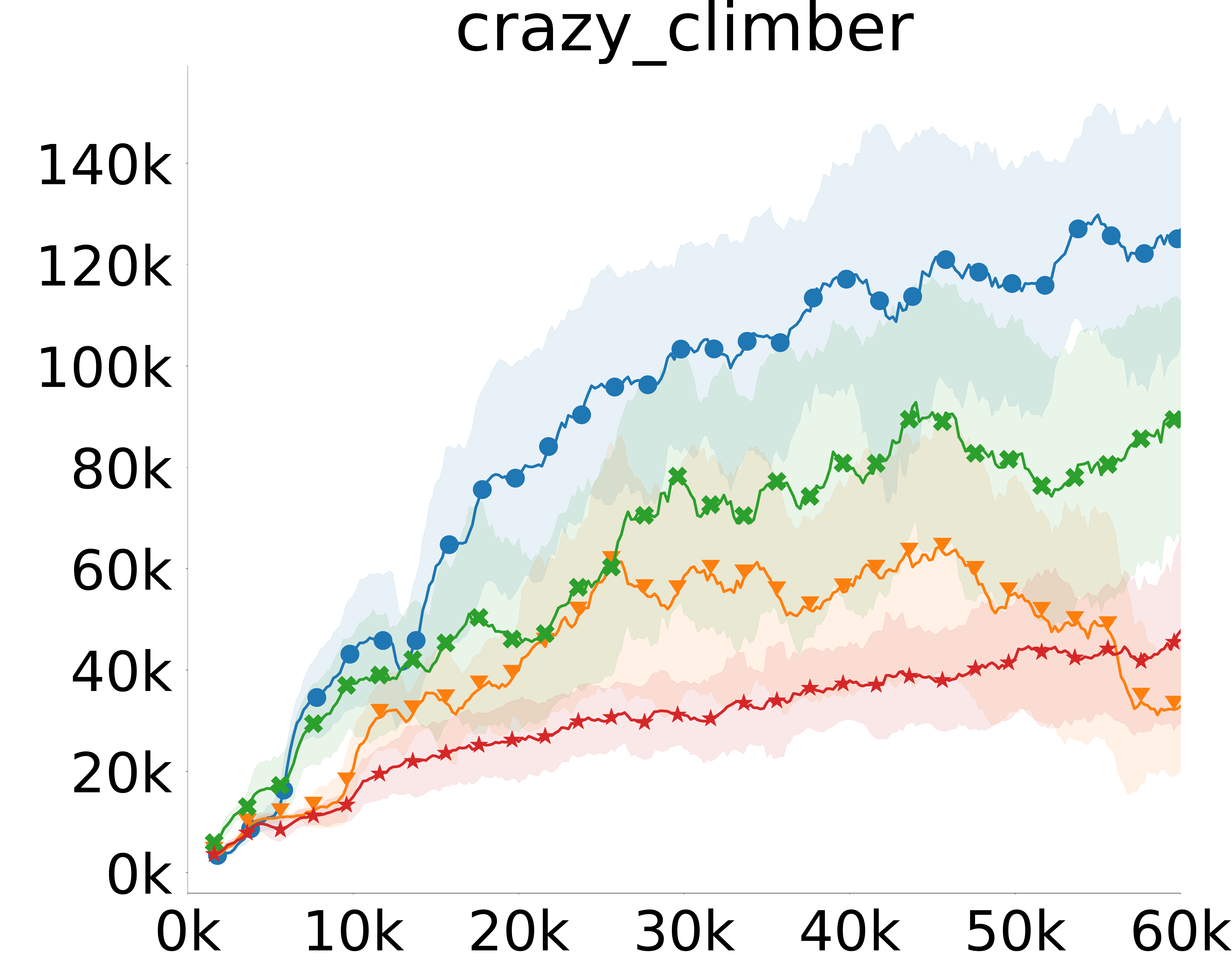}}
\subfloat{
    \includegraphics[width=0.15\textwidth]{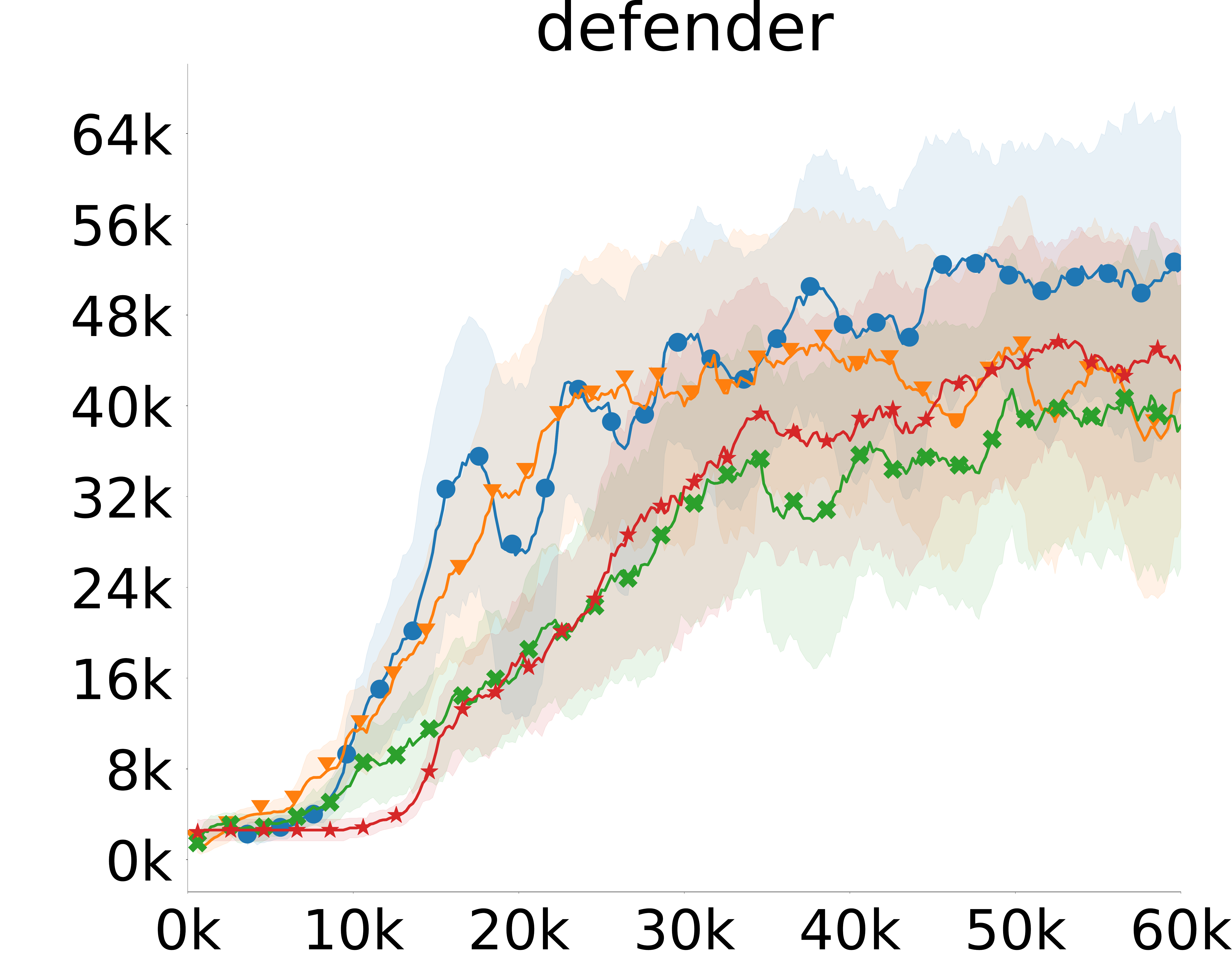}}
\subfloat{
    \includegraphics[width=0.15\textwidth]{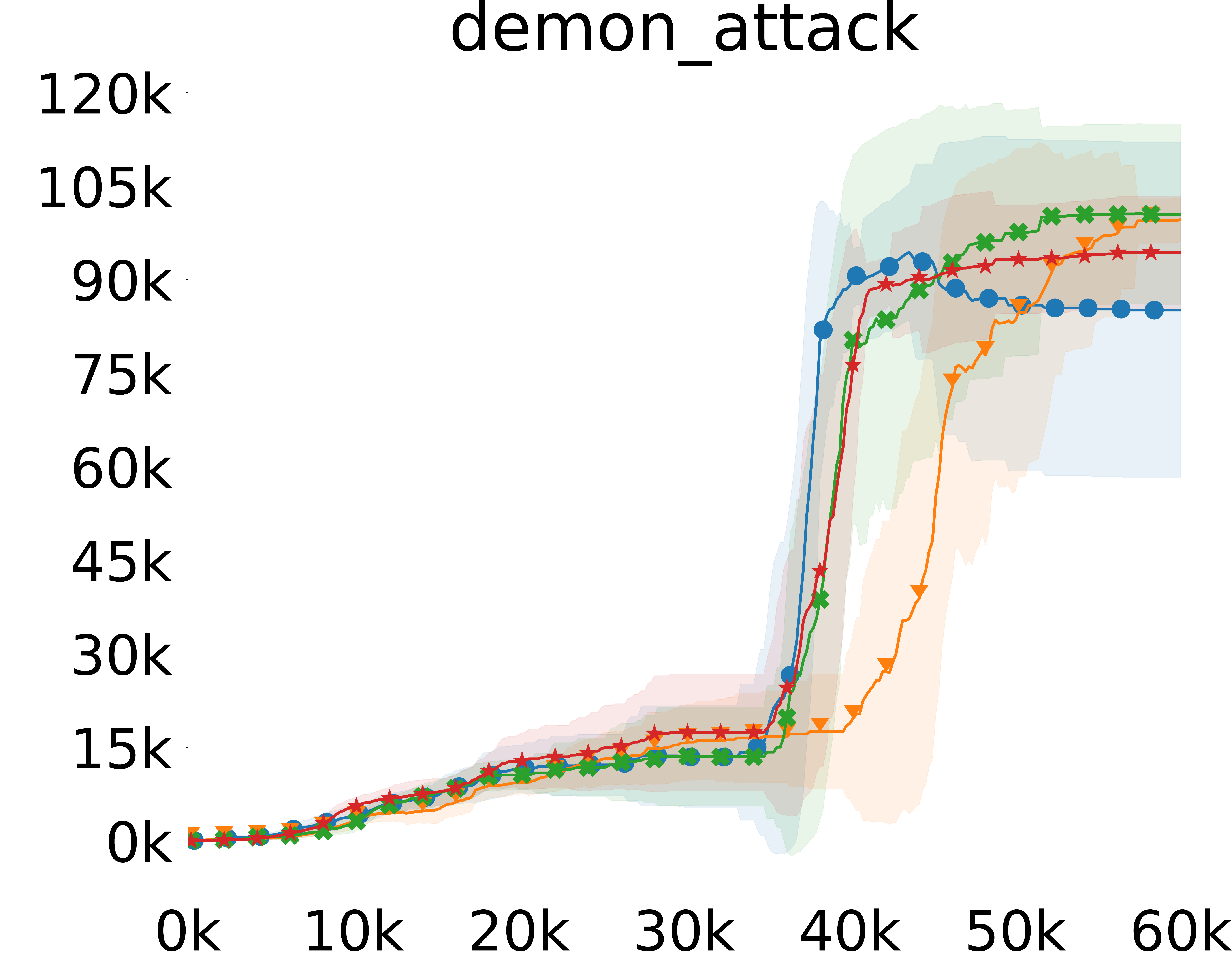}}
\\\vspace*{-0.8em}
\subfloat{
    \includegraphics[width=0.15\textwidth]{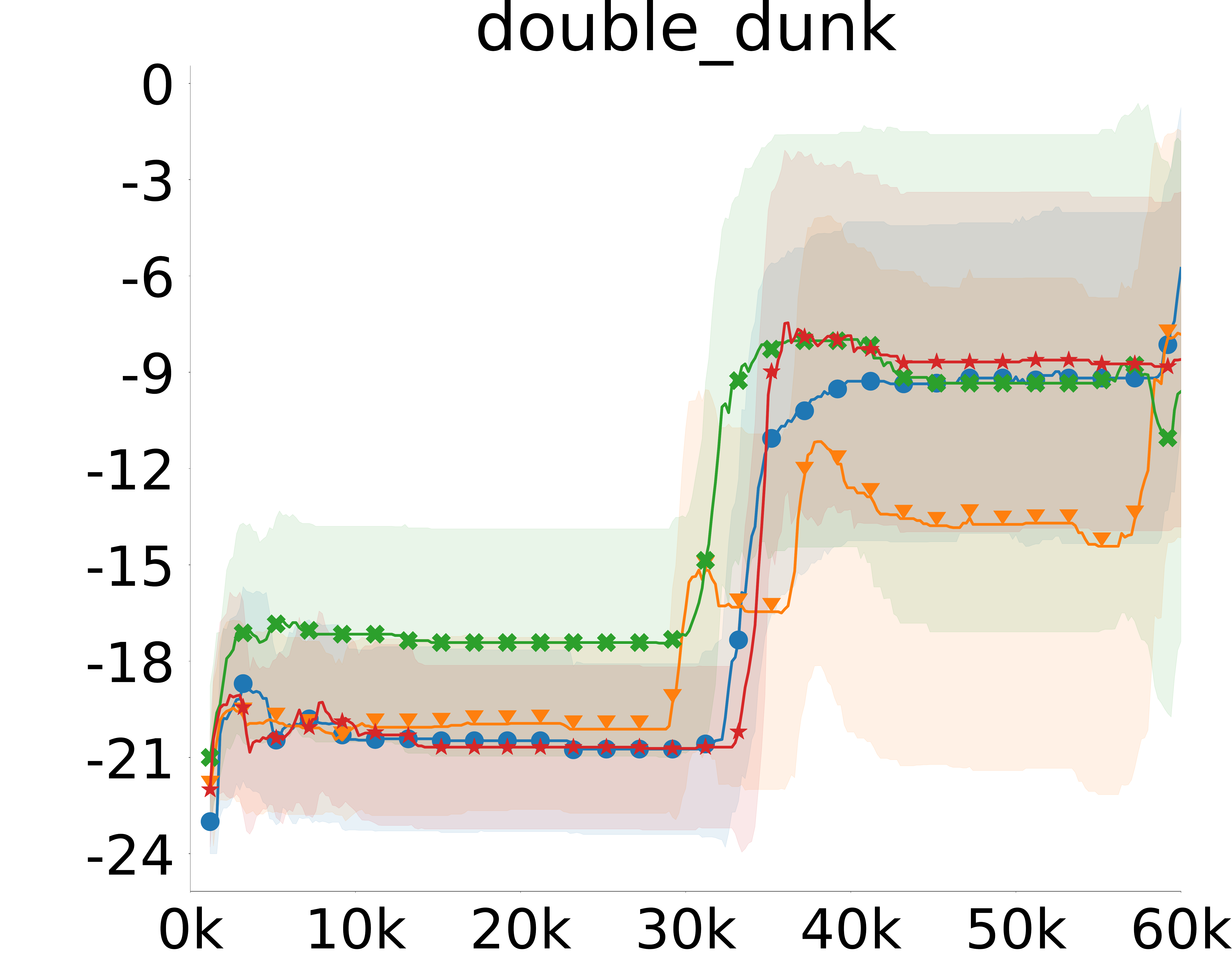}}
\subfloat{
    \includegraphics[width=0.15\textwidth]{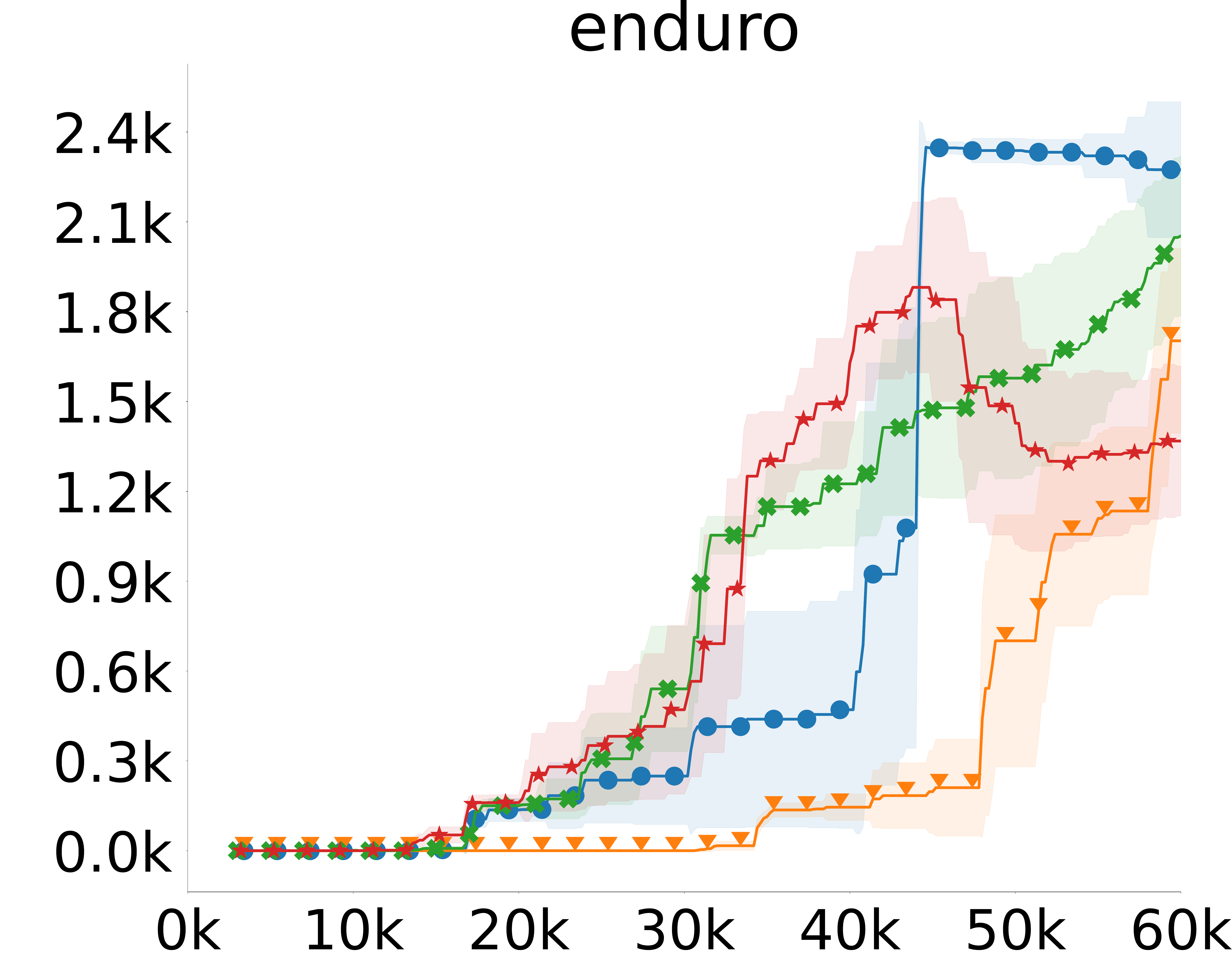}}
\subfloat{
    \includegraphics[width=0.15\textwidth]{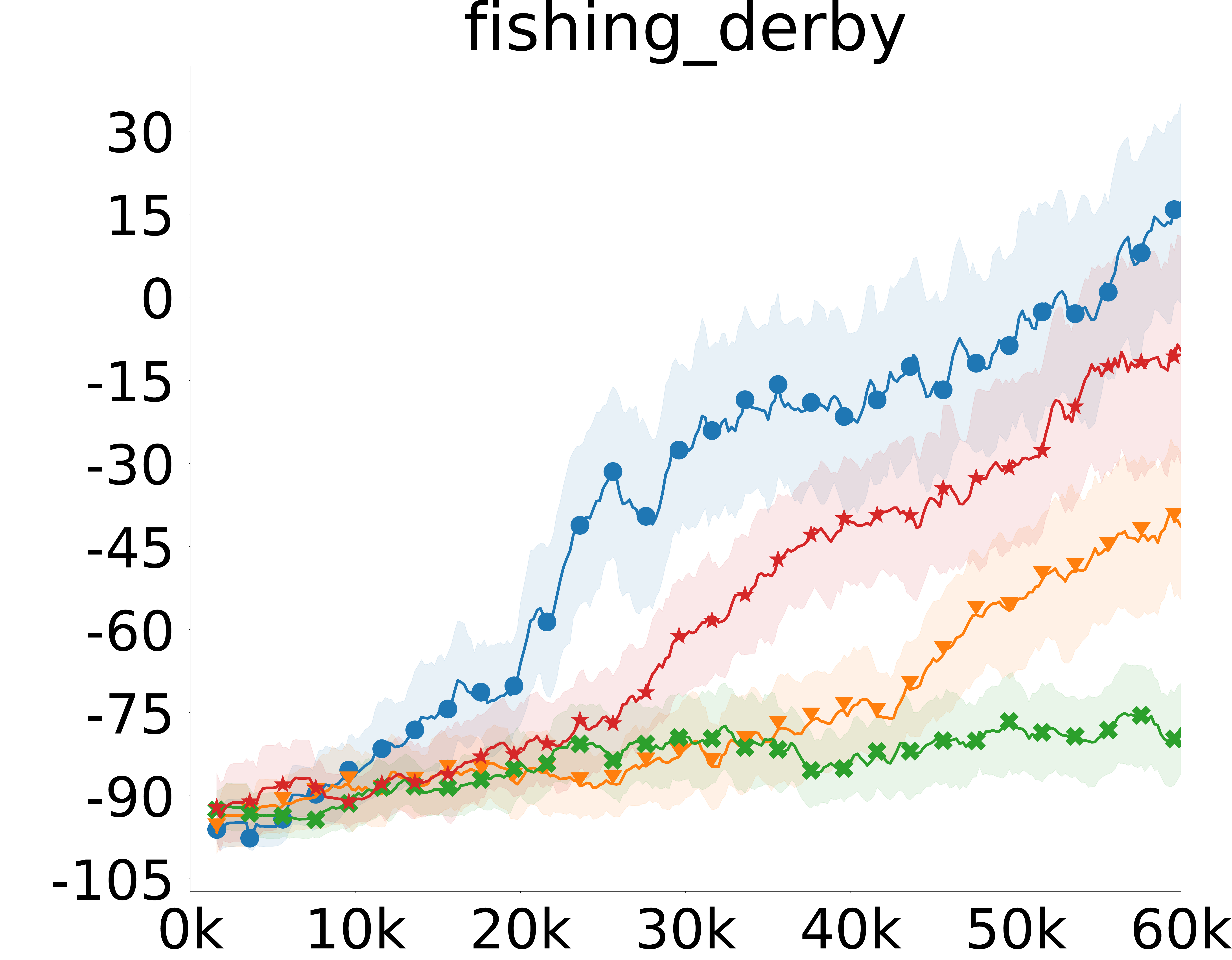}}
\subfloat{
    \includegraphics[width=0.15\textwidth]{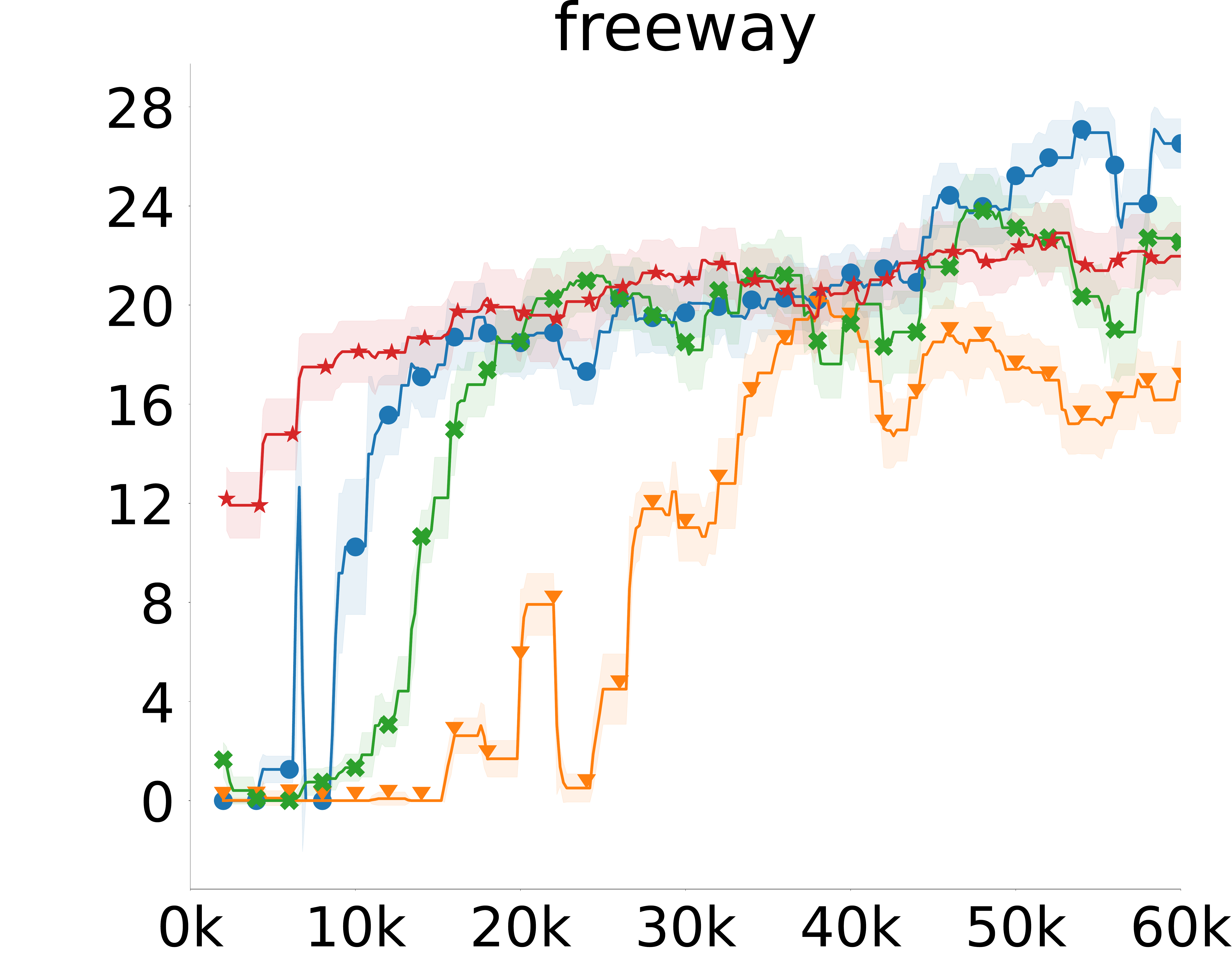}}
\subfloat{
    \includegraphics[width=0.15\textwidth]{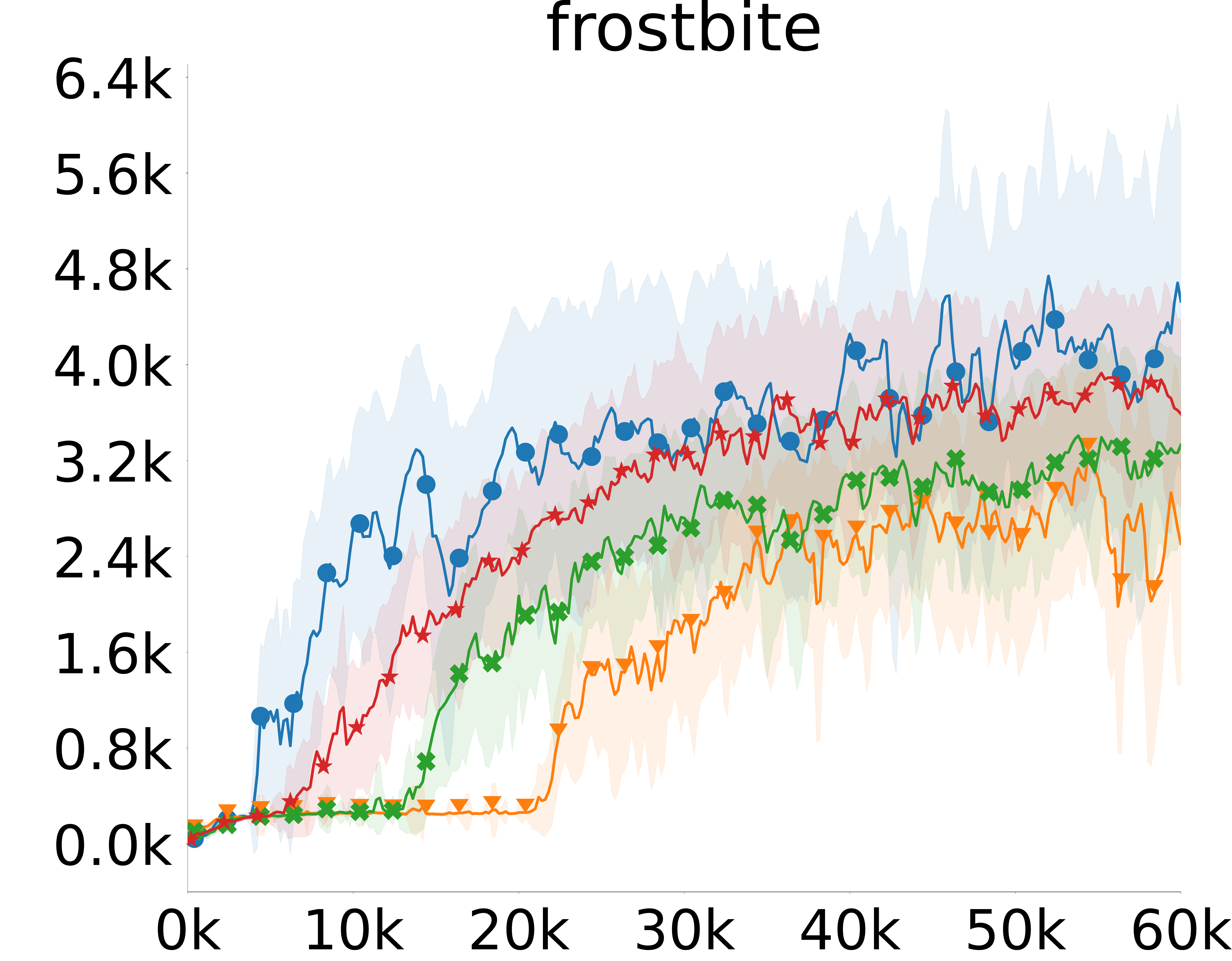}}
\subfloat{
    \includegraphics[width=0.15\textwidth]{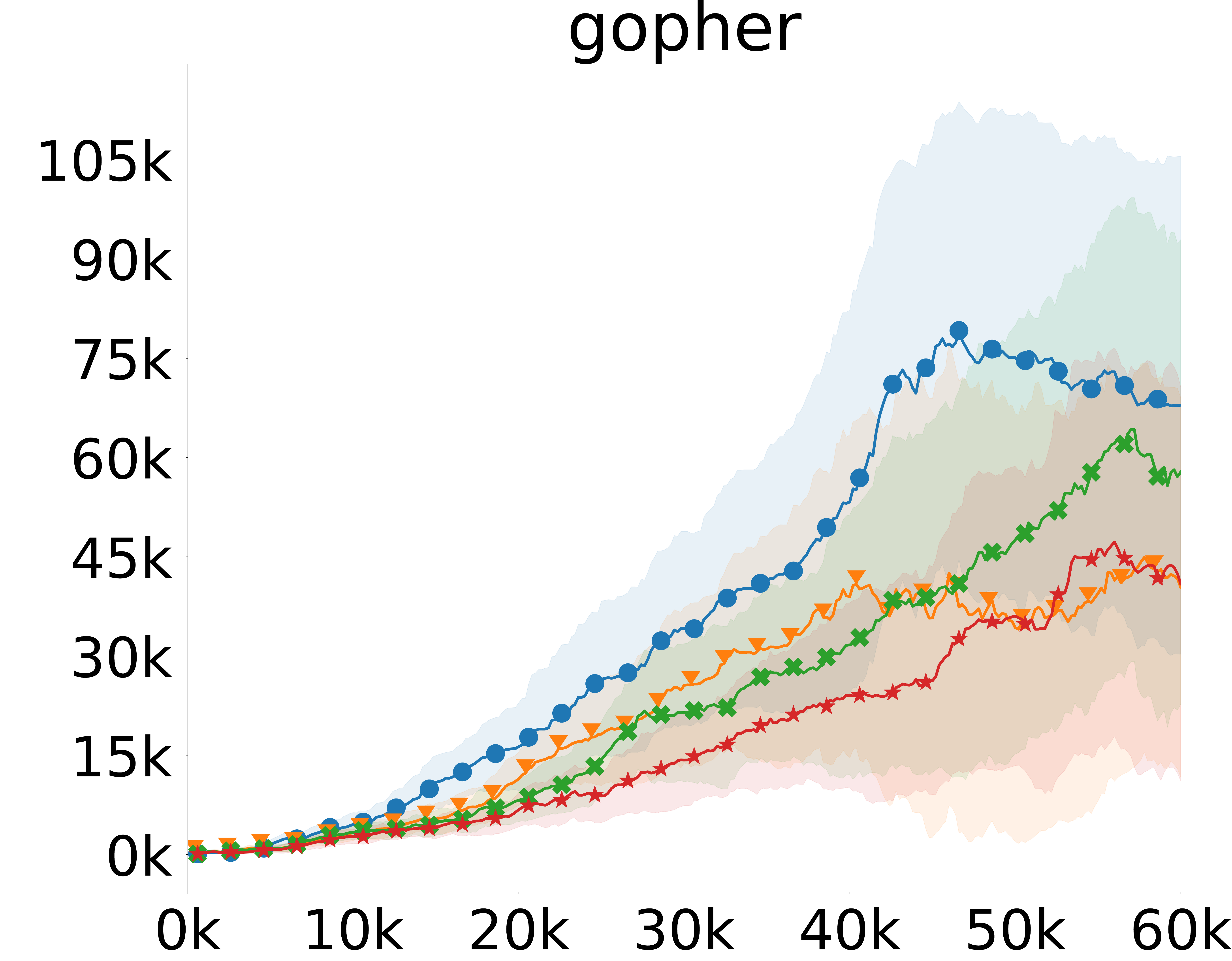}}
\\\vspace*{-0.8em}
\subfloat{
    \includegraphics[width=0.15\textwidth]{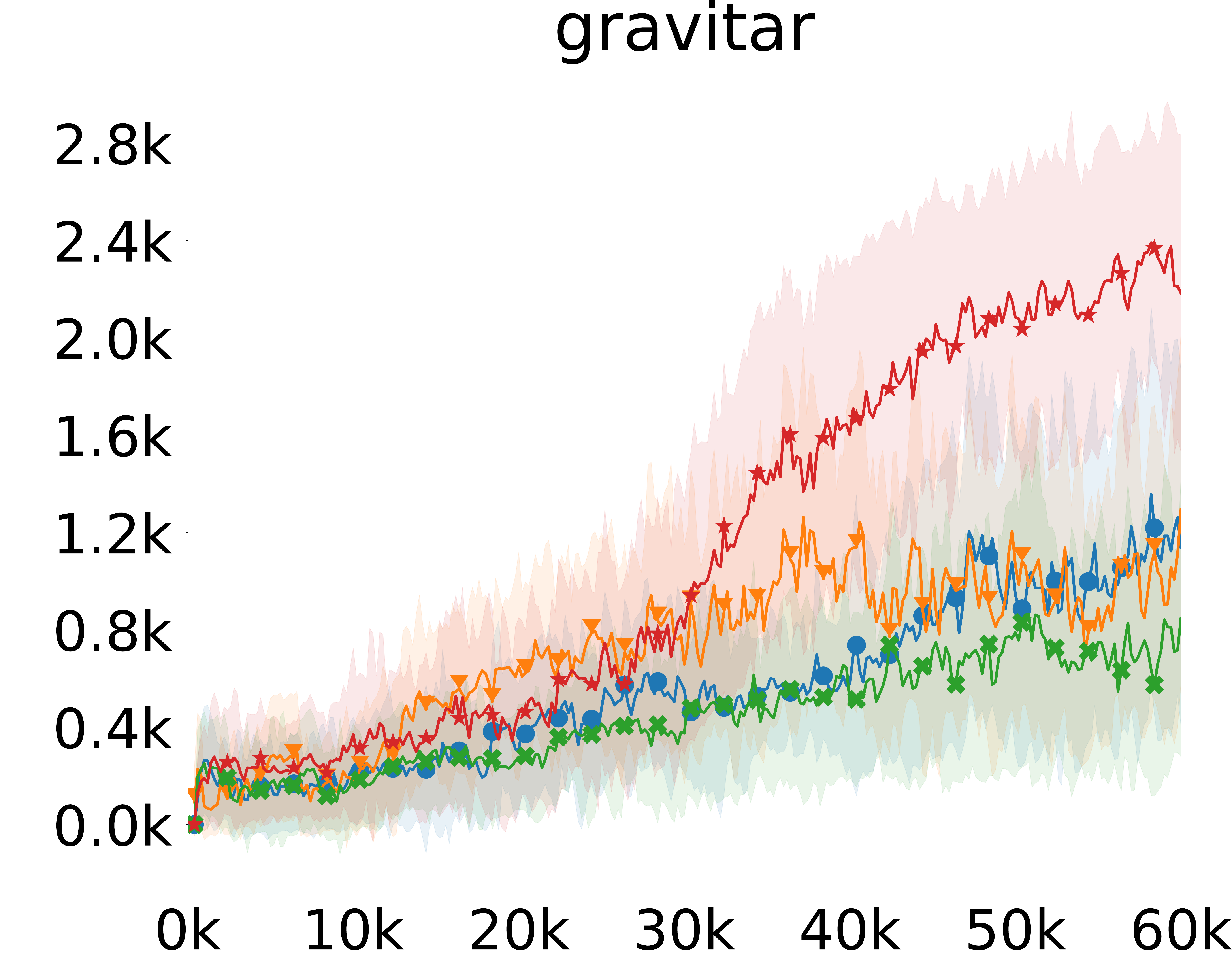}}
\subfloat{
    \includegraphics[width=0.15\textwidth]{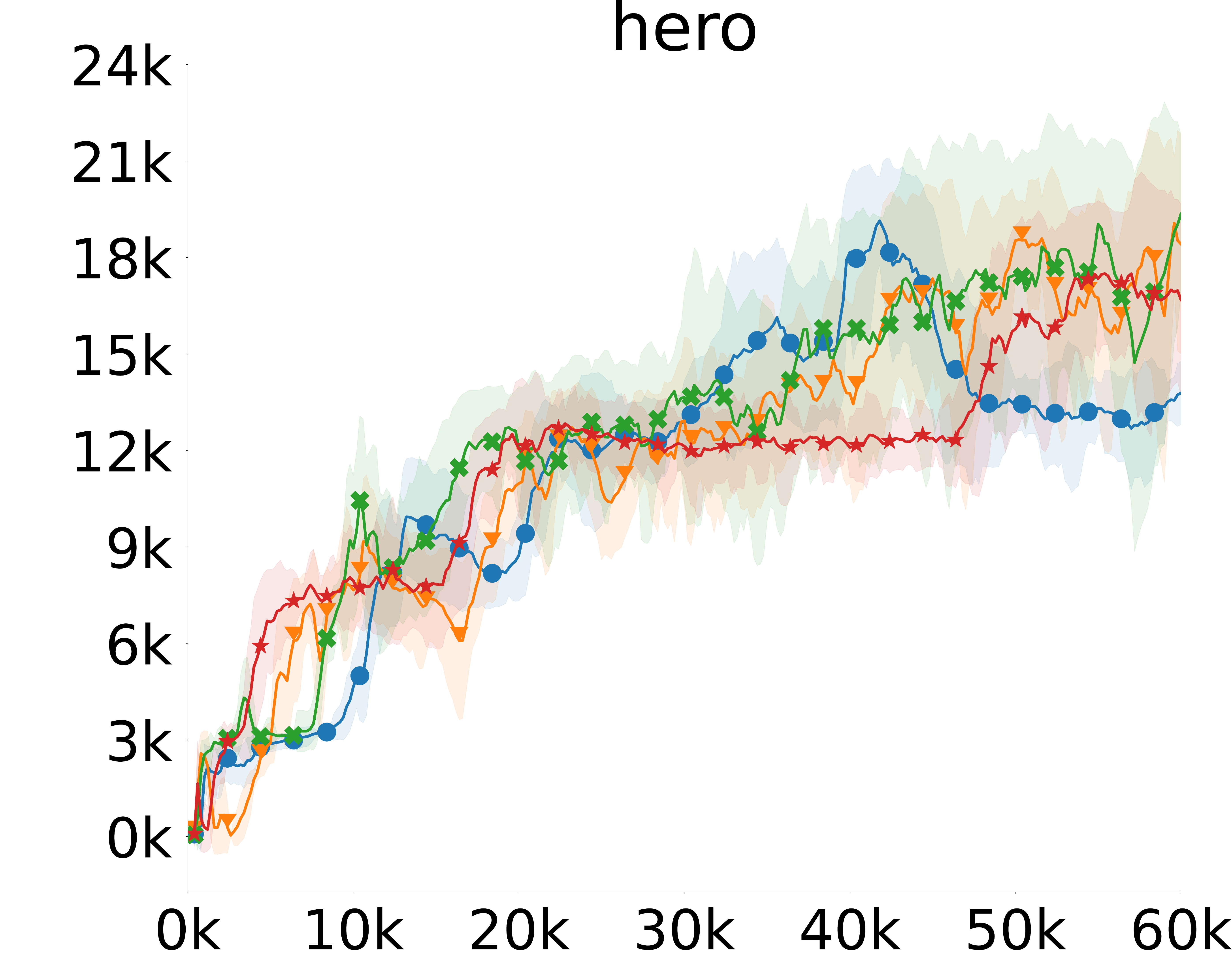}}
\subfloat{
    \includegraphics[width=0.15\textwidth]{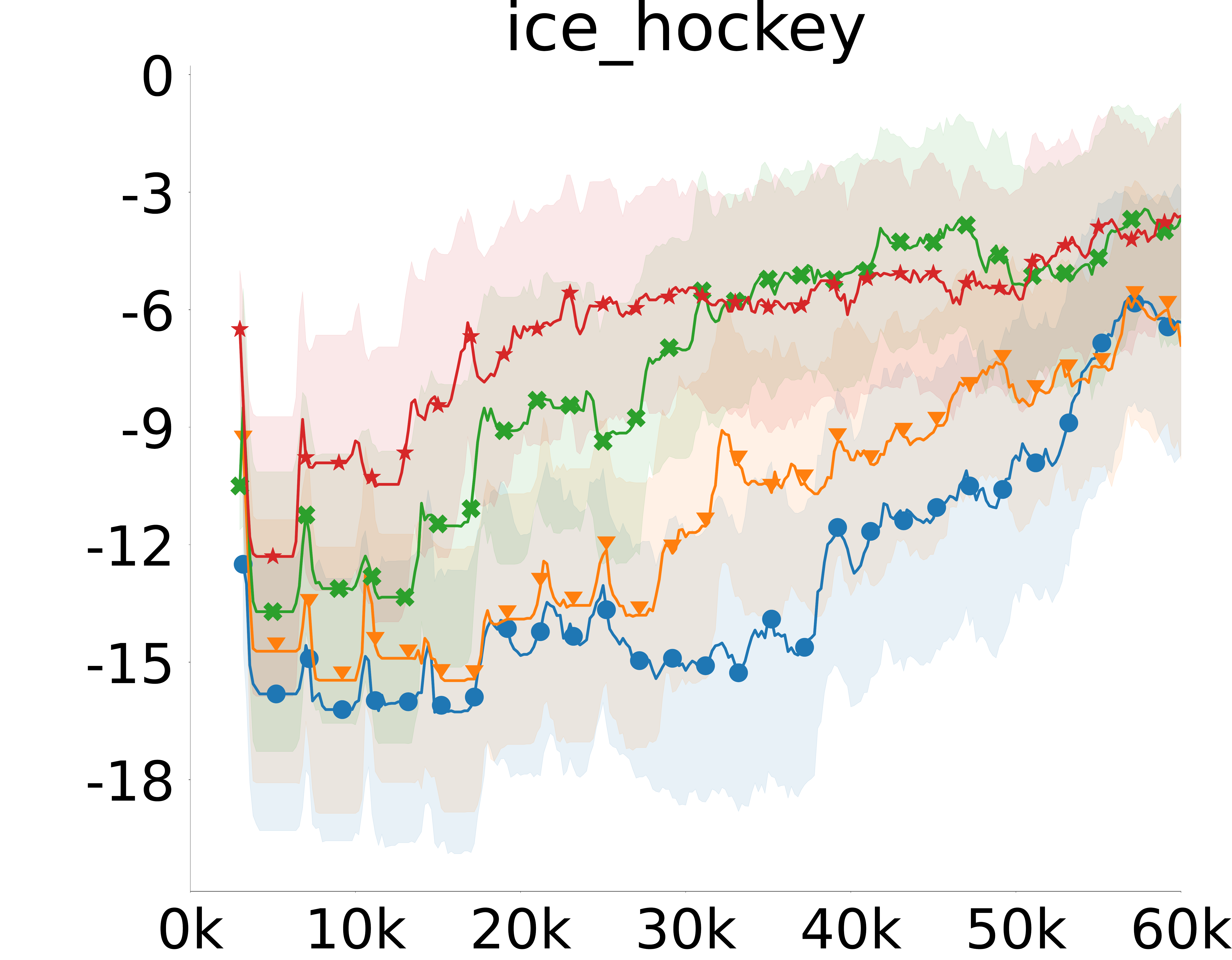}}
\subfloat{
    \includegraphics[width=0.15\textwidth]{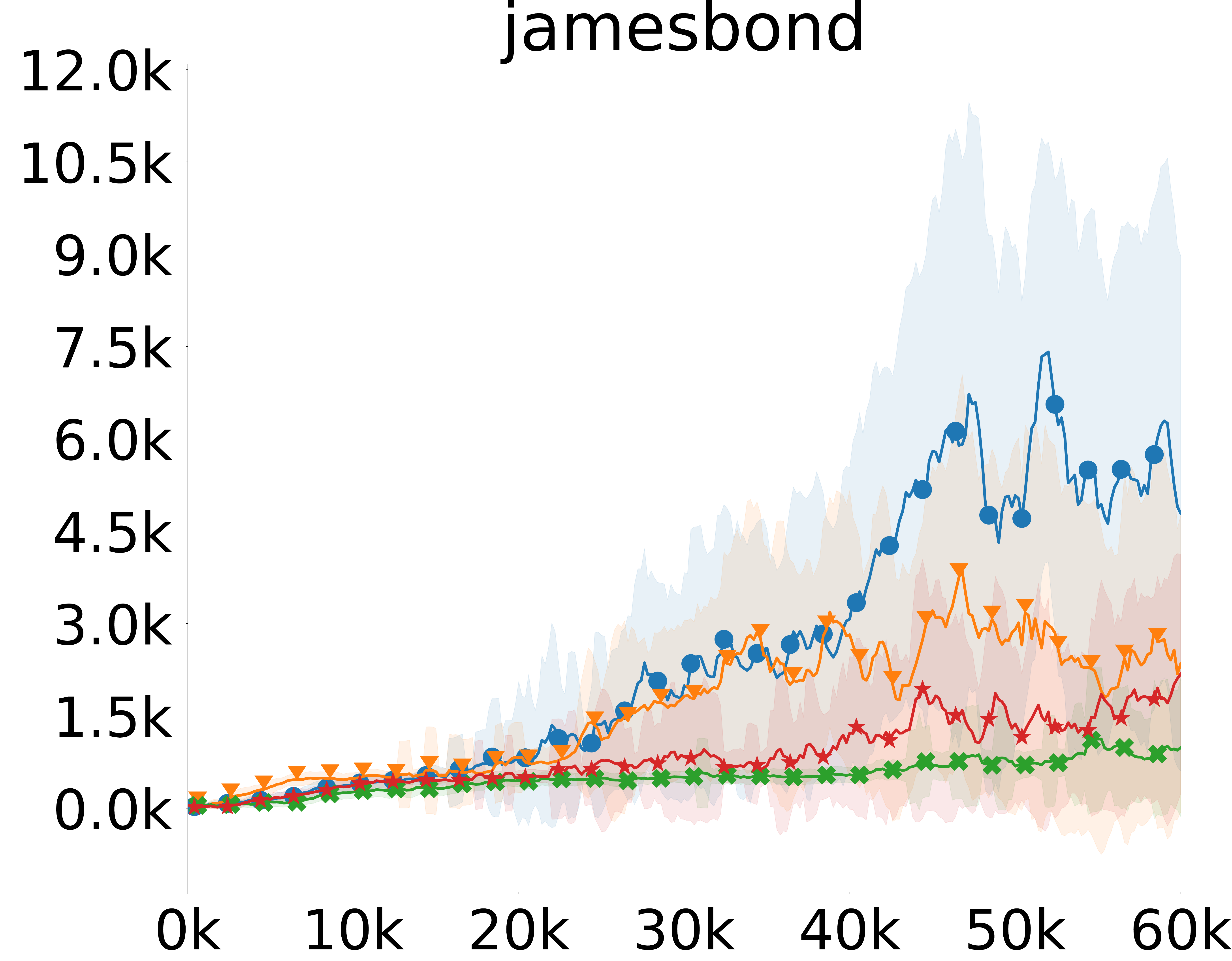}}
\subfloat{
    \includegraphics[width=0.15\textwidth]{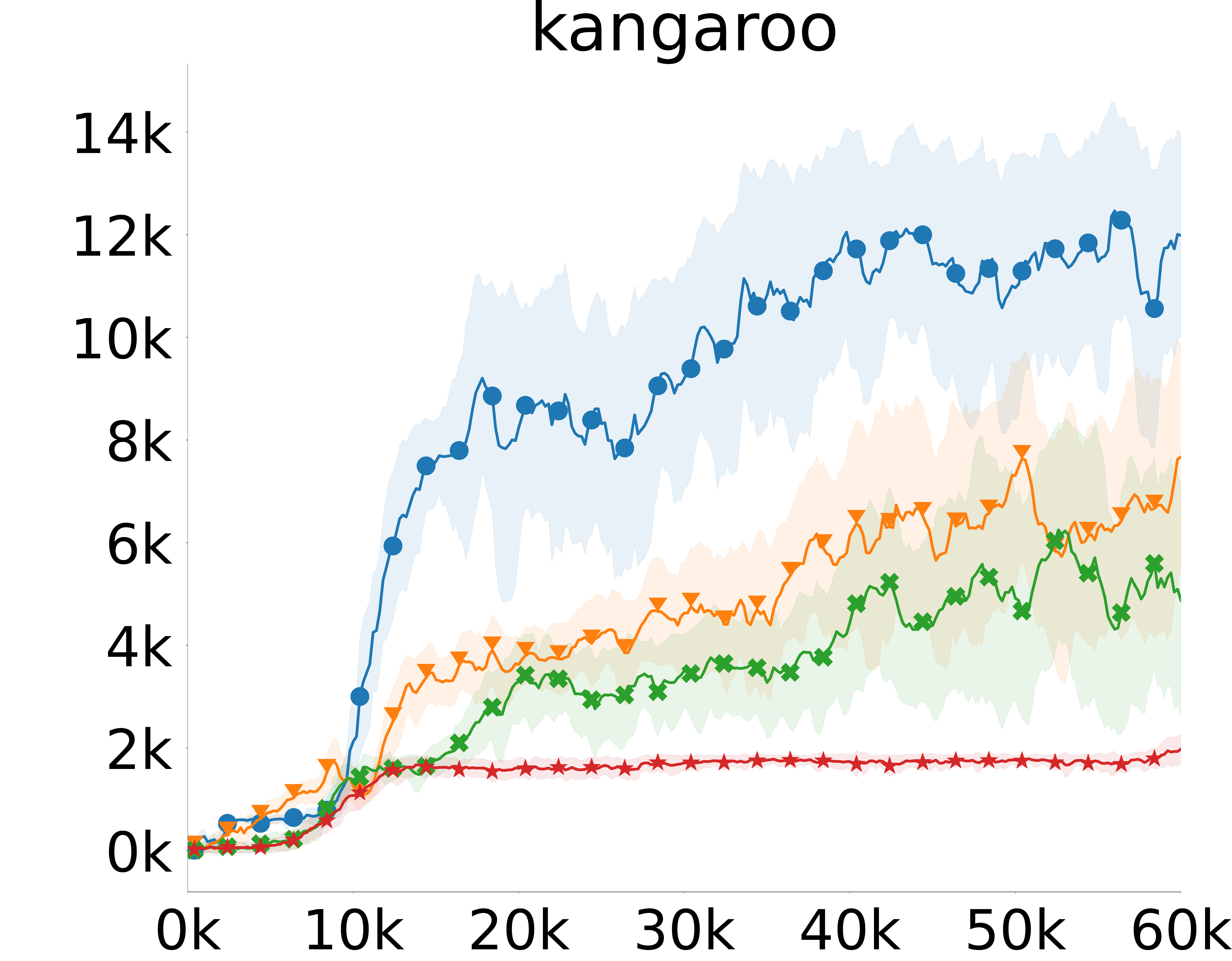}}
\subfloat{
    \includegraphics[width=0.15\textwidth]{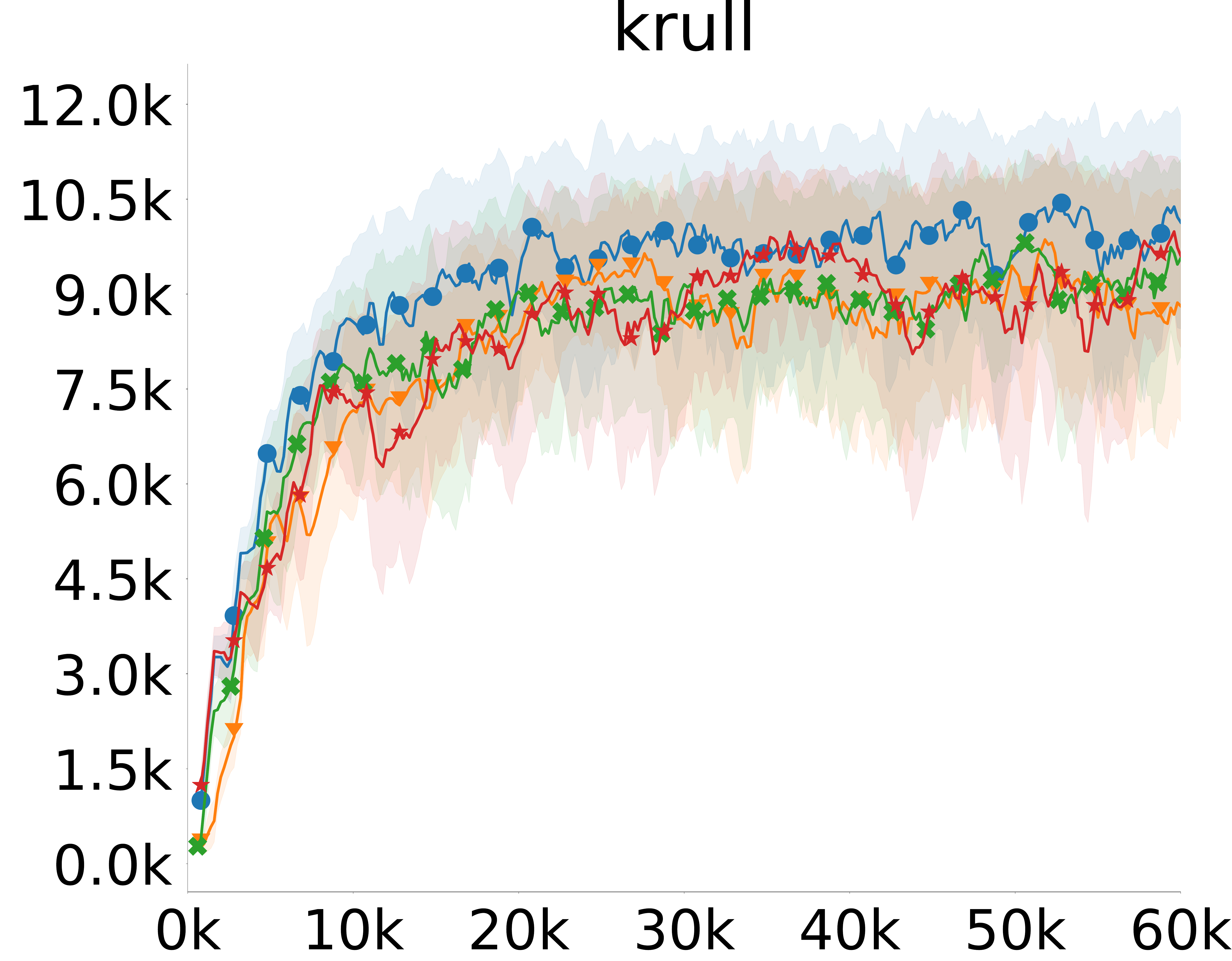}}
\\\vspace*{-0.8em}
\subfloat{
    \includegraphics[width=0.15\textwidth]{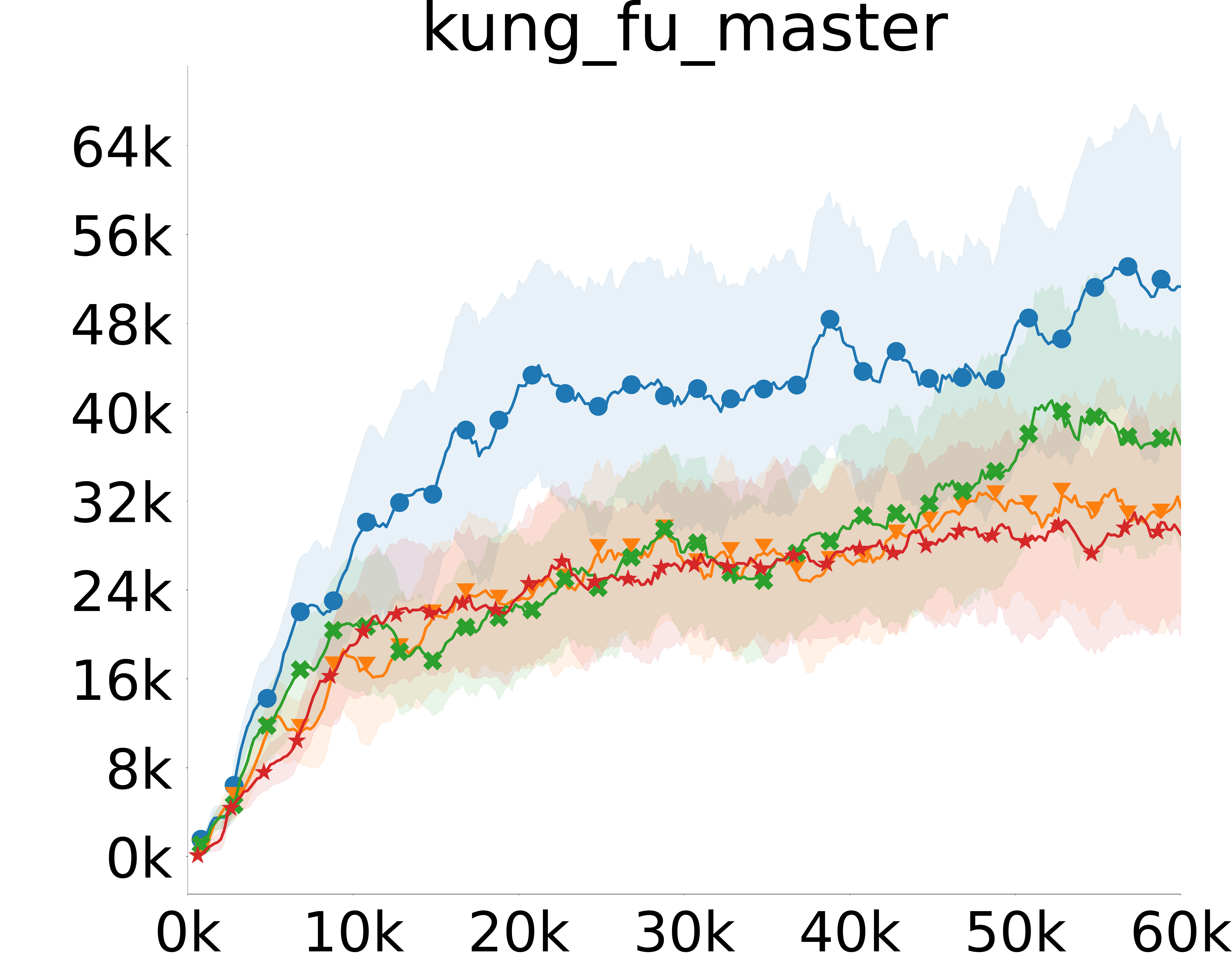}}
\subfloat{
    \includegraphics[width=0.15\textwidth]{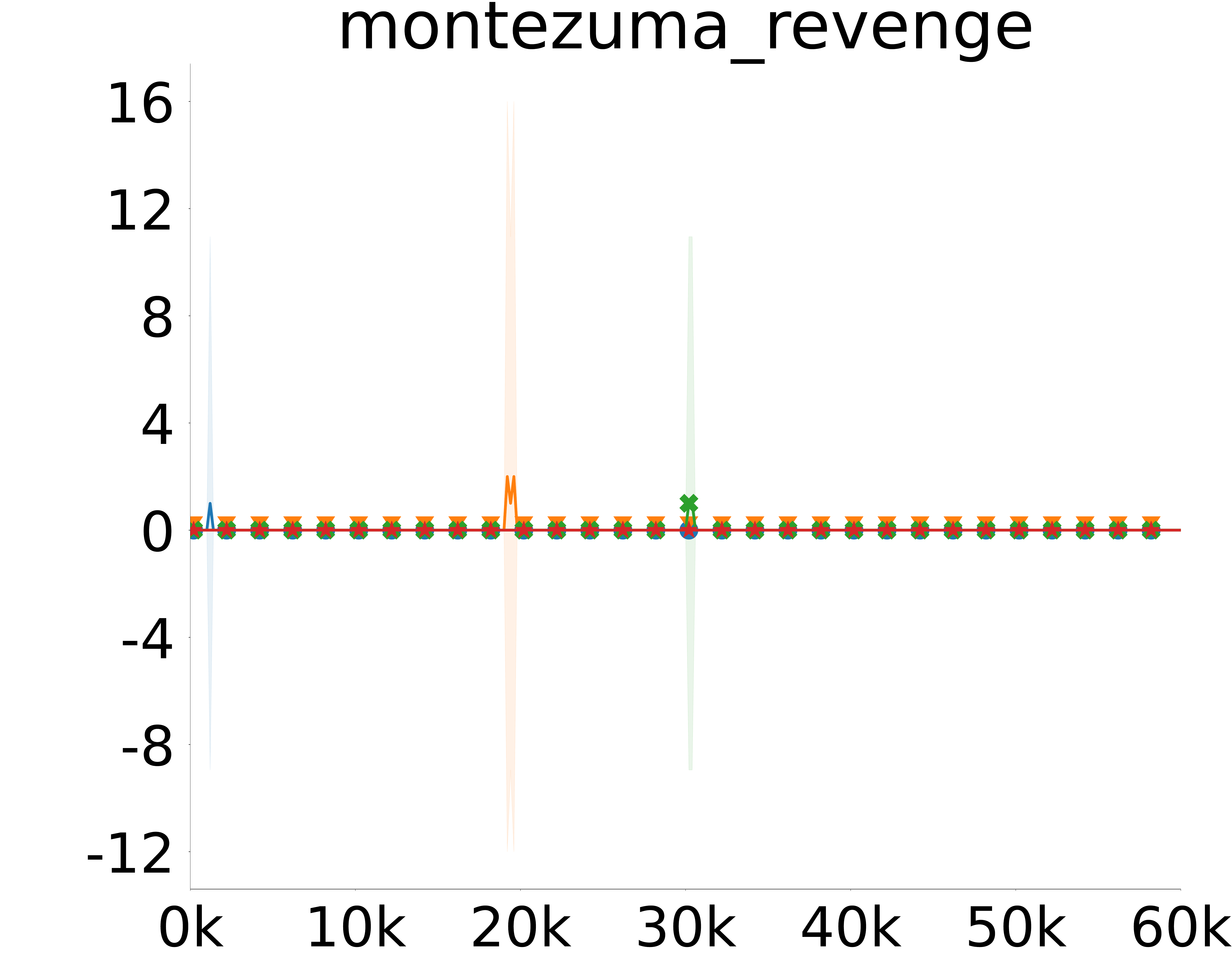}}
\subfloat{
    \includegraphics[width=0.15\textwidth]{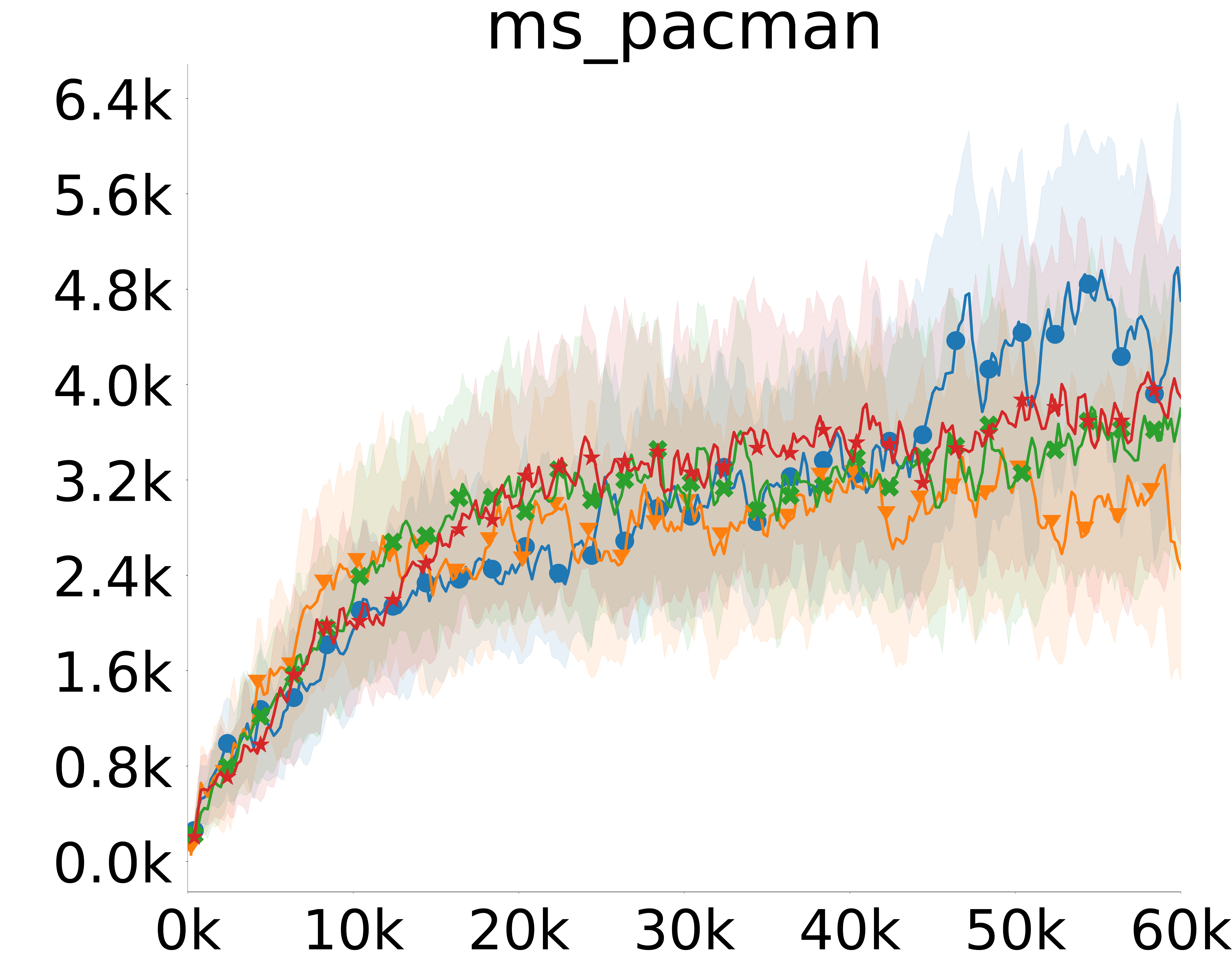}}
\subfloat{
    \includegraphics[width=0.15\textwidth]{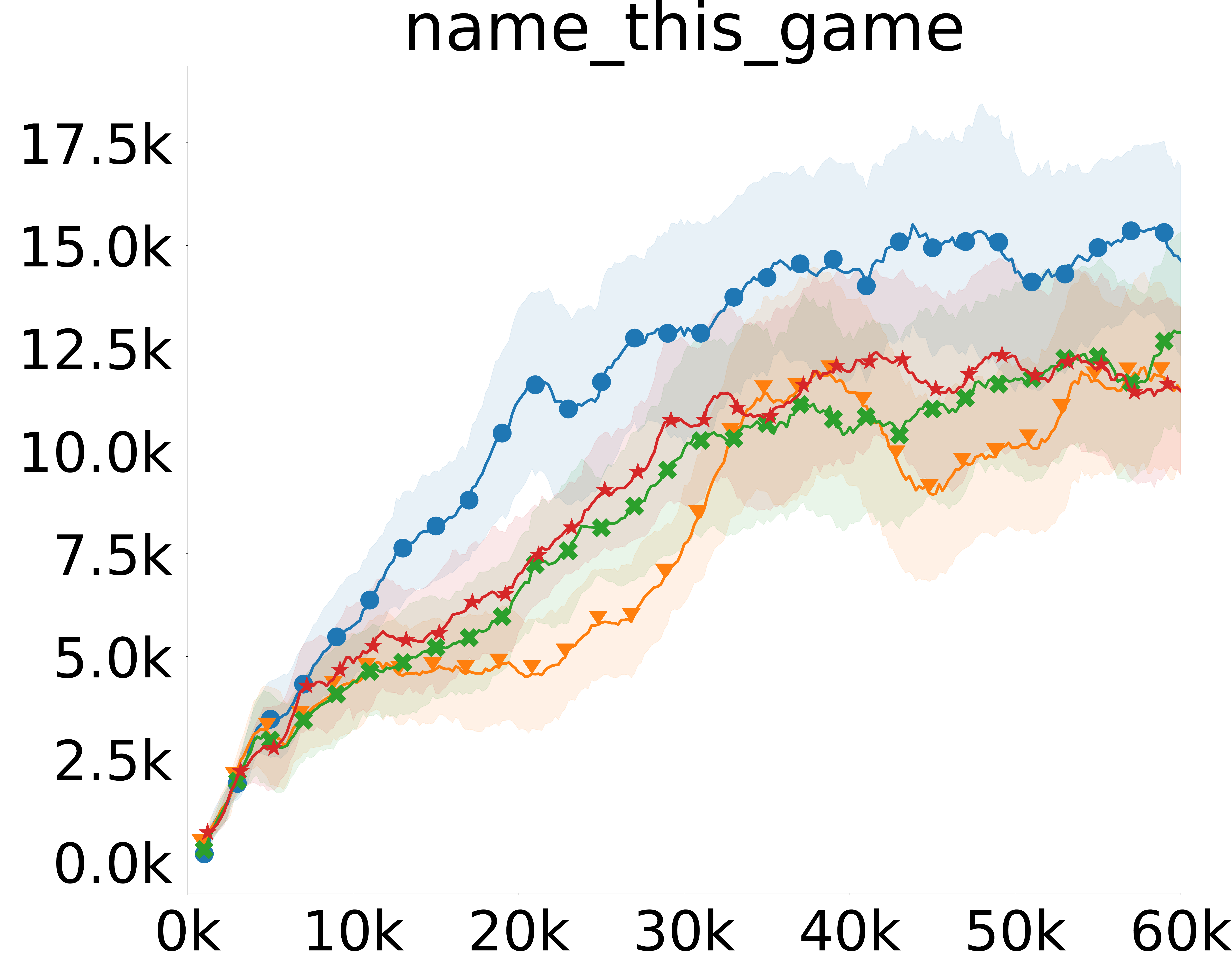}}
\subfloat{
    \includegraphics[width=0.15\textwidth]{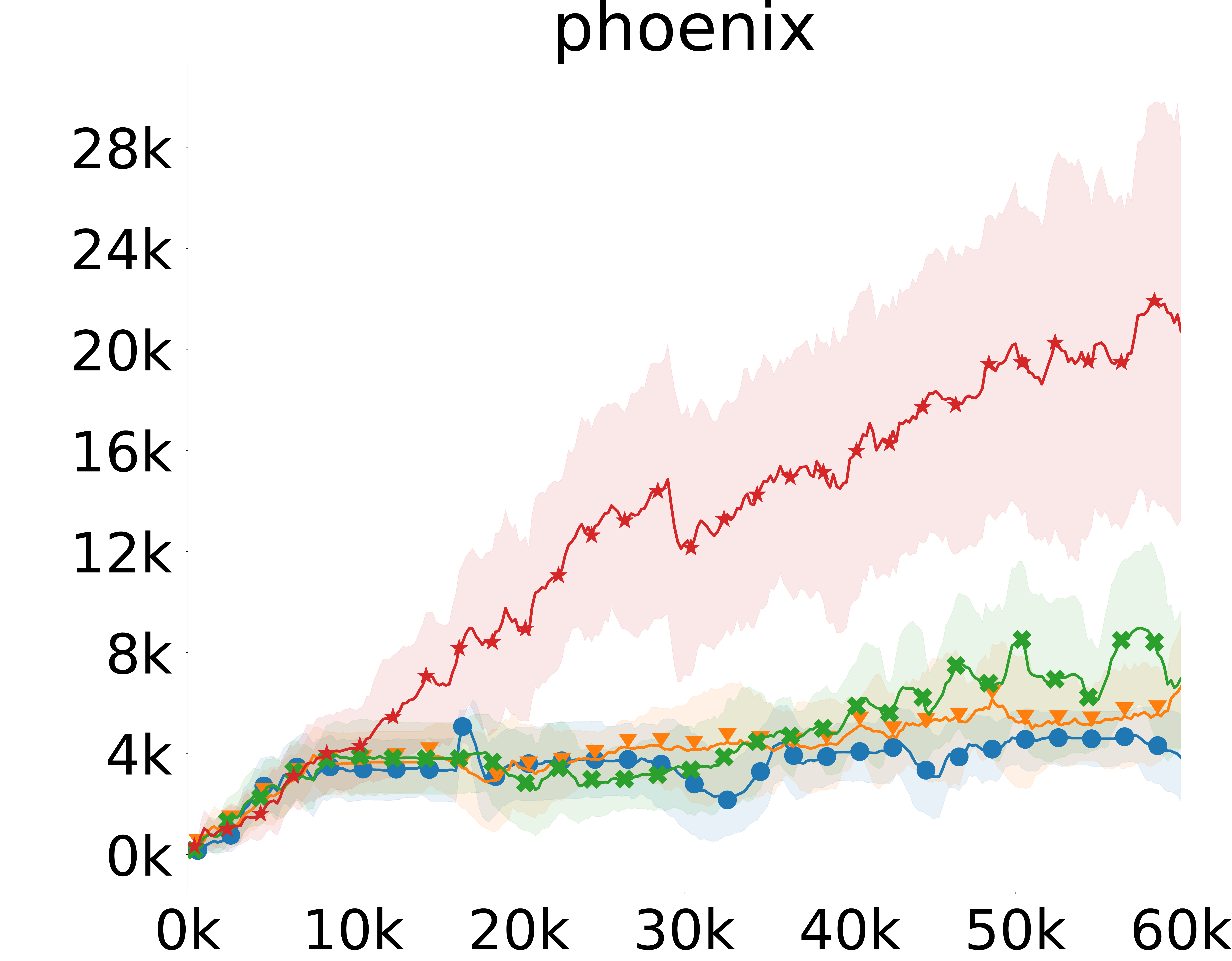}}
\subfloat{
    \includegraphics[width=0.15\textwidth]{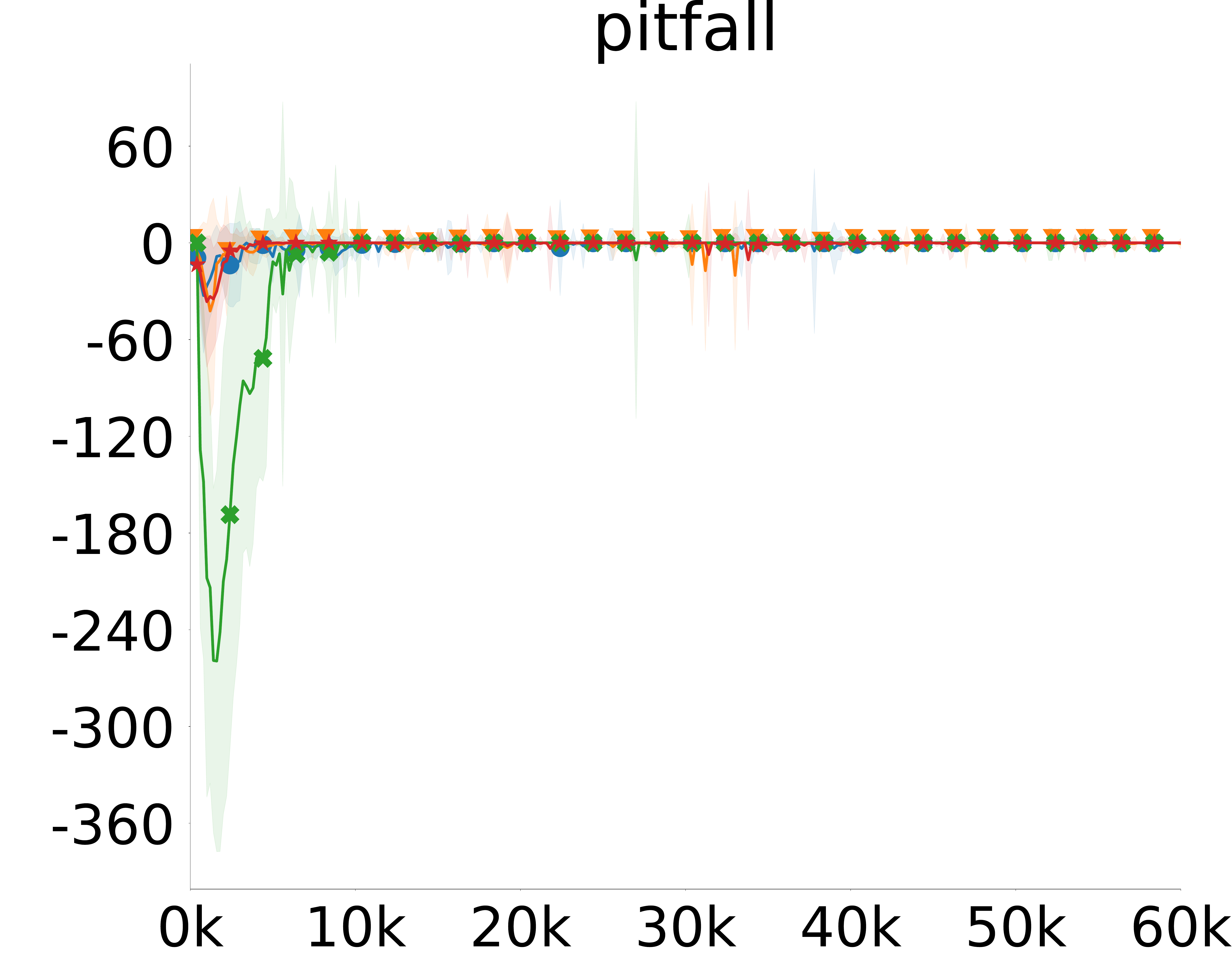}}
\\\vspace*{-0.8em}
\subfloat{
    \includegraphics[width=0.15\textwidth]{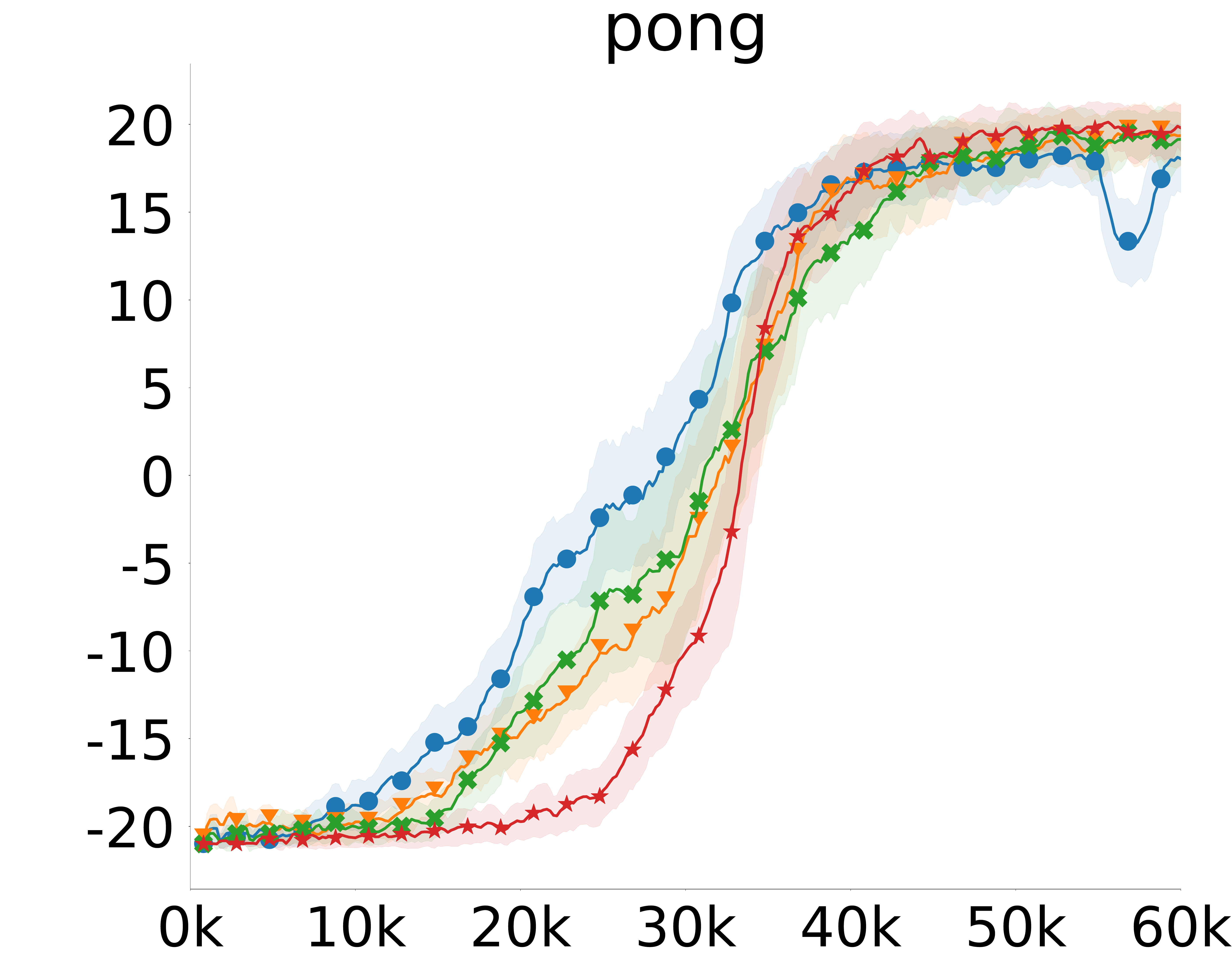}}
\subfloat{
    \includegraphics[width=0.15\textwidth]{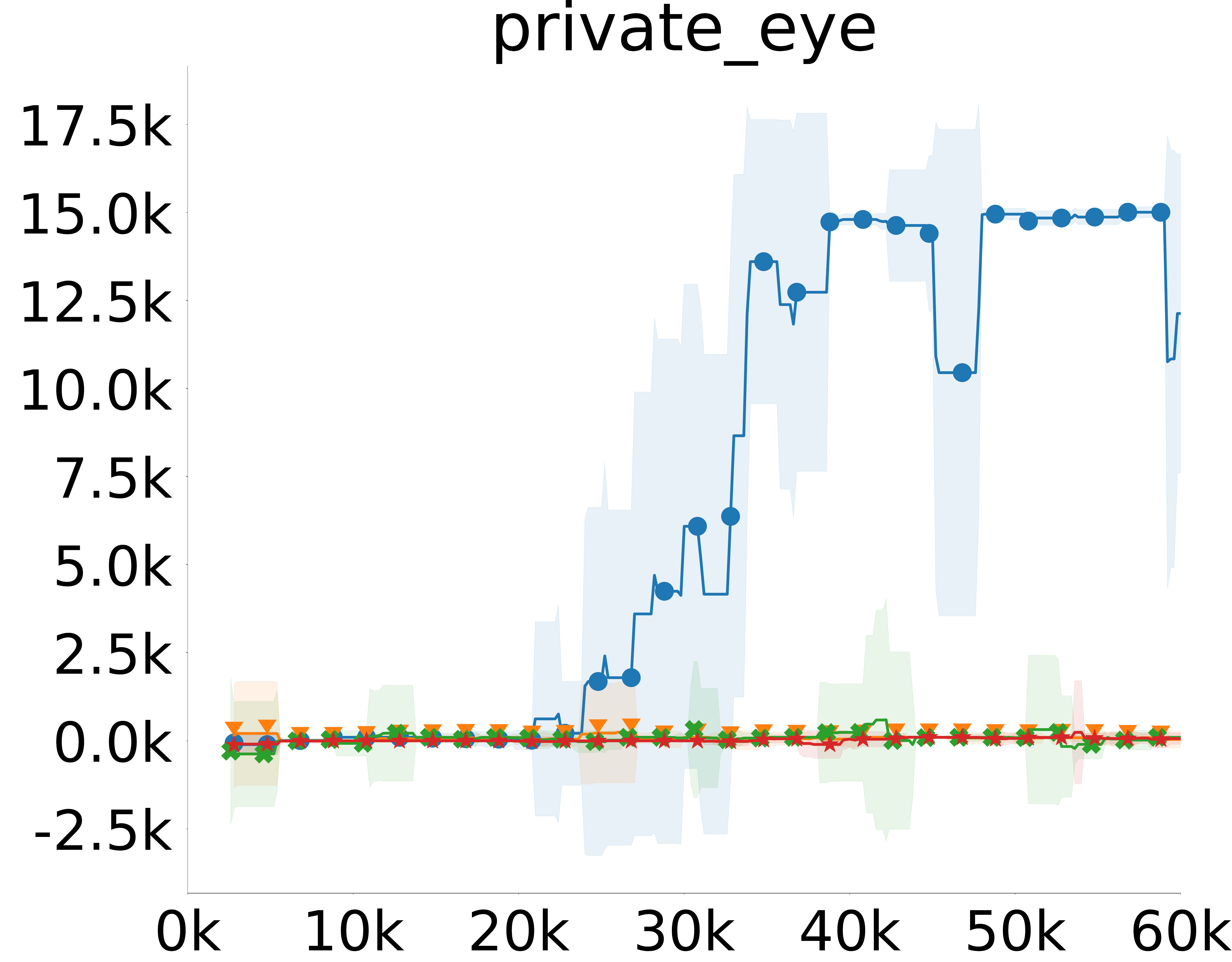}}
\subfloat{
    \includegraphics[width=0.15\textwidth]{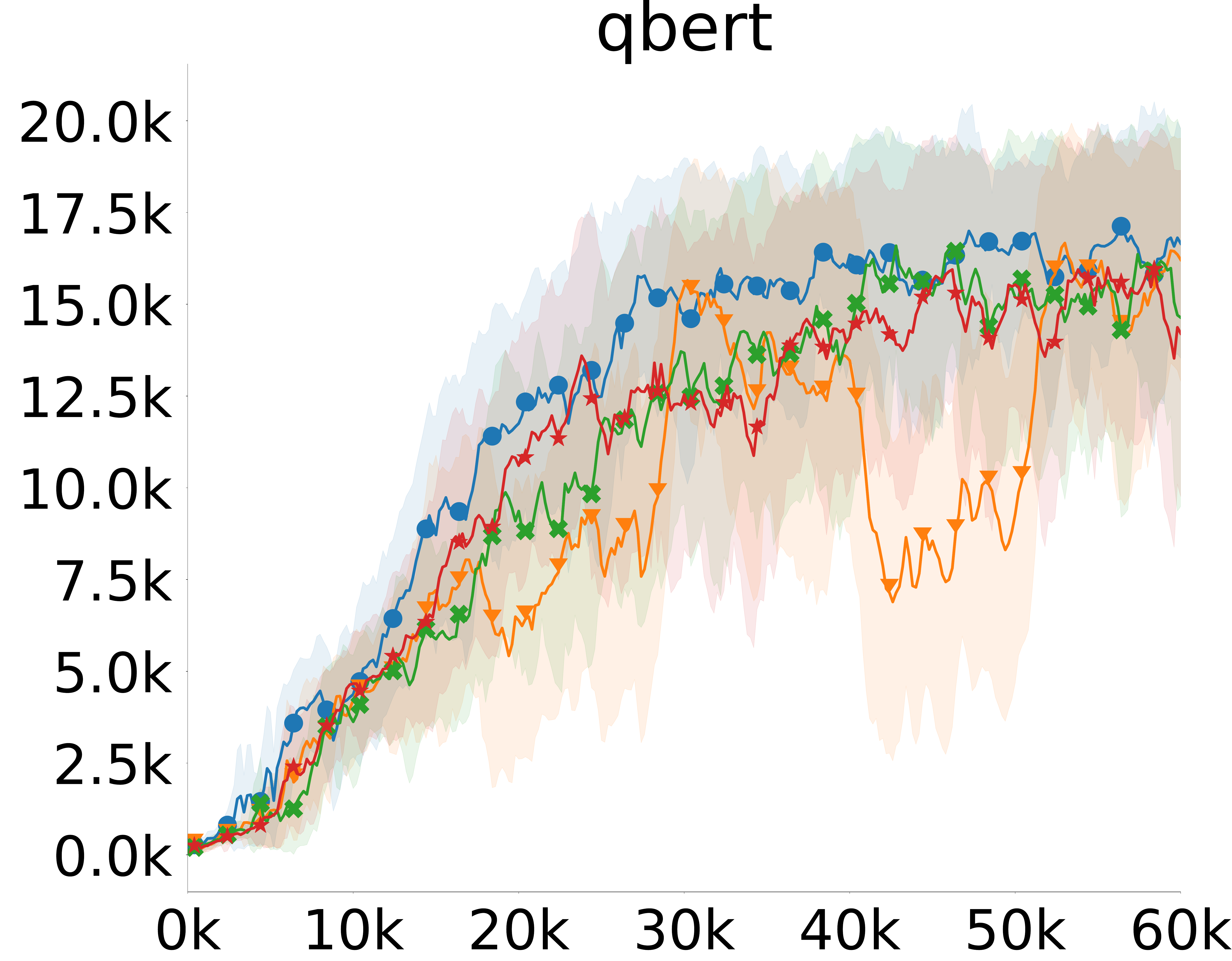}}
\subfloat{
    \includegraphics[width=0.15\textwidth]{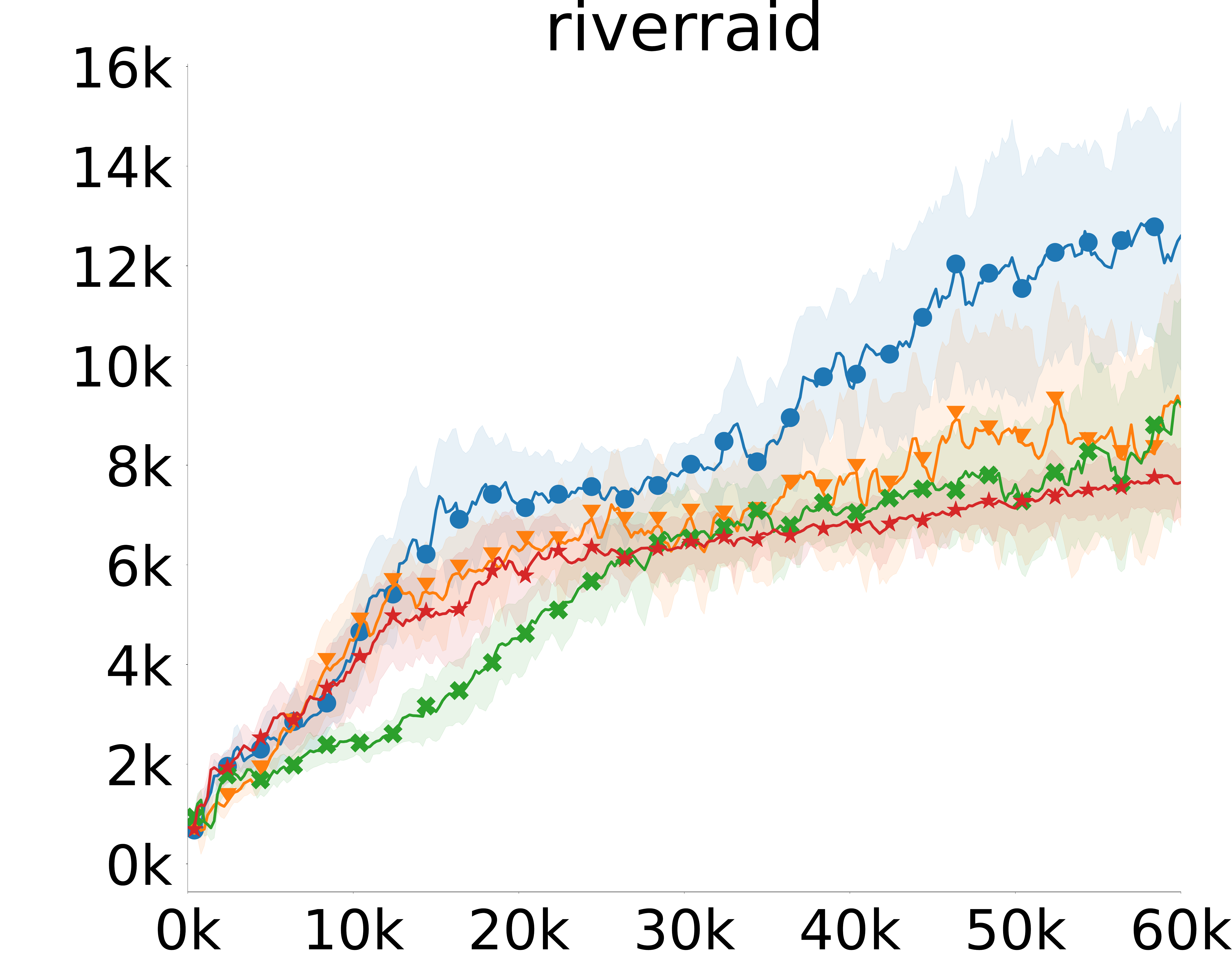}}
\subfloat{
    \includegraphics[width=0.15\textwidth]{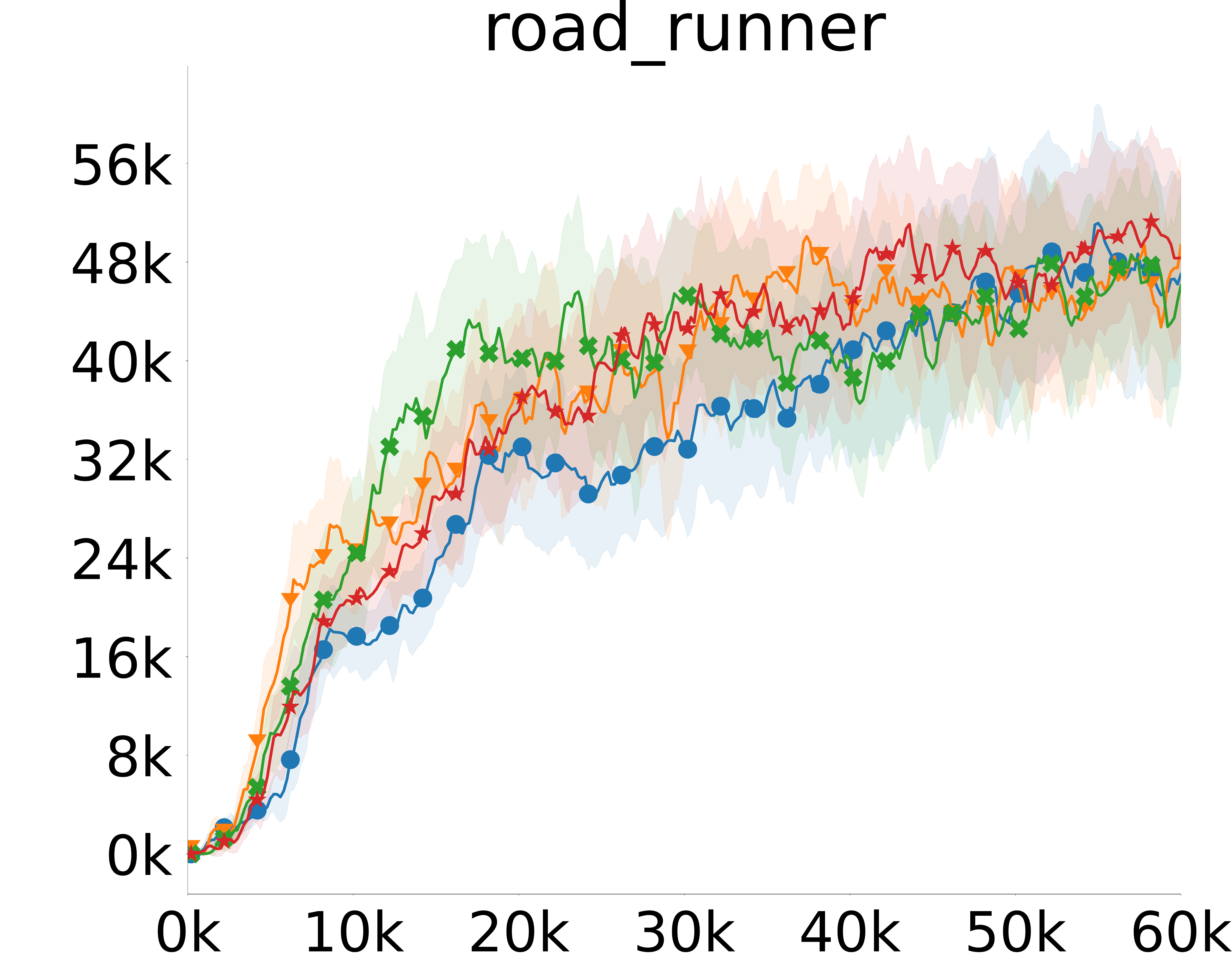}}
\subfloat{
    \includegraphics[width=0.15\textwidth]{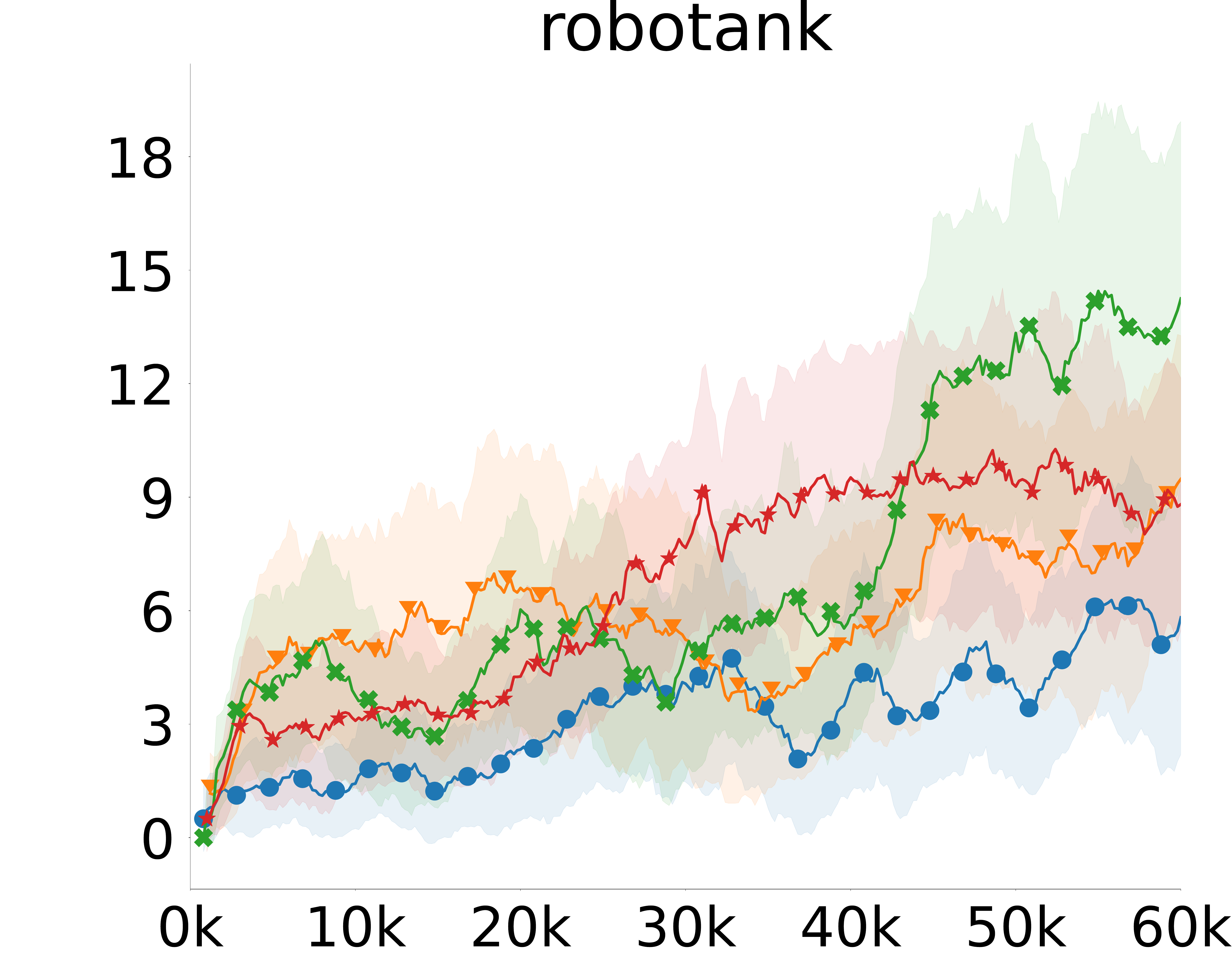}}
\\\vspace*{-0.8em}
\subfloat{
    \includegraphics[width=0.15\textwidth]{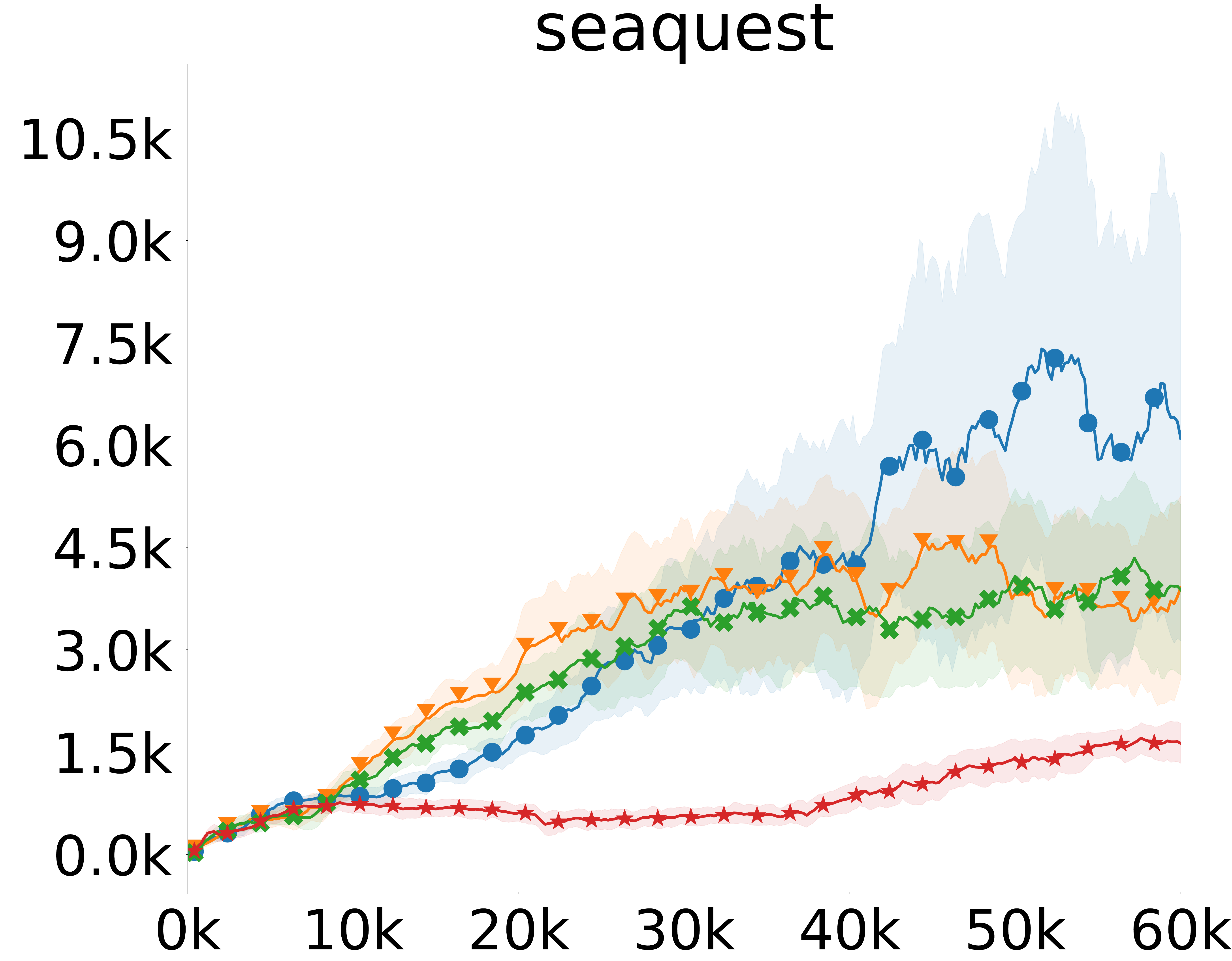}}
\subfloat{
    \includegraphics[width=0.15\textwidth]{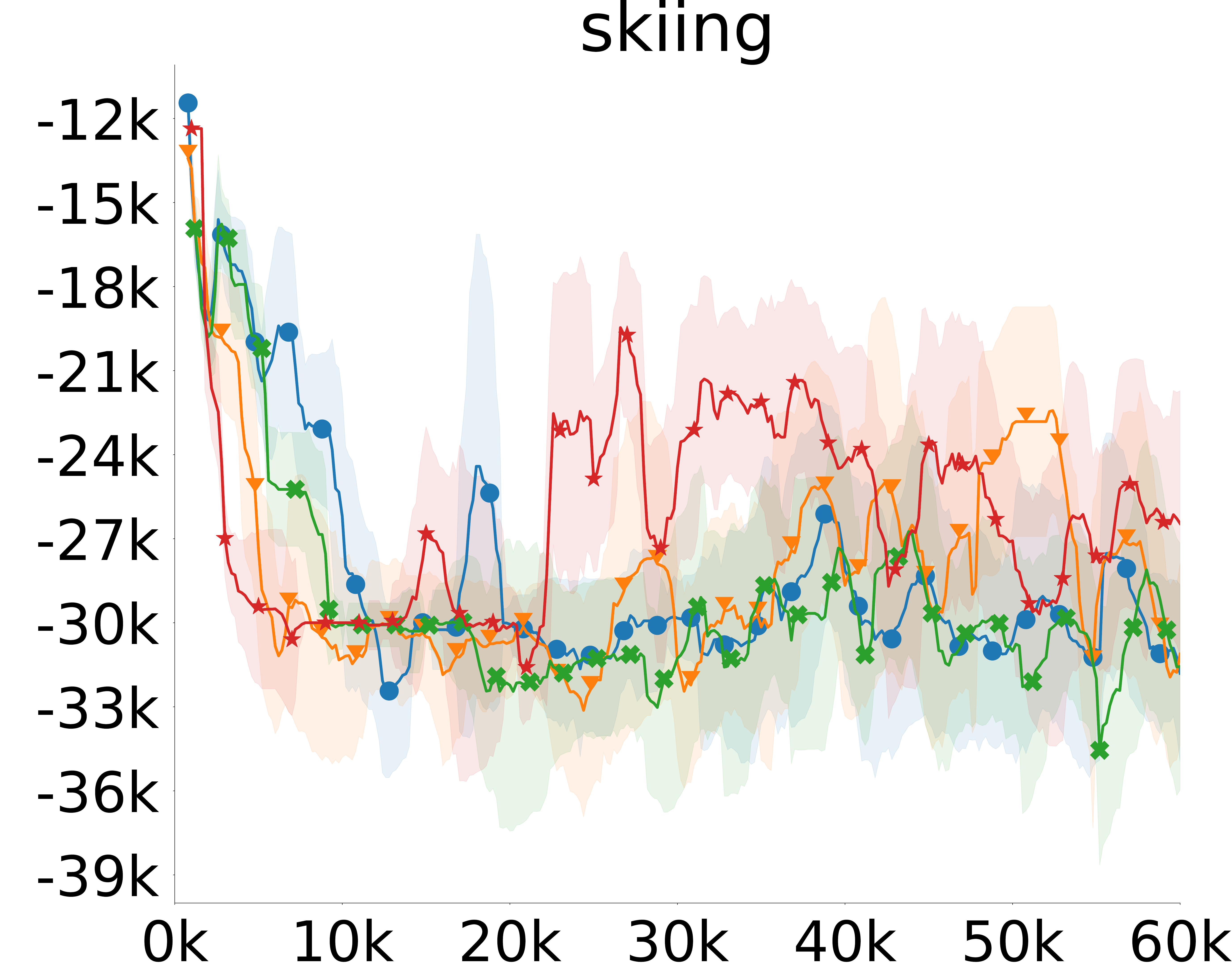}}
\subfloat{
    \includegraphics[width=0.15\textwidth]{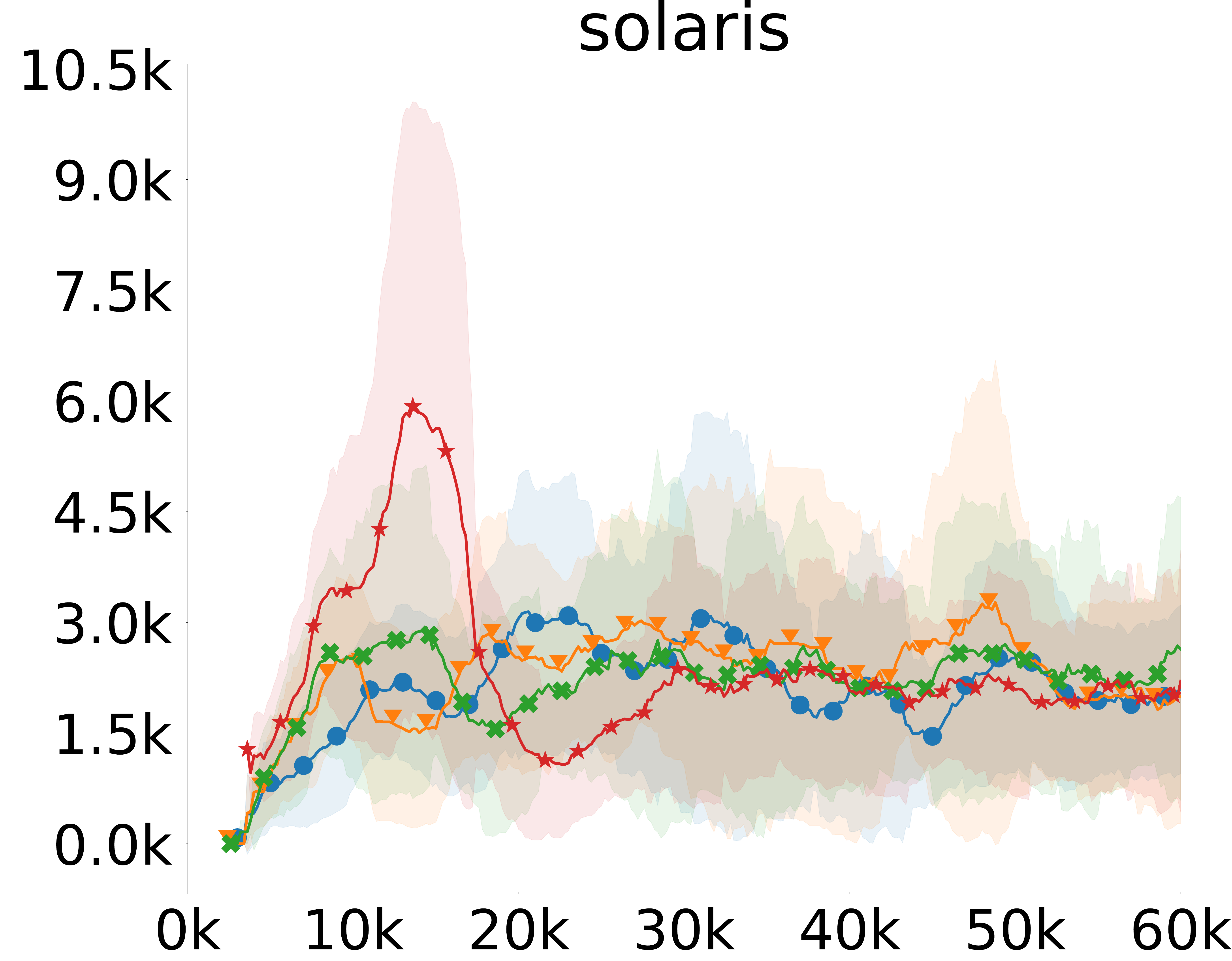}}
\subfloat{
    \includegraphics[width=0.15\textwidth]{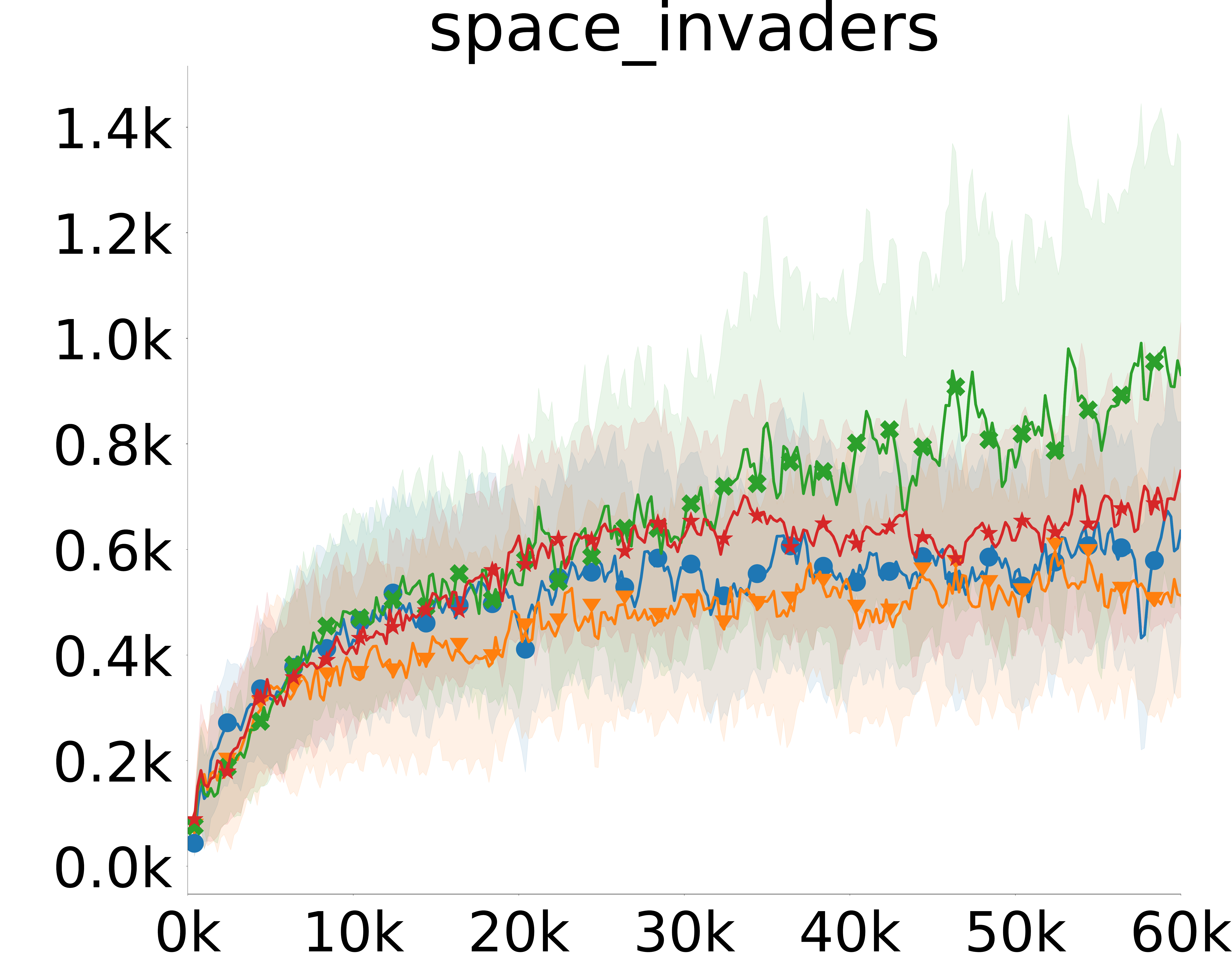}}
\subfloat{
    \includegraphics[width=0.15\textwidth]{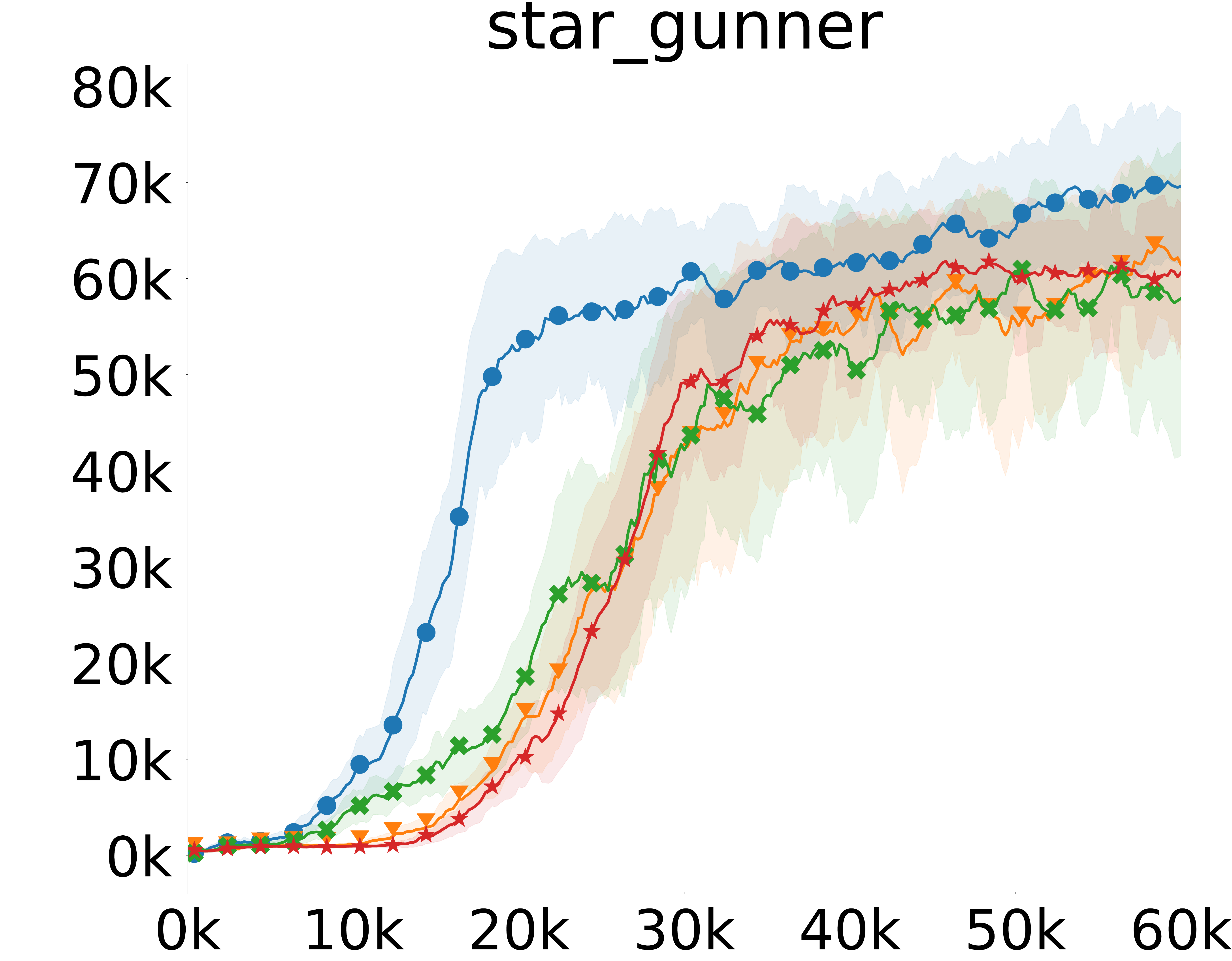}}
\subfloat{
    \includegraphics[width=0.15\textwidth]{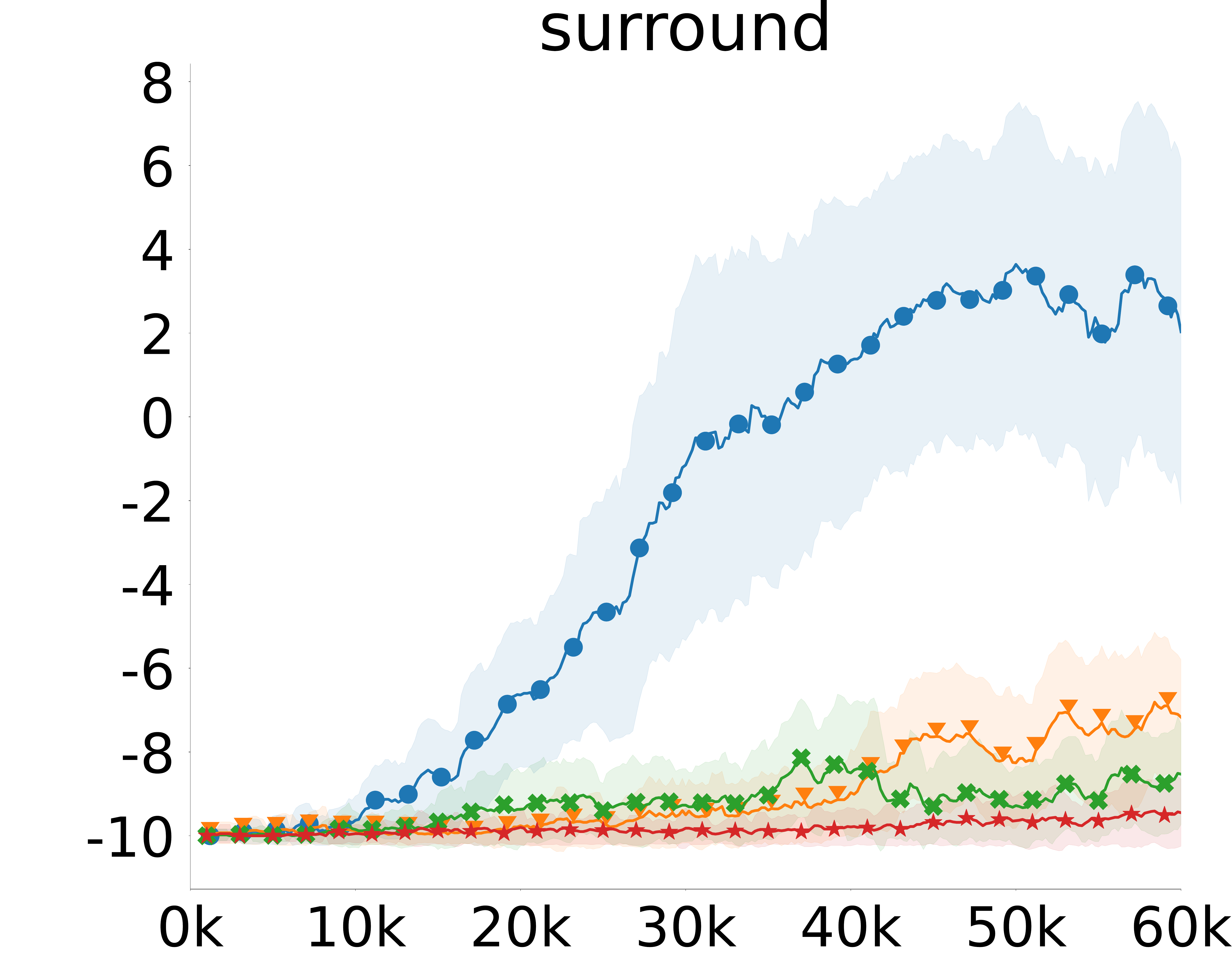}}
\\\vspace*{-0.8em}
\subfloat{
    \includegraphics[width=0.15\textwidth]{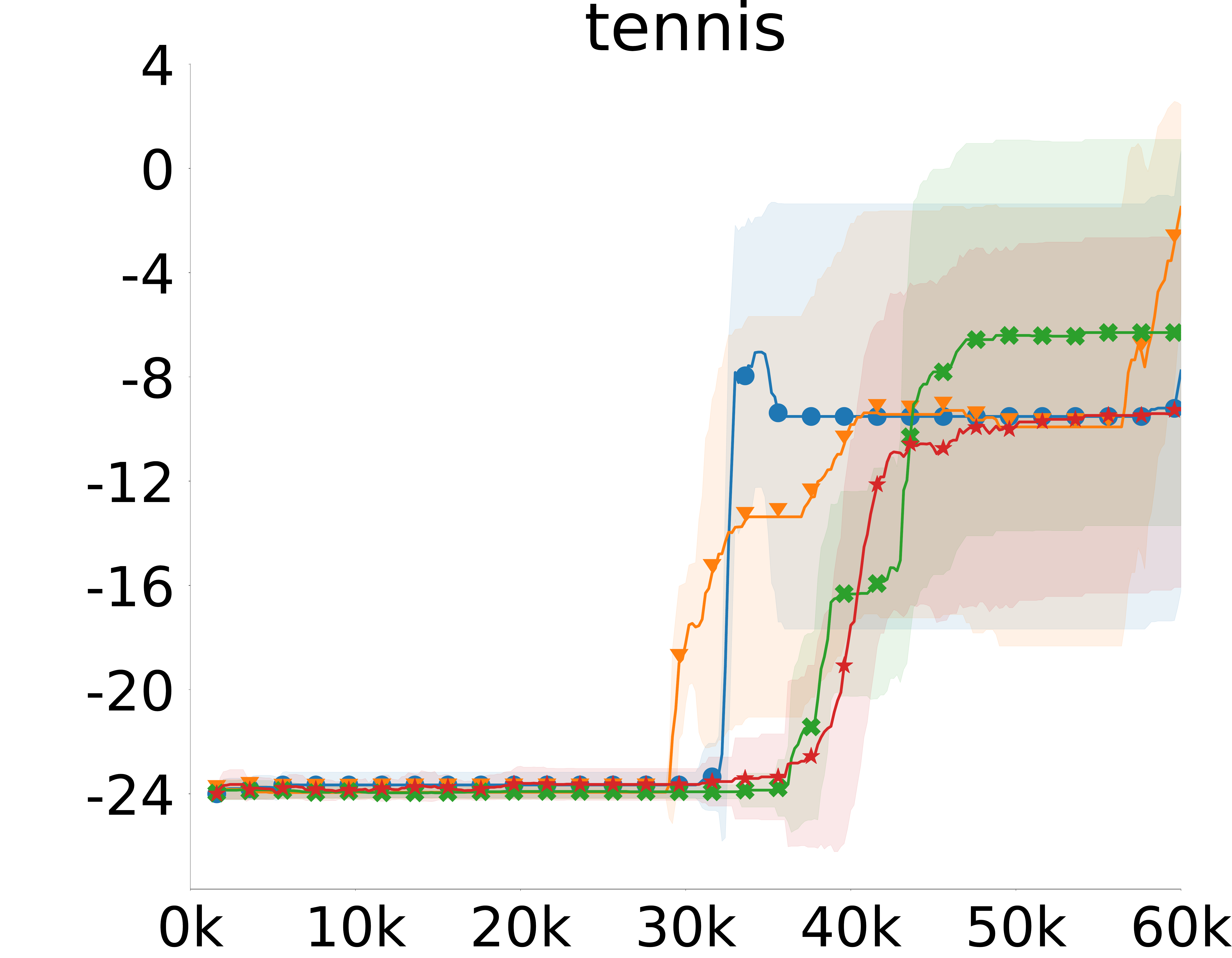}}
\subfloat{
    \includegraphics[width=0.15\textwidth]{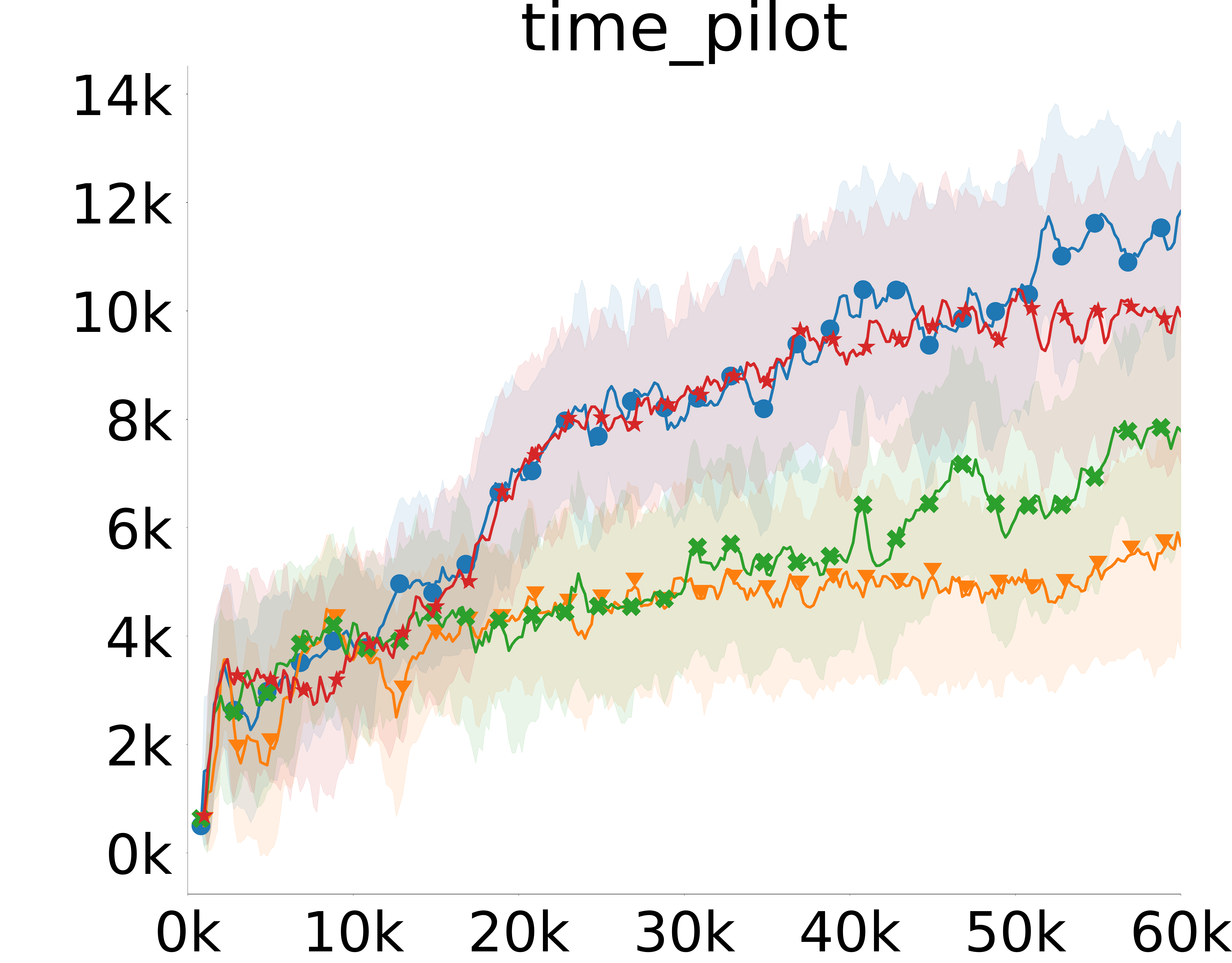}}
\subfloat{
    \includegraphics[width=0.15\textwidth]{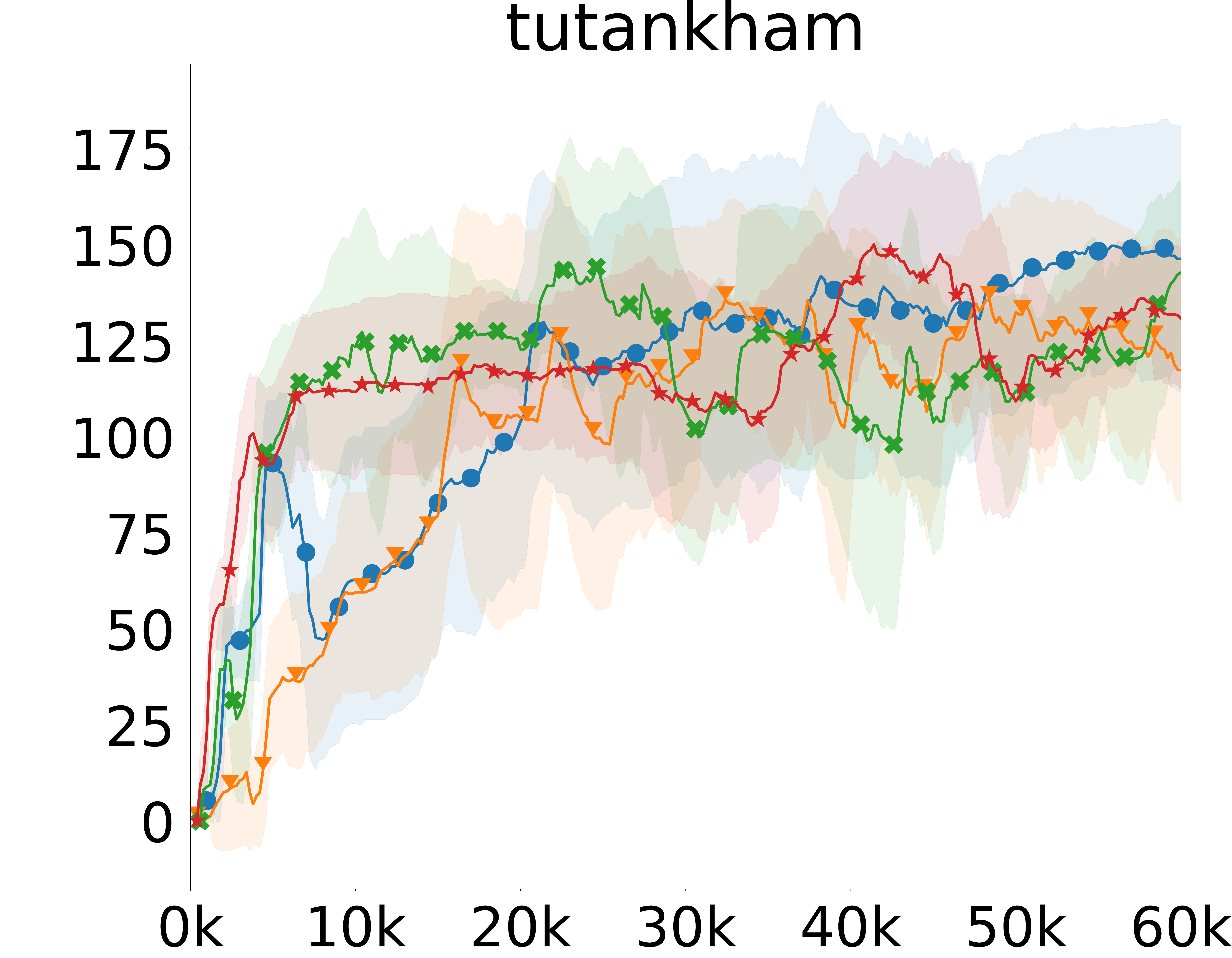}}
\subfloat{
    \includegraphics[width=0.15\textwidth]{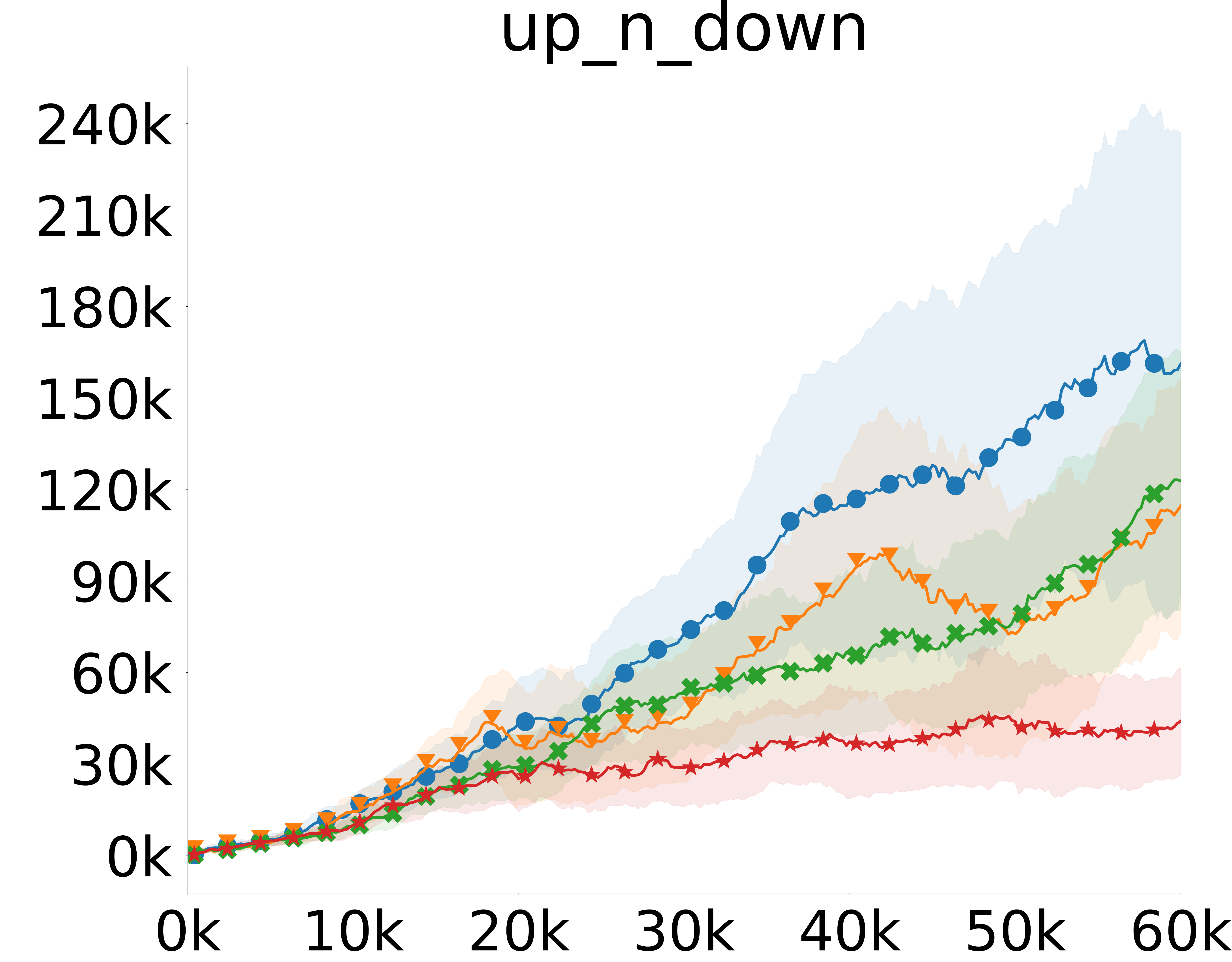}}
\subfloat{
    \includegraphics[width=0.15\textwidth]{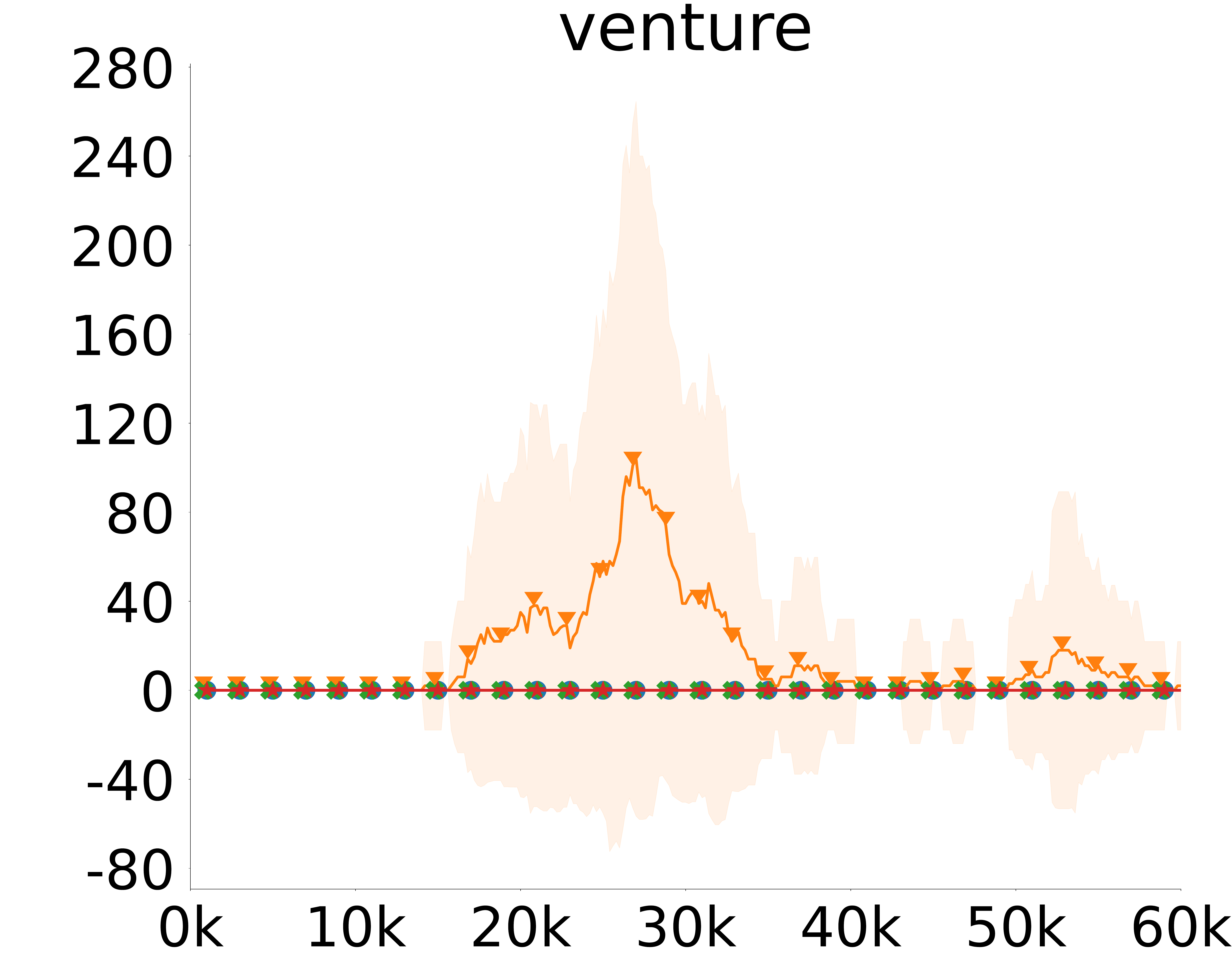}}
\subfloat{
    \includegraphics[width=0.15\textwidth]{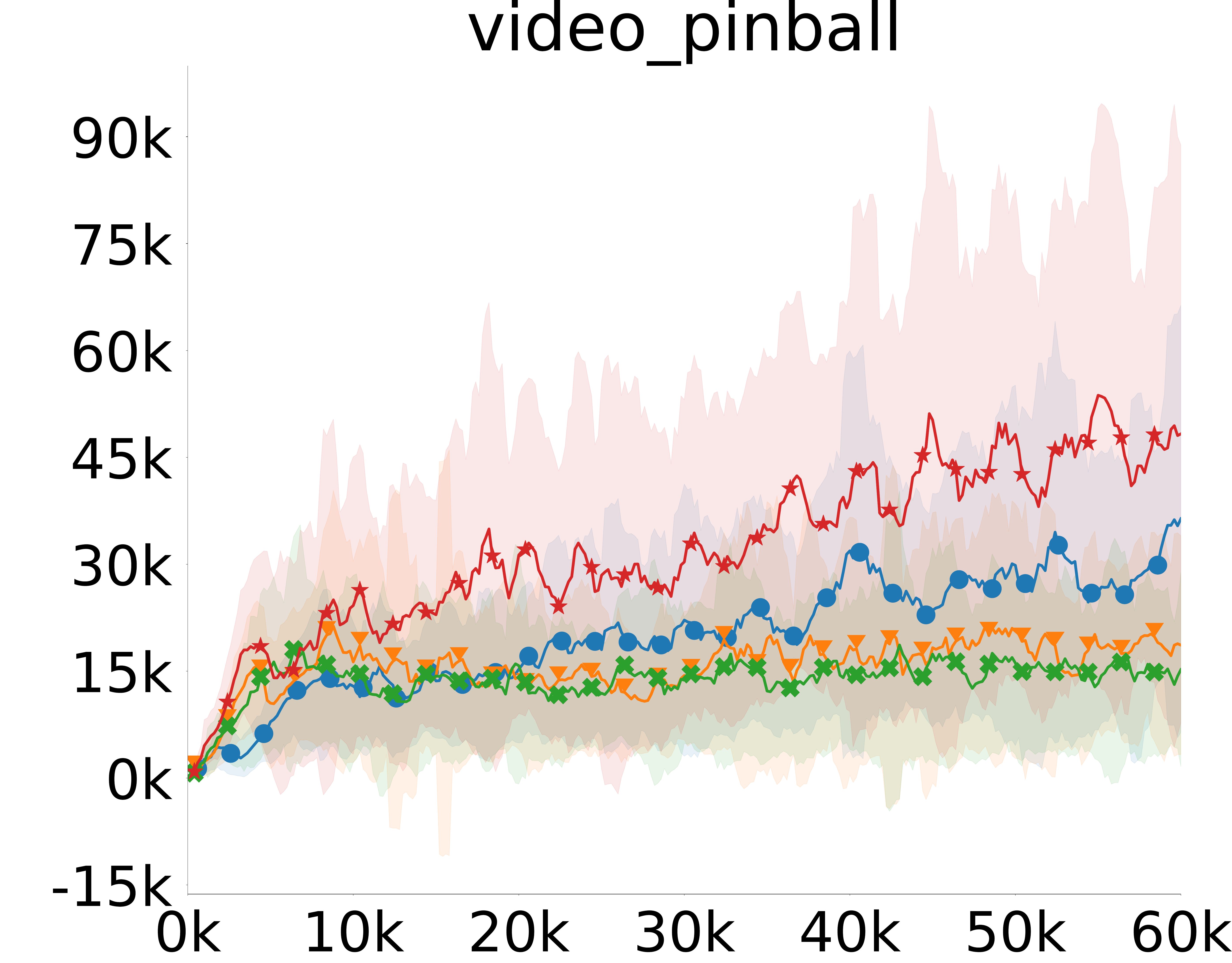}}
\\\vspace*{-0.8em}
\subfloat{
    \includegraphics[width=0.15\textwidth]{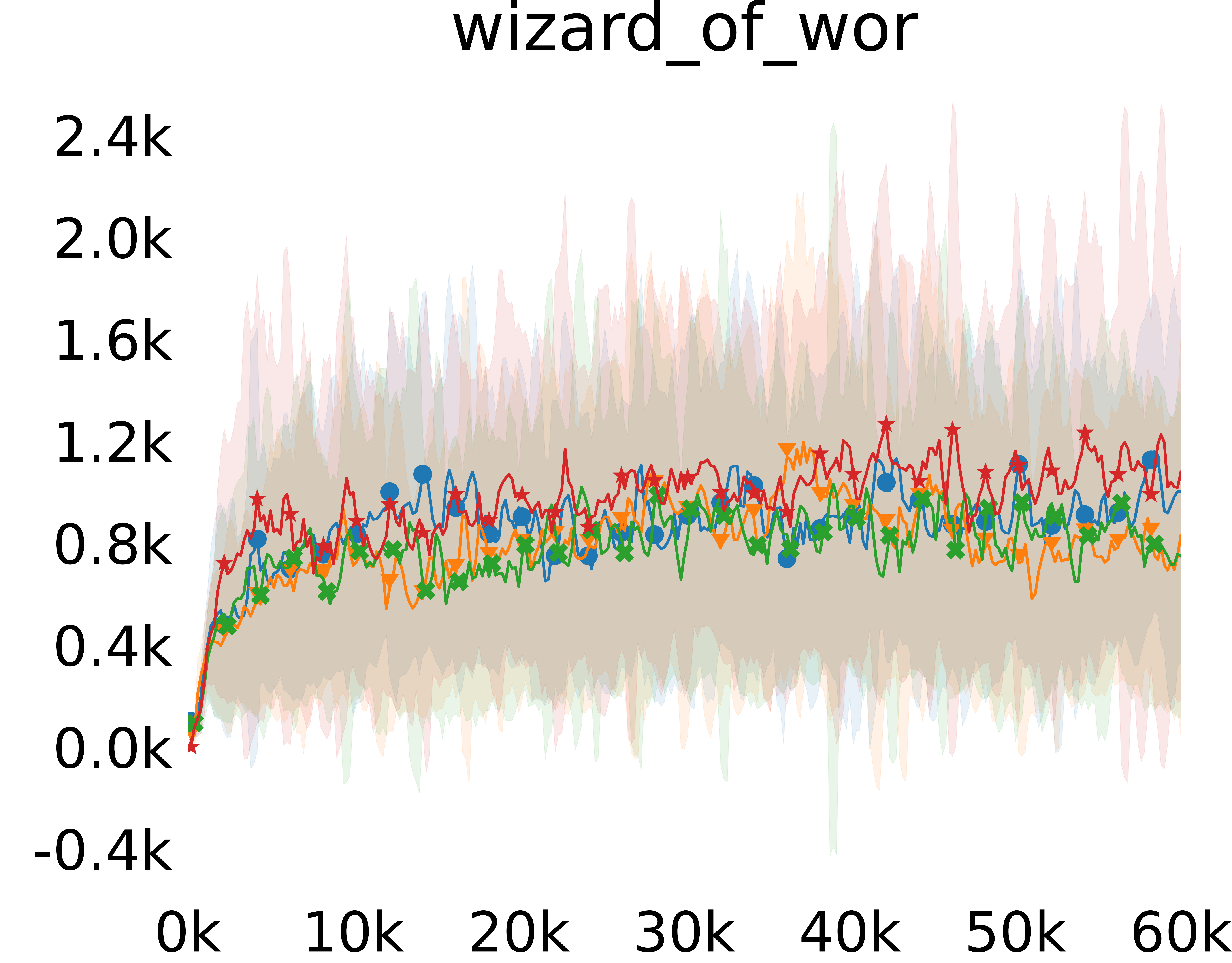}}
\subfloat{
    \includegraphics[width=0.15\textwidth]{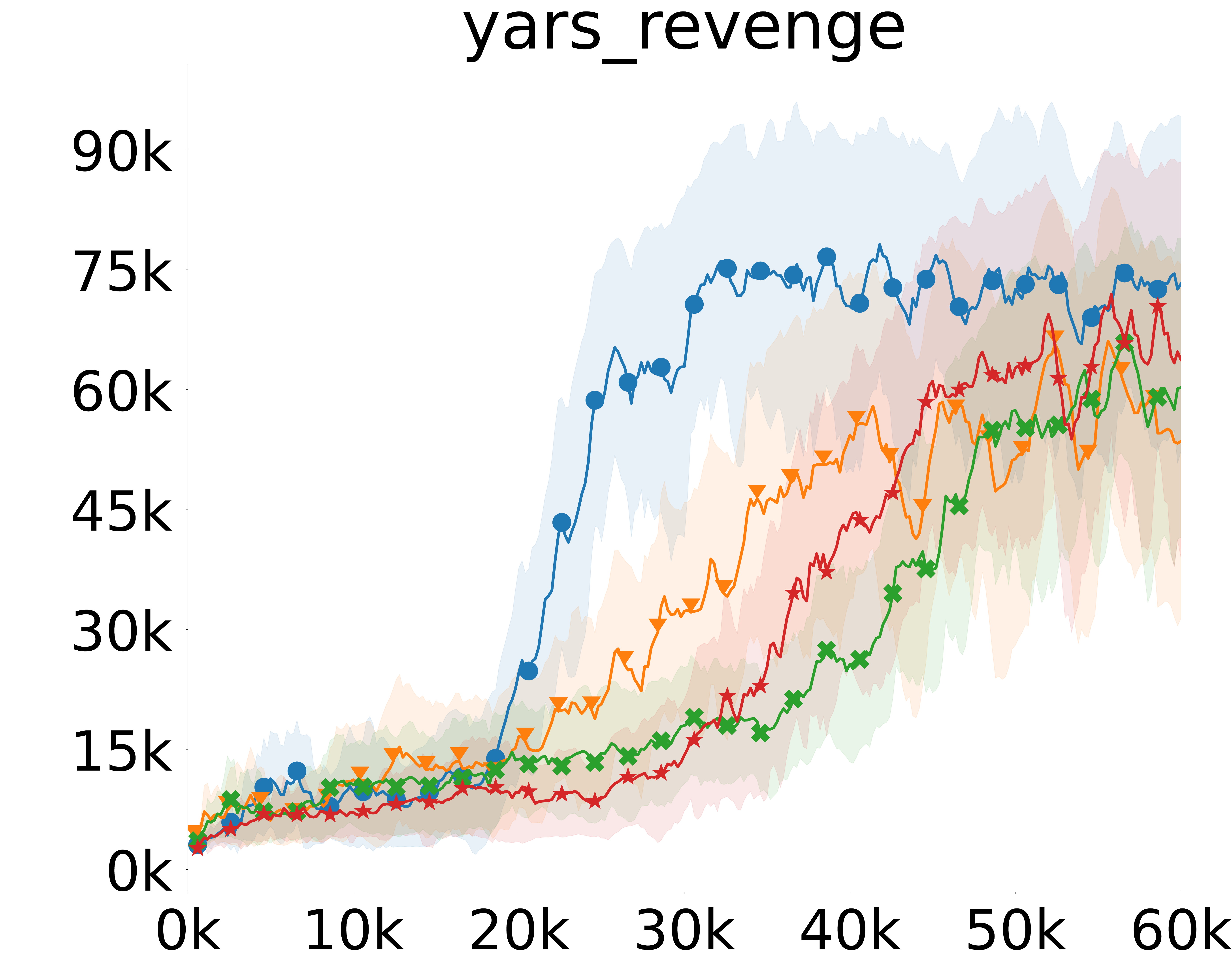}}
\subfloat{
    \includegraphics[width=0.15\textwidth]{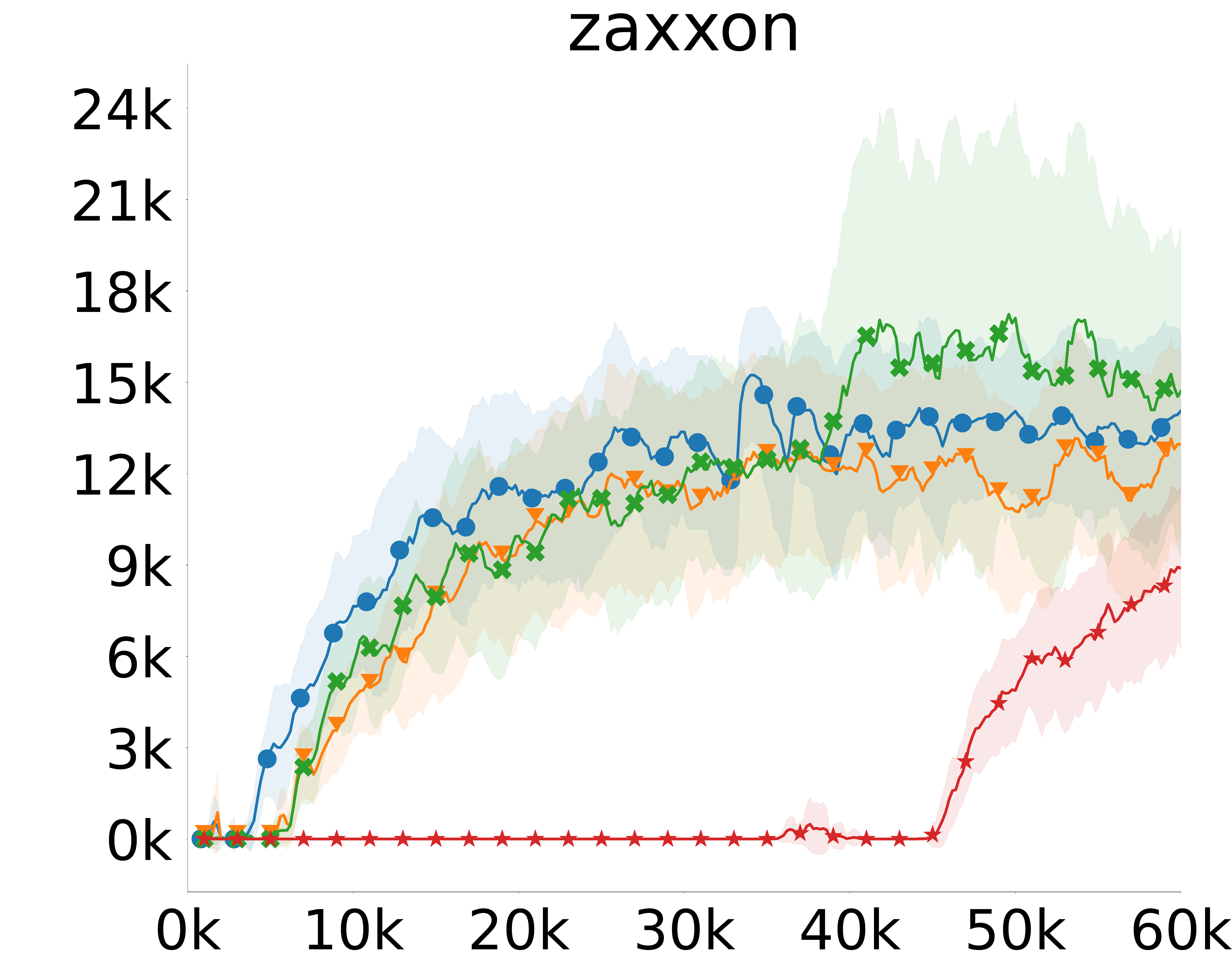}}
\subfloat{
    \includegraphics[width=0.15\textwidth]{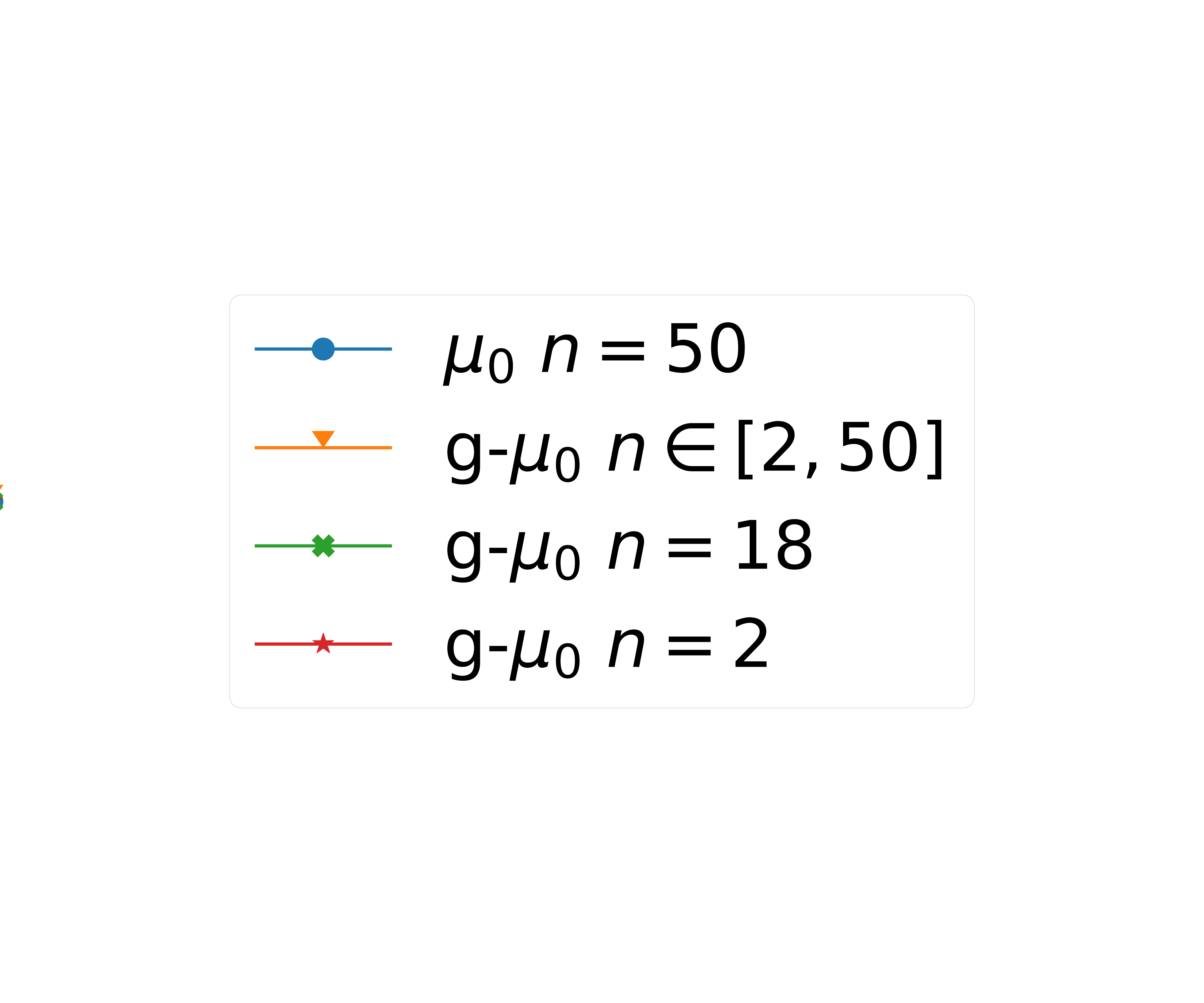}}
\caption{Training curves on 57 Atari games of different zero-knowledge learning algorithms. The x-axis represents the number of neural network training steps, while the y-axis represents the average returns collected from the last 100 training episodes. The shaded area indicates standard deviation.}
\label{fig:Atari57-score}
\end{figure*}

Next, we evaluate the progressive simulation method on 57 Atari games by training g-$\mu_0$ with a simulation boundary $(N_{min}, N_{max})=(2, 50)$, where the simulation budget is the same as $n=18$.
The results are shown in Table \ref{tab:Atari57-score} and Fig. \ref{fig:Atari57-score}.
From the table, we find that progressive simulation achieves lower human normalized mean returns (359.97\%) compared to the baseline model g-$\mu_0$ $n=18$ (395.87\%), which uses the same simulation budget.
Overall, progressive simulation only outperforms g-$\mu_0$ $n=18$ in 25 games among 57 games.

For cases that use progressive simulation significantly outperforms g-$\mu_0$ $n=18$, such as \textit{battle\_zone}, \textit{kangaroo}, and \textit{jamesbond}.
The average returns of these three games gradually increase during training.
This situation is similar to that of board games, where fewer simulations are allocated during the early stages of training.
Then, simulations are gradually increased to perform meticulous planning with higher simulation counts, obtaining better scores.

However, not all Atari games exhibit the same characteristics as board games in terms of increasing simulations.
For games such as \textit{gravitar}, training with fewer simulations outperforms training with more simulations.
Thus, using progressive simulation causes training to become less efficient as it progresses.
Specifically, it performs well in the early training stages but converges when the number of simulations becomes excessive.
For games such as \textit{asterix}, \textit{private\_eye}, and \textit{surround}, the opposite is that training requires sufficient simulations.
Namely, the use of progressive simulation results in wasted early stages of training.

Furthermore, we observe that several games\footnote{\textit{amidar}, \textit{berzerk}, \textit{breakout}, \textit{centipede}, \textit{crazy\_climber}, \textit{qbert}, and \textit{robotank}.} experience a significant decline in score when the simulation counts are changed.
Despite this decline, it usually recovers after several training steps.
We hypothesize that the agent might settle into a sub-optimal strategy when training with a fixed simulation.
As the number of simulations increases, the strategy shifts into different playing styles, which may impact the performance both positively and negatively, before long-term improvements.
Therefore, the model may require a significantly larger number of training steps to accommodate these shifts and subsequent adjustments.

In conclusion, the ablation study on the progressive simulation method suggests that adjusting the simulation count during training does not necessarily lead to improvement in Atari games.
Since each Atari game has distinct characteristics, the results vary, as demonstrated in section \ref{exp:atari}.
A universally beneficial method for training Atari games is yet to be discovered.

\section{Conclusion}
This paper introduces \textit{MiniZero}, a zero-knowledge learning framework that supports four algorithms, including AlphaZero, MuZero, Gumbel AlphaZero, and Gumbel MuZero.
We compare the performance of these algorithms with different simulations on 9x9 Go, 8x8 Othello, and 57 Atari games.

For board games, given the same amount of training data, using more simulations generally results in higher performance for both AlphaZero and MuZero algorithms.
Nevertheless, when trained for an equivalent amount of time (computing resources), Gumbel Zero with fewer simulations can achieve performance nearly on par with AlphaZero/MuZero using more simulations.
We also propose an efficient approach, named \textit{progressive simulation}, to reallocate the simulation budget under limited computing resources.
This approach begins with fewer simulations, which is gradually increased throughout training.
Our experiments demonstrate that models trained with progressive simulation outperform those trained without it.
Furthermore, the choice between AlphaZero and MuZero may differ according to specific game properties.
Our experiments indicate that MuZero excels in 9x9 Go, whereas AlphaZero is superior in 8x8 Othello.

The results for Atari games are different from board games.
While training with 50 simulations generally yields better results than 18 and 2 simulations in terms of human normalized mean returns, it is noteworthy that using 50 simulations does not consistently outperform 18 and 2 simulations in every game.
Our experiments suggest that the performance of different simulation settings is highly correlated to distinct game characteristics.
Therefore, the progressive simulation method may not be necessarily optimal for all cases.

Currently, our framework only supports four fundamental zero-knowledge learning algorithms.
The framework can still be improved with several extensions, such as incorporating Reanalyze techniques \cite{schrittwieser_mastering_2020}, providing continuous action spaces \cite{antonoglou_planning_2021}, supporting stochastic environments \cite{hubert_learning_2021}, and including more games. 
Besides, we can design different progressive simulation strategies, such as adjusting simulations during training without a pre-determined total budget.
In conclusion, our framework, along with all trained models, is publicly available online.
We expect that these trained models can serve as benchmarks for the game community in the future.

\bibliography{references}

\begin{thebibliography}{10}
\providecommand{\url}[1]{#1}
\csname url@samestyle\endcsname
\providecommand{\newblock}{\relax}
\providecommand{\bibinfo}[2]{#2}
\providecommand{\BIBentrySTDinterwordspacing}{\spaceskip=0pt\relax}
\providecommand{\BIBentryALTinterwordstretchfactor}{4}
\providecommand{\BIBentryALTinterwordspacing}{\spaceskip=\fontdimen2\font plus
\BIBentryALTinterwordstretchfactor\fontdimen3\font minus \fontdimen4\font\relax}
\providecommand{\BIBforeignlanguage}[2]{{%
\expandafter\ifx\csname l@#1\endcsname\relax
\typeout{** WARNING: IEEEtran.bst: No hyphenation pattern has been}%
\typeout{** loaded for the language `#1'. Using the pattern for}%
\typeout{** the default language instead.}%
\else
\language=\csname l@#1\endcsname
\fi
#2}}
\providecommand{\BIBdecl}{\relax}
\BIBdecl

\bibitem{silver_mastering_2016}
D.~Silver, A.~Huang, C.~J. Maddison, A.~Guez, L.~Sifre, G.~{van den Driessche}, J.~Schrittwieser, I.~Antonoglou, V.~Panneershelvam, M.~Lanctot, S.~Dieleman, D.~Grewe, J.~Nham, N.~Kalchbrenner, I.~Sutskever, T.~Lillicrap, M.~Leach, K.~Kavukcuoglu, T.~Graepel, and D.~Hassabis, ``Mastering the game of {{Go}} with deep neural networks and tree search,'' \emph{Nature}, vol. 529, no. 7587, pp. 484--489, Jan. 2016.

\bibitem{silver_mastering_2017}
D.~Silver, J.~Schrittwieser, K.~Simonyan, I.~Antonoglou, A.~Huang, A.~Guez, T.~Hubert, L.~Baker, M.~Lai, A.~Bolton, Y.~Chen, T.~Lillicrap, F.~Hui, L.~Sifre, G.~{van den Driessche}, T.~Graepel, and D.~Hassabis, ``Mastering the game of {{Go}} without human knowledge,'' \emph{Nature}, vol. 550, no. 7676, pp. 354--359, Oct. 2017.

\bibitem{silver_general_2018}
D.~Silver, T.~Hubert, J.~Schrittwieser, I.~Antonoglou, M.~Lai, A.~Guez, M.~Lanctot, L.~Sifre, D.~Kumaran, T.~Graepel, T.~Lillicrap, K.~Simonyan, and D.~Hassabis, ``A general reinforcement learning algorithm that masters chess, shogi, and {{Go}} through self-play,'' \emph{Science}, vol. 362, no. 6419, pp. 1140--1144, Dec. 2018.

\bibitem{schrittwieser_mastering_2020}
J.~Schrittwieser, I.~Antonoglou, T.~Hubert, K.~Simonyan, L.~Sifre, S.~Schmitt, A.~Guez, E.~Lockhart, D.~Hassabis, T.~Graepel, T.~Lillicrap, and D.~Silver, ``Mastering {{Atari}}, {{Go}}, chess and shogi by planning with a learned model,'' \emph{Nature}, vol. 588, no. 7839, pp. 604--609, Dec. 2020.

\bibitem{danihelka_policy_2022}
I.~Danihelka, A.~Guez, J.~Schrittwieser, and D.~Silver, ``Policy improvement by planning with {{Gumbel}},'' in \emph{International {{Conference}} on {{Learning Representations}}}, Apr. 2022.

\bibitem{browne_survey_2012}
C.~B. Browne, E.~Powley, D.~Whitehouse, S.~M. Lucas, P.~I. Cowling, P.~Rohlfshagen, S.~Tavener, D.~Perez, S.~Samothrakis, and S.~Colton, ``A {{Survey}} of {{Monte Carlo Tree Search Methods}},'' \emph{IEEE Transactions on Computational Intelligence and AI in Games}, vol.~4, no.~1, pp. 1--43, Mar. 2012.

\bibitem{coulom_efficient_2007}
R.~Coulom, ``Efficient {{Selectivity}} and {{Backup Operators}} in {{Monte-Carlo Tree Search}},'' in \emph{Computers and {{Games}}}, ser. Lecture {{Notes}} in {{Computer Science}}.\hskip 1em plus 0.5em minus 0.4em\relax {Berlin, Heidelberg}: {Springer}, 2007, pp. 72--83.

\bibitem{kocsis_bandit_2006}
L.~Kocsis and C.~Szepesv{\'a}ri, ``Bandit {{Based Monte-Carlo Planning}},'' in \emph{Machine {{Learning}}: {{ECML}} 2006}, ser. Lecture {{Notes}} in {{Computer Science}}.\hskip 1em plus 0.5em minus 0.4em\relax {Berlin, Heidelberg}: {Springer}, 2006, pp. 282--293.

\bibitem{wu_accelerating_2020a}
D.~J. Wu, ``Accelerating {{Self-Play Learning}} in {{Go}},'' in \emph{Proceedings of the {{AAAI Workshop}} on {{Reinforcement Learning}} in {{Games}}}, Nov. 2020.

\bibitem{pascutto_leela_2023}
\BIBentryALTinterwordspacing
G.-C. Pascutto and {GitHub contributors}, ``Leela {{Zero}}: {{A Go}} program with no human provided knowledge,'' Leela Zero, Oct. 2023. [Online]. Available: \url{https://github.com/leela-zero/leela-zero}
\BIBentrySTDinterwordspacing

\bibitem{tian_elf_2019}
Y.~Tian, J.~Ma, Q.~Gong, S.~Sengupta, Z.~Chen, J.~Pinkerton, and L.~Zitnick, ``{{ELF OpenGo}}: An analysis and open reimplementation of {{AlphaZero}},'' in \emph{Proceedings of the 36th {{International Conference}} on {{Machine Learning}}}.\hskip 1em plus 0.5em minus 0.4em\relax {PMLR}, May 2019, pp. 6244--6253.

\bibitem{wu_accelerating_2020}
T.-R. Wu, T.-H. Wei, and I.-C. Wu, ``Accelerating and {{Improving AlphaZero Using Population Based Training}},'' \emph{Proceedings of the AAAI Conference on Artificial Intelligence}, vol.~34, no.~01, pp. 1046--1053, Apr. 2020.

\bibitem{gary_leela_2023}
\BIBentryALTinterwordspacing
L.~Gary, L.~Alexander, H.~Folkert, and {GitHub contributors}, ``Leela {{Chess Zero}}: A {{UCI-compliant}} chess engine designed to play chess via neural network,'' LCZero, Oct. 2023. [Online]. Available: \url{https://github.com/LeelaChessZero/lc0}
\BIBentrySTDinterwordspacing

\bibitem{ye_mastering_2021}
W.~Ye, S.~Liu, T.~Kurutach, P.~Abbeel, and Y.~Gao, ``Mastering {{Atari Games}} with {{Limited Data}},'' in \emph{Advances in {{Neural Information Processing Systems}}}, vol.~34.\hskip 1em plus 0.5em minus 0.4em\relax {Curran Associates, Inc.}, 2021, pp. 25\,476--25\,488.

\bibitem{lanctot_openspiel_2020}
M.~Lanctot, E.~Lockhart, J.-B. Lespiau, V.~Zambaldi, S.~Upadhyay, J.~P{\'e}rolat, S.~Srinivasan, F.~Timbers, K.~Tuyls, S.~Omidshafiei, D.~Hennes, D.~Morrill, P.~Muller, T.~Ewalds, R.~Faulkner, J.~Kram{\'a}r, B.~De~Vylder, B.~Saeta, J.~Bradbury, D.~Ding, S.~Borgeaud, M.~Lai, J.~Schrittwieser, T.~Anthony, E.~Hughes, I.~Danihelka, and J.~{Ryan-Davis}, ``{{OpenSpiel}}: {{A Framework}} for {{Reinforcement Learning}} in {{Games}},'' Sep. 2020.

\bibitem{cazenave_polygames_2020}
T.~Cazenave, Y.-C. Chen, G.-W. Chen, S.-Y. Chen, X.-D. Chiu, J.~Dehos, M.~Elsa, Q.~Gong, H.~Hu, V.~Khalidov, C.-L. Li, H.-I. Lin, Y.-J. Lin, X.~Martinet, V.~Mella, J.~Rapin, B.~Roziere, G.~Synnaeve, F.~Teytaud, O.~Teytaud, S.-C. Ye, Y.-J. Ye, S.-J. Yen, and S.~Zagoruyko, ``Polygames: {{Improved Zero Learning}},'' Jan. 2020.

\bibitem{thakoor_learning_2016}
S.~Thakoor, S.~Nair, and M.~Jhunjhunwala, ``Learning to play othello without human knowledge,'' 2016.

\bibitem{wernerduvaud_muzero_2019}
\BIBentryALTinterwordspacing
A.~H. Werner~Duvaud, ``{{MuZero}} general: {{Open}} reimplementation of {{MuZero}},'' 2019. [Online]. Available: \url{https://github.com/werner-duvaud/muzero-general}
\BIBentrySTDinterwordspacing

\bibitem{niu2023lightzero}
Y.~Niu, Y.~Pu, Z.~Yang, X.~Li, T.~Zhou, J.~Ren, S.~Hu, H.~Li, and Y.~Liu, ``Lightzero: A unified benchmark for monte carlo tree search in general sequential decision scenarios,'' in \emph{Thirty-seventh Conference on Neural Information Processing Systems Datasets and Benchmarks Track}, 2023.

\bibitem{he_deep_2016}
K.~He, X.~Zhang, S.~Ren, and J.~Sun, ``Deep {{Residual Learning}} for {{Image Recognition}},'' in \emph{Proceedings of the {{IEEE Conference}} on {{Computer Vision}} and {{Pattern Recognition}}}, 2016, pp. 770--778.

\bibitem{rosin_multiarmed_2011}
C.~D. Rosin, ``Multi-armed bandits with episode context,'' \emph{Annals of Mathematics and Artificial Intelligence}, vol.~61, no.~3, pp. 203--230, Mar. 2011.

\bibitem{fawzi_discovering_2022}
A.~Fawzi, M.~Balog, A.~Huang, T.~Hubert, B.~{Romera-Paredes}, M.~Barekatain, A.~Novikov, F.~J. R.~Ruiz, J.~Schrittwieser, G.~Swirszcz, D.~Silver, D.~Hassabis, and P.~Kohli, ``Discovering faster matrix multiplication algorithms with reinforcement learning,'' \emph{Nature}, vol. 610, no. 7930, pp. 47--53, Oct. 2022.

\bibitem{mankowitz_faster_2023}
D.~J. Mankowitz, A.~Michi, A.~Zhernov, M.~Gelmi, M.~Selvi, C.~Paduraru, E.~Leurent, S.~Iqbal, J.-B. Lespiau, A.~Ahern, T.~K{\"o}ppe, K.~Millikin, S.~Gaffney, S.~Elster, J.~Broshear, C.~Gamble, K.~Milan, R.~Tung, M.~Hwang, T.~Cemgil, M.~Barekatain, Y.~Li, A.~Mandhane, T.~Hubert, J.~Schrittwieser, D.~Hassabis, P.~Kohli, M.~Riedmiller, O.~Vinyals, and D.~Silver, ``Faster sorting algorithms discovered using deep reinforcement learning,'' \emph{Nature}, vol. 618, no. 7964, pp. 257--263, Jun. 2023.

\bibitem{hubert_learning_2021}
T.~Hubert, J.~Schrittwieser, I.~Antonoglou, M.~Barekatain, S.~Schmitt, and D.~Silver, ``Learning and {{Planning}} in {{Complex Action Spaces}},'' in \emph{Proceedings of the 38th {{International Conference}} on {{Machine Learning}}}.\hskip 1em plus 0.5em minus 0.4em\relax {PMLR}, Jul. 2021, pp. 4476--4486.

\bibitem{antonoglou_planning_2021}
I.~Antonoglou, J.~Schrittwieser, S.~Ozair, T.~K. Hubert, and D.~Silver, ``Planning in {{Stochastic Environments}} with a {{Learned Model}},'' in \emph{International {{Conference}} on {{Learning Representations}}}, Oct. 2021.

\bibitem{kool_stochastic_2019}
W.~Kool, H.~V. Hoof, and M.~Welling, ``Stochastic {Beams} and {Where} {To} {Find} {Them}: {The} {Gumbel}-{Top}-k {Trick} for {Sampling} {Sequences} {Without} {Replacement},'' in \emph{Proceedings of the 36th {International} {Conference} on {Machine} {Learning}}.\hskip 1em plus 0.5em minus 0.4em\relax PMLR, May 2019, pp. 3499--3508, iSSN: 2640-3498.

\bibitem{karnin_almost_2013}
Z.~Karnin, T.~Koren, and O.~Somekh, ``\BIBforeignlanguage{en}{Almost {Optimal} {Exploration} in {Multi}-{Armed} {Bandits}},'' in \emph{\BIBforeignlanguage{en}{Proceedings of the 30th {International} {Conference} on {Machine} {Learning}}}.\hskip 1em plus 0.5em minus 0.4em\relax PMLR, May 2013, pp. 1238--1246, iSSN: 1938-7228.

\bibitem{paszke_pytorch_2019}
A.~Paszke, S.~Gross, F.~Massa, A.~Lerer, J.~Bradbury, G.~Chanan, T.~Killeen, Z.~Lin, N.~Gimelshein, L.~Antiga, A.~Desmaison, A.~Kopf, E.~Yang, Z.~DeVito, M.~Raison, A.~Tejani, S.~Chilamkurthy, B.~Steiner, L.~Fang, J.~Bai, and S.~Chintala, ``{PyTorch}: {An} {Imperative} {Style}, {High}-{Performance} {Deep} {Learning} {Library},'' in \emph{Advances in {Neural} {Information} {Processing} {Systems}}, vol.~32.\hskip 1em plus 0.5em minus 0.4em\relax Curran Associates, Inc., 2019.

\bibitem{bellemare_arcade_2013}
M.~G. Bellemare, Y.~Naddaf, J.~Veness, and M.~Bowling, ``The arcade learning environment: an evaluation platform for general agents,'' \emph{Journal of Artificial Intelligence Research}, vol.~47, no.~1, pp. 253--279, May 2013.

\bibitem{machado_revisiting_2018}
M.~C. Machado, M.~G. Bellemare, E.~Talvitie, J.~Veness, M.~Hausknecht, and M.~Bowling, ``Revisiting the arcade learning environment: evaluation protocols and open problems for general agents,'' \emph{Journal of Artificial Intelligence Research}, vol.~61, no.~1, pp. 523--562, Jan. 2018.

\bibitem{hessel_muesli_2021}
M.~Hessel, I.~Danihelka, F.~Viola, A.~Guez, S.~Schmitt, L.~Sifre, T.~Weber, D.~Silver, and H.~V. Hasselt, ``Muesli: {{Combining Improvements}} in {{Policy Optimization}},'' in \emph{Proceedings of the 38th {{International Conference}} on {{Machine Learning}}}.\hskip 1em plus 0.5em minus 0.4em\relax {PMLR}, Jul. 2021, pp. 4214--4226.

\bibitem{haque_road_2022}
R.~Haque, T.~H. Wei, and M.~M{\"u}ller, ``On the~{{Road}} to~{{Perfection}}? {{Evaluating Leela Chess Zero Against Endgame Tablebases}},'' in \emph{Advances in {{Computer Games}}}, ser. Lecture {{Notes}} in {{Computer Science}}.\hskip 1em plus 0.5em minus 0.4em\relax {Cham}: {Springer International Publishing}, 2022, pp. 142--152.

\end{thebibliography}
\begin{IEEEbiography}[{\includegraphics[width=1in,height=1.25in,clip,keepaspectratio]{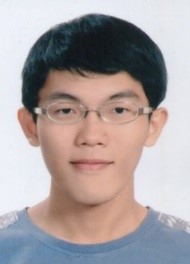}}]{Ti-Rong Wu} 
(Member, IEEE) received the B.S. degree in computer science from National Chung Cheng University, Chiayi, Taiwan, in 2014, and the Ph.D. degree in computer science from National Chiao Tung University, Hsinchu, Taiwan, in 2020.

He is currently an assistant research fellow with the Institute of Information Science, Academia Sinica, Taipei, Taiwan. His research interests include machine learning, deep reinforcement learning, artificial intelligence, and computer games. He led a team for developing the Go program, CGI, which won many competitions including second place in the first World AI Open held in Ordos, China, in 2017. He published several papers in top-tier conferences, such as AAAI, ICLR, and NeurIPS. 
\end{IEEEbiography}

\begin{IEEEbiography}[{\includegraphics[width=0.9in,height=1.1in,clip,keepaspectratio]{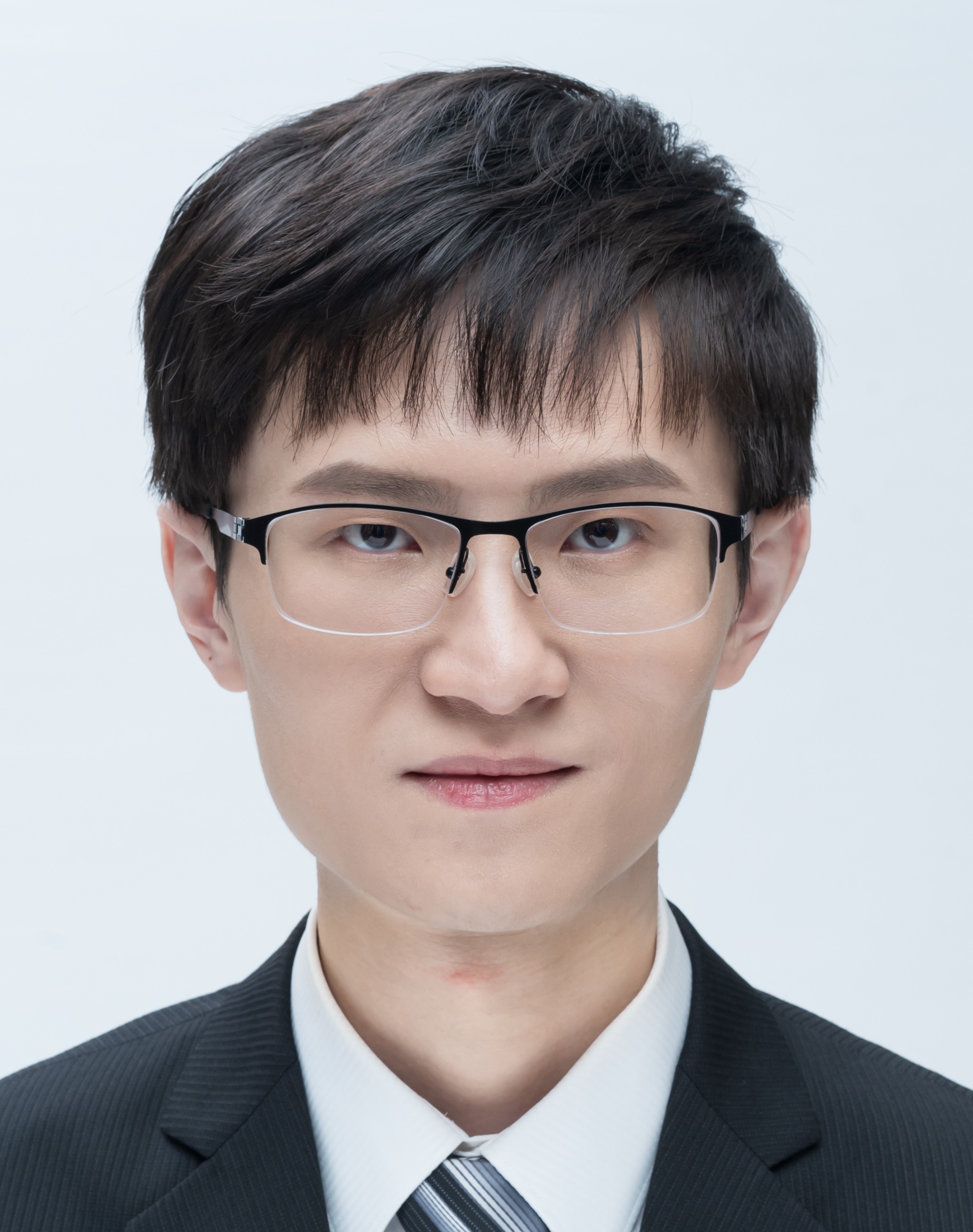}}]{Hung Guei}
(Member, IEEE) received the B.S. degree in computer science from National Central University, Taoyuan, Taiwan, in 2015, and the Ph.D. degree in computer science from National Yang Ming Chiao Tung University, Hsinchu, Taiwan, in 2023.

He is currently a postdoctoral scholar with the Institute of Information Science, Academia Sinica, Taipei, Taiwan. His research interests include reinforcement learning, artificial intelligence, and computer games. His research achievements include a state-of-the-art game-playing program for the game of 2048, which is the best-performing program based on reinforcement learning as of 2023.
\end{IEEEbiography}

\begin{IEEEbiography}[{\includegraphics[width=0.9in,height=1.1in,clip,keepaspectratio]{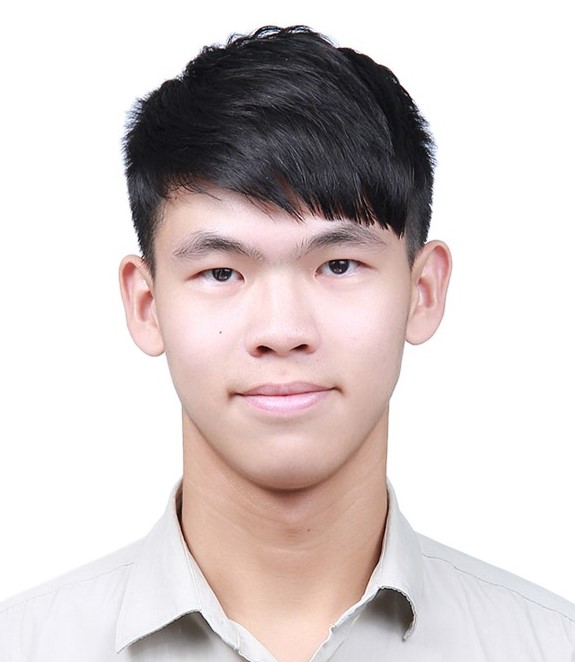}}]{Pei-Chiun Peng}
is currently working toward the M.S. degree in computer science with the Department of Computer Science, National Yang Ming Chiao Tung University, Hsinchu, Taiwan. 

His research interests include deep reinforcement learning and computer games. 
\end{IEEEbiography}

\begin{IEEEbiography}[{\includegraphics[width=0.9in,height=1.1in,clip,keepaspectratio]{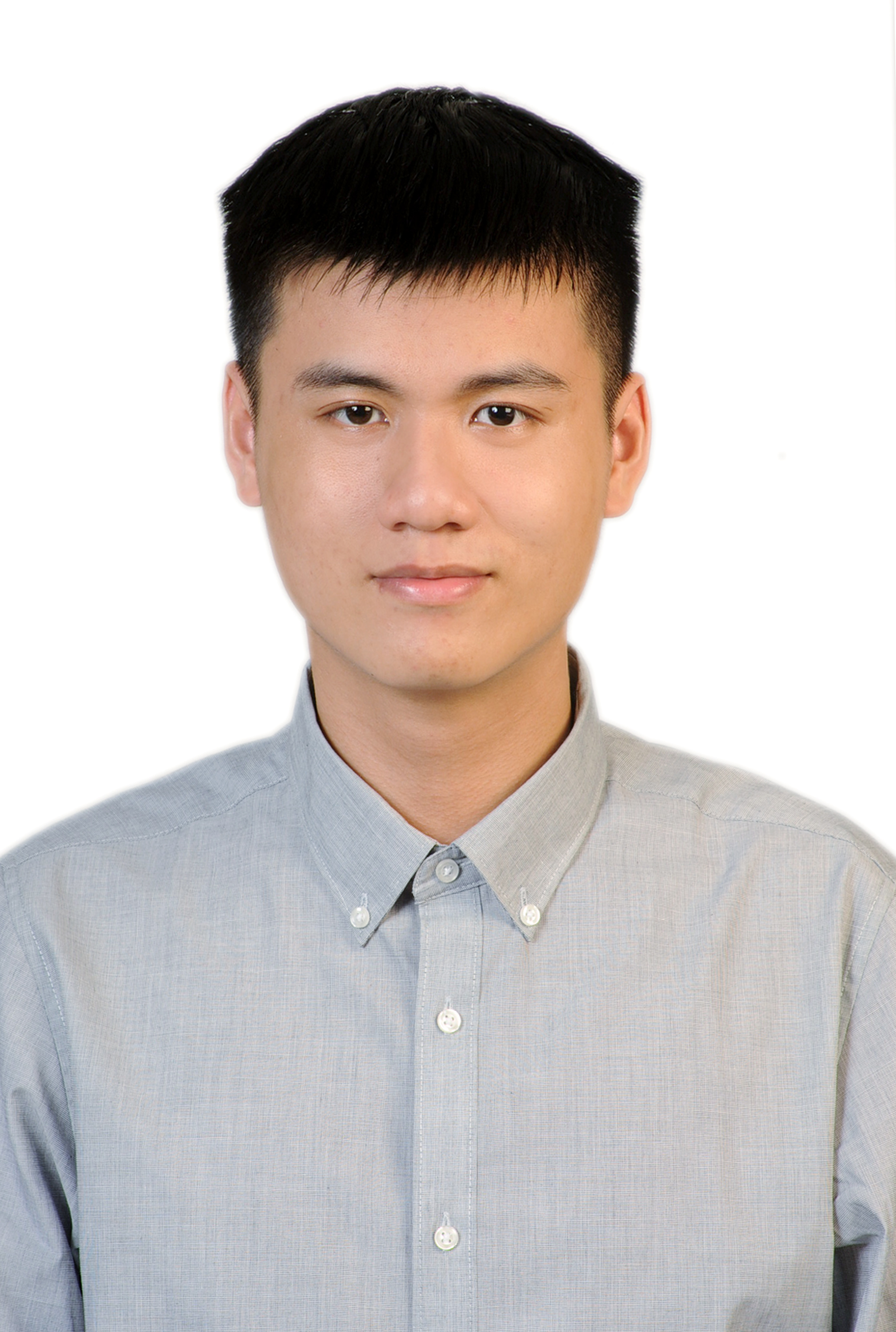}}]{Po-Wei Huang}
is currently working toward the M.S. degree in computer science with the Department of Computer Science, National Yang Ming Chiao Tung University, Hsinchu, Taiwan. 

His research interests include deep reinforcement learning and computer games. 
\end{IEEEbiography}

\begin{IEEEbiography}[{\includegraphics[width=1in,height=1.25in,clip,keepaspectratio]{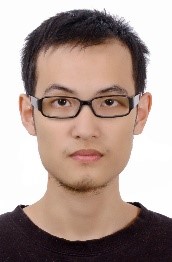}}]{Ting Han Wei}
received the Ph.D. degree in computer science from National Chiao Tung University, Hsinchu, Taiwan, in 2019.

He is currently a Professor at the School of Informatics, Kochi University of Technology, Japan. His research interests include computer games, reinforcement learning, heuristic search, and AI safety. His publications include papers in top-tier conferences, such as AAAI, IJCAI, NeurIPS, and ICLR.
\end{IEEEbiography}

\begin{IEEEbiography}[{\includegraphics[width=0.9in,height=1.1in,clip,keepaspectratio]{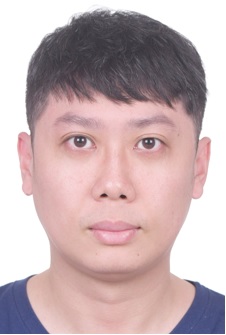}}]{Chung-Chin Shih}
(Member, IEEE) received the B.S. degree in computer science from National Chiao Tung University, Hsinchu, Taiwan, in 2011, and earned his Ph.D. degree in computer science from National Yang Ming Chiao Tung University, Hsinchu, Taiwan, in 2023.

He is currently a postdoctoral scholar with the Institute of Information Science, Academia Sinica, Taipei, Taiwan. 
His publications include papers in top-tier conferences, such as AAAI, ICLR, and NeurIPS.

His research interests include machine learning, reinforcement learning, and computer games.
\end{IEEEbiography}

\begin{IEEEbiography}[{\includegraphics[width=0.9in,height=1.1in,clip,keepaspectratio]{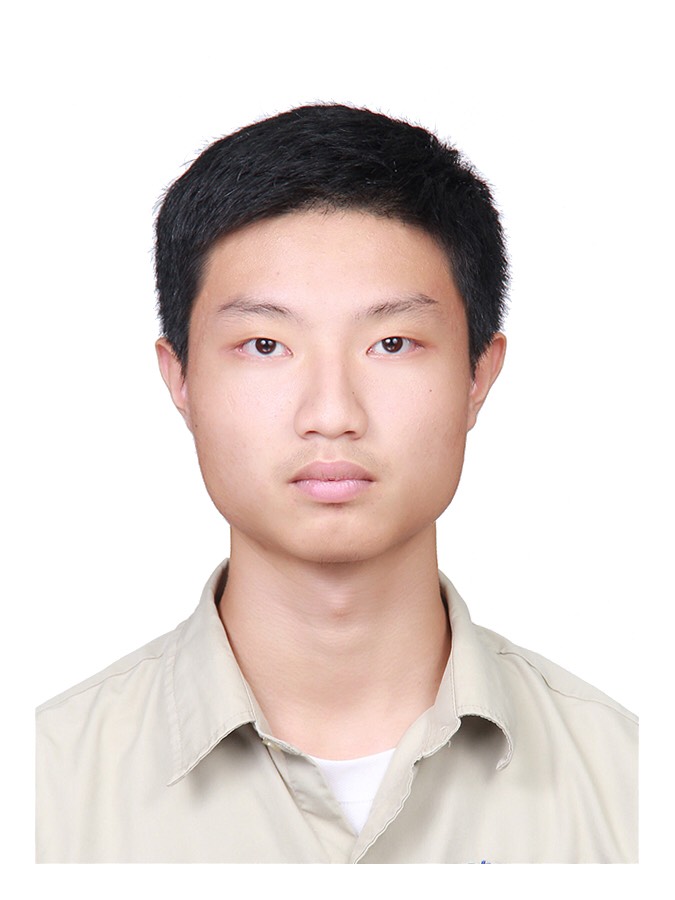}}]{Yun-Jui Tsai}
is currently working toward the M.S. degree in computer science with the Department of Computer Science, National Yang Ming Chiao Tung University, Hsinchu, Taiwan. 

His research interests include deep reinforcement learning and computer games. 
\end{IEEEbiography}

\clearpage

\clearpage
\begin{table*}[h!t]
    \centering
    \caption{Scores on 57 Atari games of different zero-knowledge learning algorithms}
    \begin{tabular}{l|rrrrr|r}
    \toprule
        Game  & ${\mu_0}$ $n=50$ & g-${\mu_0}$ $n=18$ & g-${\mu_0}$ $n=2$ & g-$\mu_0$ $n\in[2, 50]$\\ 
    \midrule
        alien  & \textbf{3,384.00} & 1,461.30 & 1,873.60 & 1,724.70 \\ 
        amidar  & \textbf{710.25} & 475.70 & 546.91 & 517.58 \\ 
        assault  & \textbf{19,447.85} & 14,933.13 & 10,786.07 & 13,870.61 \\ 
        asterix & \textbf{160,700.00} & 149,249.50 & 51,530.00 & 78,716.00 \\ 
        asteroids & 2,549.40 & 2,100.30 & \textbf{3,382.10} & 2,034.00 \\ 
        atlantis & \textbf{70,064.00} & 64,986.00 & 62,005.00 & 49,899.00 \\ 
        bank\_heist & \textbf{998.70} & 697.40 & 934.50 & 494.50 \\ 
        battle\_zone & \textbf{50,130.00} & 18,530.00 & 22,250.00 & 24,570.00 \\ 
        beam\_rider & \textbf{24,393.68} & 17,331.04 & 17,370.70 & 15,816.62 \\ 
        berzerk & 741.50 & 663.80 & \textbf{1,187.80} & 945.50 \\ 
        bowling & \textbf{57.30} & 55.07 & 54.39 & 56.75 \\ 
        boxing & \textbf{99.23} & 98.14 & 99.01 & 98.20 \\ 
        breakout & \textbf{413.02} & 333.43 & 289.65 & 304.23 \\ 
        centipede & \textbf{49,823.07} & 30,749.87 & 9,946.83 & 19,242.80 \\ 
        chopper\_command & \textbf{40,159.00} & 7,578.00 & 24,153.00 & 11,245.00 \\ 
        crazy\_climber & \textbf{126,874.00} & 89,117.00 & 47,722.00 & 32,886.00 \\ 
        defender & \textbf{52,140.00} & 38,230.50 & 43,240.50 & 41,418.50 \\ 
        demon\_attack & 85,108.10 & \textbf{100,499.85} & 94,344.80 & 99,637.15 \\ 
        double\_dunk & \textbf{-5.76} & -9.60 & -8.60 & -7.82 \\ 
        enduro & \textbf{2,274.20} & 2,052.87 & 1,368.13 & 1,702.76 \\ 
        fishing\_derby & \textbf{17.02} & -77.88 & -9.55 & -41.32 \\ 
        freeway & \textbf{26.52} & 22.54 & 21.97 & 16.92 \\ 
        frostbite & \textbf{4,530.00} & 3,331.30 & 3,589.00 & 2,503.70 \\ 
        gopher & \textbf{67,909.00} & 57,815.40 & 40,903.40 & 40,373.20 \\ 
        gravitar & 1,138.50 & 847.50 & \textbf{2,184.00} & 1,294.00 \\ 
        hero& 13,786.05 & \textbf{19,352.80} & 16,683.65 & 18,418.80 \\ 
        ice\_hockey & -6.32 & -3.66 & \textbf{-3.61} & -6.91 \\ 
        jamesbond & \textbf{4,788.00} & 984.50 & 2,181.50 & 2,347.00 \\ 
        kangaroo & \textbf{11,984.00} & 4,876.00 & 1,978.00 & 7,671.00 \\ 
        krull & \textbf{10,134.10} & 9,567.80 & 9,606.50 & 8,811.30 \\ 
        kung\_fu\_master & \textbf{51,312.00} & 37,164.00 & 29,010.00 & 31,432.00 \\ 
        montezuma\_revenge & \textbf{0.00} & \textbf{0.00} & \textbf{0.00} & \textbf{0.00} \\ 
        ms\_pacman & \textbf{4,705.00} & 3,797.70 & 3,892.30 & 2,456.50 \\ 
        name\_this\_game & \textbf{14,633.20} & 12,875.80 & 11,462.40 & 11,469.80 \\ 
        phoenix & 3,837.30 & 6,964.30 & \textbf{20,722.00} & 6,631.20 \\ 
        pitfall & \textbf{0.00} & \textbf{0.00} & \textbf{0.00} & -0.56 \\ 
        pong & 18.04 & 19.15 & \textbf{19.80} & 19.36 \\ 
        private\_eye & \textbf{12,132.96} & 98.12 & 63.91 & 39.68 \\ 
        qbert & \textbf{16,654.75} & 14,637.00 & 14,203.50 & 16,211.50 \\ 
        riverraid & \textbf{12,596.00} & 9,231.60 & 7,653.40 & 9,180.80 \\ 
        road\_runner & 47,020.00 & 46,111.00 & 48,345.00 & \textbf{49,357.00} \\ 
        robotank & 5.82 & \textbf{14.25} & 8.81 & 9.48 \\ 
        seaquest & \textbf{6,094.40} & 3,877.80 & 1,626.60 & 3,913.00 \\ 
        skiing & -31,808.33 & -31,568.41 & \textbf{-26,459.35} & -31,117.25 \\ 
        solaris & 2,090.80 & \textbf{2,637.20} & 2,202.60 & 1,993.40 \\ 
        space\_invaders & 635.10 & \textbf{931.20} & 748.50 & 512.80 \\ 
        star\_gunner & \textbf{69,601.00} & 57,910.00 & 60,586.00 & 61,437.00 \\ 
        surround & \textbf{2.03} & -8.53 & -9.45 & -7.17 \\ 
        tennis & -7.77 & -6.30 & -9.27 & \textbf{-1.49} \\ 
        time\_pilot & \textbf{11,836.00} & 7,786.00 & 9,908.00 & 5,675.00 \\ 
        tutankham & \textbf{146.33} & 142.76 & 130.77 & 117.39 \\ 
        up\_n\_down & \textbf{161,069.70} & 122,743.00 & 44,026.50 & 114,536.10 \\ 
        venture & 0.00 & 0.00 & 0.00 & \textbf{2.00} \\ 
        video\_pinball & 36,401.50 & 15,225.94 & \textbf{48,304.05} & 18,657.30 \\ 
        wizard\_of\_wor & 1,001.00 & 748.00 & \textbf{1,078.00} & 829.00 \\ 
        yars\_revenge & \textbf{73,230.71} & 60,261.63 & 63,747.76 & 53,542.43 \\ 
        zaxxon & 14,062.00 & \textbf{14,707.00} & 8,896.00 & 12,956.00 \\ 
        \hline
        human normalized mean & \textbf{485.20\%} & 395.87\% & 341.29\% & 359.97\% \\
    \bottomrule
    \end{tabular}
    \label{tab:Atari57-score}
\end{table*}

\bibliographystyle{IEEEtran}

\end{document}